\documentclass[twoside, 11pt, preprint]{article}

\usepackage{subcaption}

\usepackage{jmlr2e}

\usepackage{mathtools}

\allowdisplaybreaks

\usepackage{bm}

\usepackage{enumitem}

\setenumerate{label = (\alph*), ref=\thelemma(\alph*)}

\usepackage{tikz}
\usepackage{pgfplots}
\usepackage{xcolor}
\usepackage{placeins}

\pgfplotsset{compat=1.18}
\definecolor{cust_col}{HTML}{680369}

\newcommand*{\E}{\mathbb{E}}
\newcommand*{\R}{\mathbb{R}}
\newcommand*{\Z}{\mathbb{Z}}

\newcommand*{\bfX}{\mathbf{X}}
\newcommand*{\bfY}{\mathbf{Y}}

\newcommand*{\bft}{\mathbf{t}}
\newcommand*{\bfu}{\mathbf{u}}
\newcommand*{\bfv}{\mathbf{v}}
\newcommand*{\bfw}{\mathbf{w}}
\newcommand*{\bfx}{\mathbf{x}}
\newcommand*{\bfy}{\mathbf{y}}
\newcommand*{\bfz}{\mathbf{z}}

\newcommand*{\bftheta}{\bm{\theta}}
\newcommand*{\bfvartheta}{\bm{\vartheta}}
\newcommand*{\bfxi}{\bm{\xi}}
\newcommand*{\bfzeta}{\bm{\zeta}}

\newcommand*{\bfzero}{\mathbf{0}}
\newcommand*{\bfone}{\mathbf{1}}

\newcommand*{\calL}{\mathcal{L}}

\newcommand*{\calN}{\mathcal{N}}

\DeclareMathOperator*{\argmin}{arg\,min}

\newcommand*{\inv}{^{-1}}
\newcommand*{\tran}{^{\mathsf{t}}}

\newcommand*{\geqZ}{\Z_{\geq 0}}
\newcommand*{\geqR}{\R_{\geq 0}}
\newcommand*{\ggR}{\R_{> 0}}
\newcommand*{\simiid}{\overset{\scriptscriptstyle i.i.d.}{\sim}}
\newcommand*{\sigmax}{\sigma_{\mathrm{max}}}
\newcommand*{\sigminp}{\sigma_{\mathrm{min}}^+}
\newcommand*{\bigmid}{\ \big\vert\ }

\newcommand*{\Diag}{\mathrm{Diag}}

\DeclarePairedDelimiter{\abs}{\lvert}{\rvert}
\DeclarePairedDelimiter{\iprod}{\langle}{\rangle}
\DeclarePairedDelimiter{\norm}{\lVert}{\rVert}

\newcommand*{\Cov}{\mathrm{Cov}}
\newcommand*{\KL}{\mathrm{KL}}
\newcommand*{\Tr}{\mathrm{Tr}}

\newcommand*{\rmd}{\mathrm{d}}
\newcommand*{\op}{\mathrm{op}}
\newcommand*{\avg}{\mathrm{avg}}

\newcommand*{\weight}{\mathbf{w}}
\newcommand*{\starweight}{\weight_{*}}

\newcommand*{\starw}{w_{*}}
\newcommand*{\starbfw}{\bfw_{*}}

\newcommand*{\wtmweight}{\widetilde{W}}
\newcommand*{\wtweight}{\widetilde{\weight}}
\newcommand*{\wtL}{\widetilde{\calL}}
\newcommand*{\wtw}{\widetilde{w}}

\newcommand{\CustomQED}{\hfill \BlackBox}
\newenvironment{CustomProof}{\par\noindent{\textbf{Proof}\ }}{\CustomQED
\medskip}

\title{Training Diagonal Linear Networks with Stochastic Sharpness-Aware
Minimization}

\jmlrheading{27}{2026}{1-63}{00/00}{00/00}{00-0000}{Gabriel Clara, Sophie
Langer, and Johannes Schmidt-Hieber}

\ShortHeadings{Diagonal Linear Networks with S-SAM}{Clara, Langer, and
Schmidt-Hieber}

\firstpageno{1}

\author{\name Gabriel Clara \email gclara@berkeley.edu \\
        \addr Simons Institute for the Theory of Computing \\
        University of California, Berkeley \\
        Berkeley, CA 94720, United States of America
        \AND
        \name Sophie Langer \email s.langer@rub.de \\
        \addr Faculty of Mathematics\\
        Ruhr University Bochum\\
        44780 Bochum, Germany
        \AND
        \name Johannes Schmidt-Hieber \email a.j.schmidt-hieber@utwente.nl \\
        \addr Faculty of Electrical Engineering, Mathematics, and Computer
        Science \\
        University of Twente \\
        7522 NB, Enschede, The Netherlands
}

\editor{NA}

\begin{document}

\maketitle

\begin{abstract}
  We analyze the landscape and training dynamics of diagonal linear networks in
  a linear regression task, with the network parameters being perturbed by
  isotropic normal noise during training. The addition of such noise may be
  interpreted as a stochastic form of sharpness-aware minimization (SAM) and we
  prove several results that relate its action on the underlying landscape and
  training dynamics to the sharpness of the loss. In particular, the noise
  induces a weighted mixture of fractional norm penalties on the network
  parameters, which forces the individual layers to balance at a fast rate and
  changes the underlying landscape to favor solutions that result from a
  shrinkage-thresholding operator applied to the true parameter. We show that
  balancing the layers equates to minimizing the average sharpness, as well as
  the trace of the Hessian matrix, among all possible factorizations of the same
  linear predictor. Further, we characterize how the noise level of the normal
  perturbations acts as a regularization parameter, with exact descriptions of
  its effect on shrinkage, thresholding, and balancing speed.
\end{abstract}


\section{Introduction}

Training deep neural networks via empirical risk minimization requires solving
difficult non-convex optimization problems \citep{li_xu_et_al_2018}. The
computed networks must strike the right balance between minimizing the empirical
loss and generalizing well to unseen data. \textit{Sharpness-aware minimization}
(SAM) refers to a family of algorithmic regularization techniques, designed to
nudge optimization routines towards flat regions of the underlying loss
landscape \citep{foret_kleiner_et_al_2021}. In the context of training neural
networks, flat minima of the empirical loss are thought to generalize well to
unseen data, due to robustness against reasonable perturbations of the loss
landscape \citep{hochreiter_schmidhuber_1994}. Although empirical studies
promise improved model generalization when training with SAM
\citep{dziugaite_roy_2017, foret_kleiner_et_al_2021}, its theoretical
underpinnings are incomplete \citep{andriushchenko_flammarion_2022,
wen_li_et_al_2023, dinh_pascanu_et_al_2017}.

SAM acts via minimization of a surrogate loss that explicitly penalizes the
sharpness of points on the original loss surface. This penalty can be defined in
various ways, leading to distinct variants of SAM. Depending on the data and the
model, these variants may generate dissimilar outcomes in practice
\citep{andriushchenko_croce_et_al_2023}, highlighting the need for thorough
investigations of their action on the underlying loss landscape and the
associated training dynamics.

For a more formal discussion, fix a loss function $\calL : \R^d \to \geqR$, for
example the empirical loss of a specific model. For a neural network model, the
empirical loss surface features many local and global minima
\citep{li_xu_et_al_2018}. In particular, while each $\bftheta_* \in \argmin
\calL(\bftheta)$ will fit the observed data well, different choices of
$\bftheta_*$ may generalize better or worse to unseen data. Consequently, we
wish to make a principled choice among minimizers $\bftheta_*$, in the hope of
guaranteeing good generalization. In practice, SAM focuses on choosing a
minimizer $\bftheta_*$ with small sharpness.

The sharpness of $\calL$ around any point in its domain may be measured via the
maximal increase in loss in a small radius $\eta > 0$, leading to the definition
$S_{\max}(\bftheta, \eta) = \sup_{\norm{\bfvartheta}_2 \leq 1} \calL(\bftheta +
\eta \cdot \bfvartheta) - \calL(\bftheta)$. A flat minimum $\bftheta_*$ should
then satisfy $\bftheta_* \in \argmin \calL(\bftheta)$, while minimizing
$S_{\max}(\bftheta_*, \eta)$ among the candidate solutions. The radius $\eta$
plays a role analogous to a regularization parameter, with larger values leading
to harsher penalization of steep landscapes. Joint minimization of
$\calL(\bftheta)$ and $S_{\max}(\bftheta, \eta)$ requires solving the min-max
problem $\inf_{\bftheta} \sup_{\norm{\bfvartheta} \leq 1} \calL(\bftheta + \eta
\cdot \bfvartheta)$, which may be intractable for reasonably complex losses,
such as those based on deep neural networks.

A different perspective on the interplay between sharpness and generalization
comes from probably approximately correct Bayesian (PAC Bayes) analysis.
Minimizing the \textit{average sharpness} $S_{\avg}(\bftheta, \eta) =
\E_{\bfxi}[\calL(\bftheta + \eta \cdot \bfxi)] - \calL(\bftheta)$, where $\bfxi$
follows a standard normal distribution, yields practical generalization gap
bounds with quantifiable probability over the draw of the training data
\citep{dziugaite_roy_2017, neyshabur_bhojanapalli_et_al_2017}. The PAC Bayes
framework also predicts tighter generalization bounds when controlling for
$S_{\avg}$, rather than $S_{\max}$, but in practice neither measure can fully
capture generalization behavior \citep{dinh_pascanu_et_al_2017,
andriushchenko_flammarion_2022}. This drives the need for a fine-grained
understanding of the effective loss landscape and training trajectories under
sharpness regularization.

Both notions of sharpness lead to mathematically impractical losses for complex
models, forcing implementable algorithms to approximate. Estimating the
sensitivity of $\calL$ to small changes via a power series expansion leads to
the surrogate $\nabla \calL(\bftheta + \eta \cdot \nabla \calL(\bftheta) /
\norm{\nabla \calL(\bftheta)}_2)$ for the gradient of
$\sup_{\norm{\bfvartheta}_2 \leq 1} \calL(\bftheta + \eta \cdot \bfvartheta)$,
see \cite{bartlett_et_al_2023, wen_ma_et_al_2023}. The gradient of
$\E_{\bfxi}[\calL(\bftheta + \eta \cdot \bfxi)]$ may be approximated by taking
independent samples of $\nabla \calL(\bfvartheta + \eta \cdot \bfxi)$ as was
done in \cite{dziugaite_roy_2017}. For reasonable loss functions this yields an
unbiased estimator of the underlying gradient, so we may refer to the
optimization algorithm based on this noisy approximation as \textit{stochastic
SAM} (S-SAM). To gain a better understanding of how the algorithmic noise acts
in a deep model, we analyze S-SAM for a diagonal linear neural network
performing a regression task, with the goal of understanding both the
qualitative nature of the stationary points generated by this method, as well as
the associated change in training dynamics.

\subsection{Contributions}

We answer the following main questions in the setting of diagonal linear
networks training on the squared Euclidean loss in a regression task:
\begin{itemize}
  \item What is the regularizer induced by the average sharpness penalty and how
    does it interact with model parameters such as network depth and level of
    the S-SAM noise? (Lemma \ref{lem::grad_exp_cond})
  \item How does this regularizer affect the stationary points of the underlying
    landscape and do these points indeed feature minimal sharpness? (Theorem
    \ref{thm::crit_set} and Lemma \ref{lem::bal_reg})
  \item How does the additional gradient noise in S-SAM influence the gradient
    descent trajectory and does it indeed lead to stationary points that
    minimize sharpness? (Theorems \ref{thm::grad_flow_bal},
    \ref{thm::conv_abs_disc}, \ref{thm::grad_desc_bal}, and
    \ref{thm::proj_sam_conv})
\end{itemize} Further, we provide a first step towards understanding the more
complex geometry of the induced regularizer in general linear networks (Section
\ref{sec::full_lnn}).

To the best of our knowledge, we provide the first toy model analysis of S-SAM
that integrates landscape regularization and gradient descent dynamics, which
provides a stepping stone towards understanding its action in more practical
situations. Further, our results shed light on some interesting connections
between algorithmic noise and regularization in training diagonal linear neural
networks, which complements the existing literature on implicit and explicit
biases of gradient descent and the wider field of non-convex optimization.
Specifically, our method does not rely on a particular initialization, or a
mirror flow analysis; we show that the added normal noise directly causes
sparsity through the induced penalty.

\subsection{Related Work}

\textit{Sharpness-Aware Minimization:} The term sharpness-aware minimization was
coined by \cite{foret_kleiner_et_al_2021}, accompanied by empirical validation
of the method on different classification tasks and the estimate $S_{\max}(w,
\eta) + O(n^{- 1 / 2})$ for the generalization gap, with $n$ the number of data
points,. This result relies on a version of the well-known McAllester Bound
\citep{mcallester_1999}, meaning it only holds for bounded losses. Analogous
generalization gap estimates based on the average sharpness $S_{\avg}(w, \eta)$
are presented in \cite{dziugaite_roy_2017, neyshabur_bhojanapalli_et_al_2017,
tsuzuku_sato_et_al_2020}. Further, \cite{dziugaite_roy_2017} argue that adding
small normally distributed parameter perturbations during each gradient descent
iteration may help optimize the generalization gap. Training under parameter
perturbations also relates to robustness against adversarial noise
\citep{wu_xia_et_al_2020}. \cite{chen_zhang_et_al_2023} show that shallow
convolutional neural networks trained with SAM exhibit benign overfitting in a
binary classification task under mild conditions.

SAM and associated methods have also been studied from the perspective of
regularization. \cite{andriushchenko_bahri_et_al_2023} exhibit a bias towards
low-rank features for neural networks trained with SAM on benchmark data sets.
Connecting SAM with Bayesian variational inference, \cite{moellenhoff_khan_2023}
show that the sharpness-aware objective may be interpreted as an optimal convex
relaxation of a Bayesian objective. Directly penalizing the trace of the Hessian
matrix to regularize for flatness leads to approximately Schatten 1-norm minimal
solutions in deep matrix factorization \citep{gatmiry_li_et_al_2023}. In
general, norm-minimal parameter estimates are thought to generalize well
\citep{d_angelo_varre_et_al_2023, krogh_hertz_1991}, but the relationship
between generalization and regularization is not necessarily clear cut since
many architectures also generalize well in practice without explicit
regularization \citep{zhang_bengio_et_al_2017, zhang_bengio_et_al_2021}.

The impact of SAM on the underlying gradient descent dynamics has mainly been
studied via a first-order approximation to $S_{\max}$, as briefly recounted in
the introduction. In this setting, \cite{bartlett_et_al_2023} analyze a
quadratic loss function and illustrate a regime where the algorithm eventually
``bounces'' between opposing sides of a valley in the loss landscape. Further,
\cite{long_bartlett_2024} exhibit an edge-of-stability (EOS) regime for SAM that
features a more intricate structure than the usual EOS for unregularized
gradient descent \citep{arora_li_et_al_2022, li_wang_et_al_2022a}.
\cite{wen_ma_et_al_2023} expose a phase transition in the dynamics of SAM when
starting in the attracting set of the manifold of global minimizers. The
algorithm follows the gradient flow of the loss until coming close to a
minimizer and then tracks a penalized gradient flow on said manifold.

The idea that flat minima should generalize well traces back to
\cite{hochreiter_schmidhuber_1994}, which also describes a regularization
strategy based on volumes of boxes around the parameter to help the training
algorithm avoid sharp landscapes. However, both theoretical
\citep{dinh_pascanu_et_al_2017} and empirical
\citep{andriushchenko_flammarion_2022} studies show that sharpness in the sense
of $S_{\max}$ does not by itself determine whether a parameter value can
generalize. In fact, \cite{wen_li_et_al_2023} exhibit a pathological example in
which flat minima never generalize. 

\textit{Diagonal Linear Networks:} The gradient descent dynamics of two-layer
diagonal linear networks performing a linear regression task were recently
investigated in \cite{even_pesme_2023, pesme_flammarion_2023,
pesme_pillaud-vivien_et_al_2021}, exhibiting their implicit bias and an EOS
regime. \cite{andriushchenko_croce_et_al_2023} analyze the Hessian matrix of
two-layer diagonal linear networks at a global minimum to show that adaptive
variants of $S_{\max}$ and $S_{\avg}$ capture different generalization measures.
Diagonal linear networks form a strict subset of linear networks, which
implement arbitrary matrix factorizations. Despite the absence of non-linear
activations, these networks feature intricate non-convex landscapes
\citep{achour_malgouyres_et_al_2024, kawaguchi_2016} and their training dynamics
are an active area of research \citep{arora_cohen_et_al_2018,
arora_cohen_et_al_2019a, arora_cohen_et_al_2019b, chitour_liao_et_al_2023,
chou_gieshoff_et_al_2024, bah_rauhut_et_al_2021, nguegnang_rauhut_et_al_2024,
zhao_xu_2024, xu_min_et_al_2023}. In comparison with analogous results for
non-linear networks \citep{arora_du_et_al_2019, chatterjee_2022,
du_lee_et_al_2019, elkabetz_cohen_et_al_2021, forti_nistri_et_al_2006}, linear
networks allow for finer characterizations of trajectories and landscape.

\textit{Noisy Gradient Descent:} Analyses of gradient descent with various forms
of additional noise exist in the literature, with a popular variant being
stochastic gradient Langevin dynamics (SGLD)
\citep{raginsky_rakhlin_et_al_2017}. In particular, the Langevin noise helps the
algorithm exit sharp minima exponentially fast for locally quadratic losses
\citep{ibayashi_imaizumi_2023, zhu_wu_et_al_2019} and can generate
differentially private estimates \citep{avella-medina_bradshaw_et_al_2023}.
Injecting parameter noise into the algorithm may lead to better generalization
\citep{orvieto_kersting_et_al_2022, orvieto_raj_et_al_2023} and has been studied
in the optimization literature under the name randomized smoothing
\citep{duchi_bartlett_et_al_2012}. The landscape and gradient descent dynamics
of the underlying regularized problem for dropout noise applied to linear neural
networks were analyzed in \cite{mianjy_arora_2019, senen-cerda_sanders_2022},
with the regularizer inducing both shrinkage and thresholding of the estimated
singular values. A full analysis of the interaction between gradient descent
dynamics and added dropout noise in a linear regression model is conducted in
\cite{clara_langer_et_al_2024}. The asymptotic dynamics of noisy gradient
descent in the small step-size limit are characterized in
\cite{li_wang_et_al_2022a, shalova_schlichting_et_al_2024}.

\textit{Mirror Flows:} The individual layers in a neural network update
separately, which induces complex dynamics on the function implemented by the
network. Under specific initializations, the predictor in a shallow diagonal
network follows a closed-form mirror flow that exhibits implicit bias towards
sparse solutions \citep{vaskevicius_kanade_et_al_2019, even_pesme_2023,
pesme_flammarion_2023}. Stochasticity in the flow can lead to better
generalization \citep{pesme_pillaud-vivien_et_al_2021}, as well as help recover
sparse solutions in settings where the gradient descent dynamics do not exhibit
an implicit sparsity bias \citep{woodworth_gunasekar_et_al_2020,
pesme_pillaud-vivien_et_al_2021}. In contrast, results for deep diagonal linear
networks seem less developed, with the corresponding mirror flow generally
characterized only existentially rather than in closed form
\citep{labarriere_molinari_et_al_2024}.

\subsection{Organization}

Section \ref{sec::lin_nn_back} starts with a brief introduction of diagonal
linear networks and the S-SAM algorithm, along with some preliminary results.
Section \ref{sec::land_no_noise} details our analysis of the expected loss
function, resulting from marginalization of the algorithmic noise inherent to
S-SAM. Section \ref{sec::conv_grad_desc} provides a dynamical analysis of the
gradient descent recursion. Section \ref{sec::full_lnn} discusses the more
general setting of fully connected linear networks. Section \ref{sec::disc}
concludes with some further discussions and an outlook on future research.

The proofs of all numbered statements in the main text are deferred to
Appendices \ref{sec::lin_nn_back_proof}, \ref{sec::land_no_noise_proof},
\ref{sec::conv_grad_desc_proof}, and \ref{sec::full_lnn_proof}. Appendix
\ref{sec::sim} details simulation settings. The remaining Appendices
\ref{sec::gd_conv} and \ref{sec::approx_root} contain auxiliary results and
discuss some technical nuances.

\subsection{Notation}

Vectors are written in boldface. We write $\bfu\tran \bfv$ and $\bfu \bfv\tran$
for the standard inner and outer products in Euclidean space. The symbols
$\bfzero$ and $\bfone$ always signify the vectors with every entry respectively
equaling $0$ and $1$. Scalar functions, such as powers and absolute value, apply
element-wise when taking a vector argument. For example, we may write
$\norm{\bfv}^q_q = \bfone\tran \abs{\bfv}^q$ for $q > 0$.  Let $\bfu \odot \bfv$
denote the element-wise product of two vectors, with the convention that
$\bigodot_{\ell = r + s}^{r} \bfw_\ell = \bfone$ whenever $r, s > 0$.

For a matrix $A$, let $A\tran$ and $\Tr(A)$ denote the transpose and trace of
$A$. We write $\sigmax(A)$ and $\sigminp(A)$ for the largest and smallest
non-zero singular values of $A$, with their ratio $\kappa(A) = \sigmax(A) /
\sigminp(A)$ giving the effective condition number. For square matrices,
$\Diag(A)$ denotes the diagonal matrix with the same main diagonal as $A$.

Given any two functions $f, g : \R^d \to \R$, we write $f(\bfx) = O(g(\bfx))$,
as $\bfx \to \bfy$, to signify that $\limsup_{\bfx \to \bfy} \abs{f(\bfx) /
g(\bfx)} < \infty$. For a finite set $C$, its cardinality is denoted by $\# C$.

\section{Learning Diagonal Linear Networks with S-SAM}
\label{sec::lin_nn_back}

Diagonal linear networks form one of the simplest non-trivial classes of neural
networks, featuring linear activations and no cross connections between hidden
layers, see Figure \ref{fig::diag_net} for an example. Despite their simplicity,
diagonal linear networks are fundamentally non-convex and provide a valuable toy
model for the mathematical study of deep learning.

\begin{figure}
  \centering
  \begin{tikzpicture}[very thick]
    \node[shape=circle, draw=cust_col, label=below:$x_1$] (i1) at (0, 0) {};
    \node[shape=circle, draw=cust_col, label=below:$x_2$] (i2) at (3, 0) {};
    \node[shape=circle, draw=cust_col, label=below:$x_d$] (id) at (6, 0) {};

    \node[shape=circle, draw=cust_col] (l11) at (0, 1.5) {};
    \node[shape=circle, draw=cust_col] (l12) at (3, 1.5) {};
    \node[shape=circle, draw=cust_col] (l1d) at (6, 1.5) {};

    \node[shape=circle, draw=cust_col] (ll1) at (0, 5) {};
    \node[shape=circle, draw=cust_col] (ll2) at (3, 5) {};
    \node[shape=circle, draw=cust_col] (lld) at (6, 5) {};

    \node[shape=circle, draw=cust_col, label=above:$\sum_{i = 1}^d w_{L, i}
    \cdots w_{1, i} x_i$] (y) at (3, 7) {};

    \node (lil1) at (0, 3) {};
    \node (lil2) at (3, 3) {};
    \node (lild) at (6, 3) {};

    \node (lil11) at (0, 2.5) {};
    \node (lil12) at (3, 2.5) {};
    \node (lil1d) at (6, 2.5) {};

    \node (liu11) at (0, 4) {};
    \node (liu12) at (3, 4) {};
    \node (liu1d) at (6, 4) {};

    \node (liu1) at (0, 3.5) {};
    \node (liu2) at (3, 3.5) {};
    \node (liud) at (6, 3.5) {};

    \node (l1l) at (4, 0.75) {};
    \node (l1r) at (5, 0.75) {};

    \node (l2l) at (4, 2.25) {};
    \node (l2r) at (5, 2.25) {};

    \node (lll) at (4, 4.25) {};
    \node (llr) at (5, 4.25) {};

    \path[dotted, draw=cust_col] (l1l) edge node {} (l1r);
    \path[dotted, draw=cust_col] (l2l) edge node {} (l2r);
    \path[dotted, draw=cust_col] (lll) edge node {} (llr);

    \path[dotted, draw=cust_col] (lil11) edge node {} (liu11);
    \path[dotted, draw=cust_col] (lil12) edge node {} (liu12);
    \path[dotted, draw=cust_col] (lil1d) edge node {} (liu1d);

    \path[-, draw=cust_col] (i1) edge node[label=left:$w_{1, 1} x_1$] {} (l11);
    \path[-, draw=cust_col] (i2) edge node[label=left:$w_{1, 2} x_2$] {} (l12);
    \path[-, draw=cust_col] (id) edge node[label=right:$w_{1, d} x_d$] {} (l1d);

    \path[-, draw=cust_col] (l11) edge node[label=left:$w_{2, 1} w_{1, 1} x_1$]
    {} (lil1);
    \path[-, draw=cust_col] (l12) edge node[label=left:$w_{2, 2} w_{1, 2} x_2$]
    {} (lil2);
    \path[-, draw=cust_col] (l1d) edge node[label=right:$w_{2, d} w_{1, d} x_d$]
    {} (lild);

    \path[-, draw=cust_col] (liu1) edge node[label=left:$w_{L, 1} \cdots w_{1,
    1} x_1$] {} (ll1);
    \path[-, draw=cust_col] (liu2) edge node[label=left:$w_{L, 2} \cdots w_{1,
    2} x_2$] {} (ll2);
    \path[-, draw=cust_col] (liud) edge node[label=right:$w_{L, d} \cdots w_{1,
    d} x_d$] {} (lld);

    \path[-, draw=cust_col] (ll1) edge node {} (y);
    \path[-, draw=cust_col] (ll2) edge node {} (y);
    \path[-, draw=cust_col] (lld) edge node {} (y);
  \end{tikzpicture}
  \caption{A diagonal linear network with $L$ layers that implements the
  non-convex function $f(x_1, \ldots, x_d) = \sum_{i = 1}^d w_{L, i} \cdots
  w_{1, i} x_i$. \label{fig::diag_net}}
\end{figure}

Given $n \geq 1$ observations of labeled data $(\bfX_i, Y_i)$, the usual linear
regression setup models the relationship between the labels $Y_i \in \R$ and the
regressor $\bfX_i \in \R^d$ as $Y_i \approx \bfX_i\tran \weight$, with the goal
of learning the linear predictor $\weight = (\bfw_1, \ldots, \bfw_d)\tran$
through minimization of the empirical loss \begin{align}
  \label{eq::lin_reg_def}
  \weight \mapsto \dfrac{1}{n} \cdot \sum_{i = 1}^n \big(Y_i - \bfX_i\tran
  \weight\big)^2.
\end{align} Given sufficient amounts of data, the empirical loss approximates
the population loss $\E\big[(Y_0 - \bfX_0\tran \weight)^2\big]$, provided the
population data distribution $(\bfX_0, Y_0)$ is square-integrable. For $d > n$,
the interpolating solutions $\starbfw$ that satisfy $Y_i = \bfX_i\tran
\starweight$ for every $i = 1, \ldots, n$ form an affine subspace of dimension
at least $d - n$. While each of these solutions behaves identically on the
training set, they can generate completely different predictions $
\bfX_{\mathrm{new}}\tran \starweight$ for a newly observed
$\bfX_{\mathrm{new}}$. We refer the reader to Chapter 2 of \cite{pesme_2024} for
further discussion.

The diagonal linear model factorizes each entry of $\weight$ into $L$ layers via
the product $\weight = \weight_L \odot \cdots \odot \weight_1$, also known as
the Hadamard parametrization \citep{hoff_2017}. Writing $\bftheta = \big(\bfw_1,
\ldots, \bfw_L\big)$ for the diagonal network parameter, this yields the
empirical risk \begin{align}
  \label{eq::diag_reg_def}
  \bftheta \mapsto \dfrac{1}{n} \cdot \sum_{i = 1}^n \left(Y_i - \bfX_i\tran
  \bigodot_{\ell = 1}^L \bfw_\ell\right)^2.
\end{align} While this problem remains convex as a function of the linear
predictor $\bfw$, it is non-convex as a joint function of the individual weight
vectors $\weight_\ell$ and can hence serve as a toy model for the non-convexity
encountered in more complex learning problems.

Write $X$ for the $(n \times d)$-matrix with rows $n^{- 1 / 2} \cdot
\bfX_i\tran$ and $\bfY$ for the length $n$ vector with entries $n^{- 1 / 2}
\cdot Y_i$. The empirical loss \eqref{eq::diag_reg_def} then admits the
representation \begin{align}
  \bftheta &\mapsto \dfrac{1}{n} \cdot \sum_{i =
  1}^n\ \Big(Y_i - \bfX_i\tran \big(\weight_L \odot \cdots \odot
  \weight_1\big)\Big)^2 \nonumber \\ 
  &= \norm[\Big]{\bfY - X \big(\weight_L \odot \cdots \odot
  \weight_1\big)}^2_2. \label{eq::emp_loss_int}
\end{align} For any product of vectors, we may write $\bfu \odot \bfv = U V
\bfone$, with $U$ and $V$ diagonal matrices that have the entries of $\bfu$ and
$\bfv$ as their main diagonal. Consequently, the loss \eqref{eq::emp_loss_int}
forms a special case of the matrix factorization loss \begin{align*}
  \big(W_1, \ldots, W_L\big) \mapsto \norm[\big]{Y -
  W_L \cdots W_1 X}^2
\end{align*} where $X, Y, W_1, \ldots, W_L$ now signify arbitrary (dense)
matrices of prescribed dimensions and $\norm{A}^2 = \Tr(A A\tran)$. The
optimization problem \eqref{eq::diag_reg_def} hence sits in-between linear
regression and training a general linear neural network, which will be further
discussed in Section \ref{sec::full_lnn}.

In practice, deep neural networks train via stochastic gradient descent on
massive data sets. In its most basic form, meaning SGD with batch-size $1$ and
without additional regularization, this leads to the recursive update rule
\begin{align}
  \label{eq::sgd_reg_def}
  \weight^{\mathrm{SGD}}_\ell(k + 1) = \weight^{\mathrm{SGD}}_\ell(k) - \alpha_k
  \cdot \nabla_{\weight^{\mathrm{SGD}}_\ell(k)} \bigg(Y_k - \bfX_k\tran
  \Big(\weight^{\mathrm{SGD}}_L(k) \odot \cdots \odot
  \weight^{\mathrm{SGD}}_1(k)\Big)\bigg)^2,
\end{align} where $(\bfX_k, Y_k)$ is drawn uniformly from the available data and
$\alpha_k > 0$ denotes a user-supplied sequence of step-sizes. Computing the
gradient on the loss of a single data point acts as a noisy evaluation of the
full-batch gradient. Accordingly, SGD defines a noisy discretization of the
continuous-time gradient system \begin{align}
  \label{eq::sgd_gf_def}
  \dfrac{\rmd}{\rmd t} \weight_\ell(t) = - \nabla_{\weight_\ell(t)}
  \norm[\Big]{\bfY - X \big(\weight_L(t) \odot \cdots \odot
  \weight_1(t)\big)}^2_2, \qquad \ell = 1, \ldots, L
\end{align} which was studied in \cite{arora_cohen_et_al_2018} and
\cite{bah_rauhut_et_al_2021} for general matrix factorizations.

In the notation of the introduction, $S_{\avg}(\bftheta, \eta) =
\E_{\bfxi}\big[\calL(\bftheta + \bfxi)\big] - \calL(\bftheta)$ for any loss
$\calL$ of $\bftheta$, with $\bfxi$ denoting isotropic normal noise of level
$\eta$. In particular, minimizing $\E_{\bfxi}\big[\calL(\bftheta + \bfxi)\big] =
\calL(\bftheta) + S_{\avg}(\bftheta, \eta)$ equates to joint minimization of the
original loss and the expected sharpness. Stochastic gradient descent with S-SAM
for the diagonal linear model \eqref{eq::diag_reg_def} hence considers the
expected objective function \begin{align*}
  \bftheta \mapsto \E\left[\dfrac{1}{n} \cdot \sum_{i = 1}^n \Big(Y_i -
  \bfX_i\tran \big(\wtweight_L \odot \cdots \odot
  \wtweight_1\big)\Big)^2\right],
\end{align*} where $\wtweight_\ell$ denotes the randomly perturbed version of
$\weight_\ell$, namely $\wtweight_\ell = \weight_\ell + \bfxi_\ell$ with
$\bfxi_\ell \simiid \calN(\bfzero, \eta^2 \cdot I_d)$. The noise level $\eta >
0$ is a tuning parameter. Using the same
argument as in \eqref{eq::emp_loss_int}, this expected loss can be rewritten as
\begin{align}
  \label{eq::sam_exp_loss}
  \bftheta \mapsto \E\bigg[\norm[\Big]{\bfY - X \big(\wtweight_L \odot \cdots
  \odot \wtweight_1\big)}^2_2\bigg].
\end{align} The marginalized noise
should encourage flat minima \citep{dziugaite_roy_2017} as
integrating over the normal perturbations averages the loss over a neighborhood
of the parameters, with the tuning parameter $\eta > 0$ determining the size of
the neighborhood. Sharp points on the original loss surface are thus penalized
via marginalization of the normal noise, see Figure \ref{fig::sam_marg} for an
illustration.

\begin{figure}[tb]
  \begin{center}
    \begin{tikzpicture}[scale = 0.85]
      \begin{axis}[very thick,
        axis lines = none]
        \addplot[cyan,
          domain = -2:2,
          samples = 500] {
            abs(x)
          };
        \addplot[magenta,
          domain = -2:2,
          samples = 500] {
            exp(-x^2)
          };
      \end{axis}
    \end{tikzpicture}
    \hspace{1cm}
    \begin{tikzpicture}[scale = 0.85]
      \begin{axis}[very thick,
        axis lines = none]
        \addplot[cyan,
          domain = -2:2,
          samples = 500] {
            x^4/8
          };
        \addplot[magenta,
          domain = -2:2,
          samples = 500] {
            exp(-x^2)
          };
      \end{axis}
    \end{tikzpicture}
  \end{center}
  \caption{Sharp minimum (left) and flat minimum (right) with normal density
  centered at the minimum. The area between the function graph and density curve
  gives a measure of flatness.}
  \label{fig::sam_marg}
\end{figure}

Taking independent samples of the underlying random variables, the S-SAM
recursion with batch-size $1$ now takes the form \begin{align}
  \label{eq::sgd_sam_def}
  \weight_\ell(k + 1) = \weight_\ell(k) - \alpha_k \cdot
  \nabla_{\weight_\ell(k)} \bigg(Y_k - \bfX_k\tran \Big(\wtweight_L(k) \odot
  \cdots \odot \wtweight_1(k)\Big)\bigg)^2,
\end{align} where the normal perturbations $\wtweight_\ell(k) = \weight_\ell(k)
+ \bfxi_\ell(k)$ are sampled independently across iterations. The additional
randomness in each iteration \eqref{eq::sgd_sam_def} depends solely on $\bfX_k$
and the perturbations $\bfxi_\ell(k)$, meaning \eqref{eq::sgd_sam_def} defines a
Markov process.

\begin{figure}[t]
  \centering

  \begin{subfigure}[t]{0.4\textwidth}
    \centering
    \includegraphics[width=\textwidth]{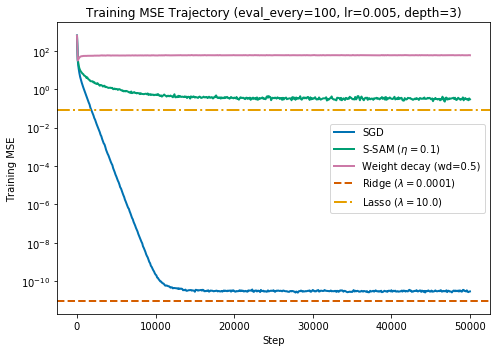}
    \caption{Training error, weights initialized close to 1}
    \label{fig::d3_init1_train}
  \end{subfigure}
  \hfill
  \begin{subfigure}[t]{0.4\textwidth}
    \centering
    \includegraphics[width=\textwidth]{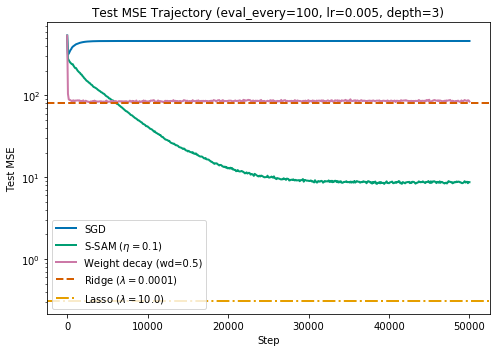}
    \caption{Test error, weights initialized close 1}
    \label{fig::d3_init1_test}
  \end{subfigure}

  \vspace{0.5cm}

  \begin{subfigure}[t]{0.4\textwidth}
    \centering
    \includegraphics[width=\textwidth]{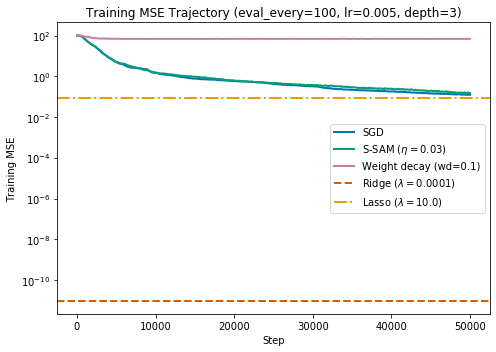}
    \caption{Training error, weights initialized close 0}
    \label{fig::d3_init0_train}
  \end{subfigure}
  \hfill
  \begin{subfigure}[t]{0.4\textwidth}
    \centering
    \includegraphics[width=\textwidth]{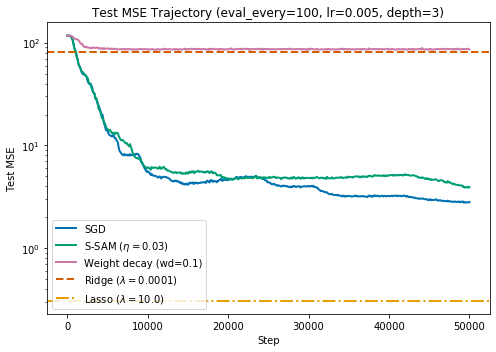}
    \caption{Test error, weights initialized close 0}
    \label{fig::d3_init0_test}
  \end{subfigure}

  \caption{Training errors (a) and test errors (b) for a diagonal linear network
  with 3 hidden layers and weights initialized closed to one versus training
  errors (c) and test errors (d) for the same network with weights initialized
  closed to zero. The $y$-axes are shown in $\log$-scale. For simulation details
  see Appendix \ref{sec::sim}.} \label{fig::depth_3_setting}
\end{figure}

As shown in Figure \ref{fig::depth_3_setting}, the addition of parameter
noise can help the algorithm to correctly identify underlying features, even in
settings where SGD, with or without weight decay, cannot. For the unregularized
algorithm, the size of the initialization determines whether it operates in the
kernel or feature learning regime, thereby changing the induced bias in the
mirror flow \citep{woodworth_gunasekar_et_al_2020}. We will show that the
ability of S-SAM to induce feature learning follows directly from landscape and
trajectory regularization, without passing to a restrictive setting that admits
a well-defined mirror flow. Simulation details and additional discussion are
provided in Appendix \ref{sec::sim}.

For the sake of brevity, we will omit the iteration index $k$ whenever
convenient and write $\bftheta_k = \big(\weight_1(k), \ldots,
\weight_L(k)\big)$, $\wtweight(k) = \wtweight_L(k) \odot \cdots \odot
\wtweight_1(k)$, and $\weight(k) = \weight_L(k) \odot \cdots \weight_1(k)$ from
this point on. Given any scalar function $f$ of the factorized predictor, the
chain rule implies $\nabla_{\weight_\ell} f = \nabla_{\weight_\ell} f \odot
\weight_L \odot \cdots \odot \weight_{\ell + 1} \odot \weight_{\ell - 1} \odot
\cdots \odot \weight_1$. Further, $\partial \wtweight_{\ell, j} / \partial
\weight_{\ell, j} = 1$ and so the gradients in \eqref{eq::sgd_sam_def} take the
form \begin{align}
  \label{eq::backprop_noise}
  \nabla_{\weight_\ell} \big(Y - \bfX\tran \wtweight\big)^2 = - 2 \cdot \bfX
  \big(Y - \bfX\tran \wtweight\big) \bigodot_{\substack{m = 1\\ m \neq \ell}}^L
  \wtweight_{m}
\end{align} Taking $\eta = 0$, in which case $\bfxi_\ell = \bfzero$ almost
surely, we also recover an expression for the gradients in
\eqref{eq::sgd_reg_def}. For $m \neq \ell$, each noisy vector $\wtweight_m$
appears squared in the gradient with respect to $\weight_\ell$. As
$\bfxi^2_{\ell, i}$ follows a $\chi^2$-distribution, $\E[\wtweight_m^2] \neq
\weight_m^2$ and so the noisy gradients in \eqref{eq::backprop_noise} do not
give an unbiased estimate for the noiseless full-batch gradients
$\nabla_{\weight_\ell} \norm{\bfY - X (\weight_L \odot \cdots \odot
\weight_1)}^2_2$. Instead, the noisy gradients estimate the gradients of a
regularized loss, which we will compute in the next lemma.

\begin{lemma}
  \label{lem::grad_exp_cond}
  Let $(\bfX, Y)$ follow the uniform distribution over available data points,
  then, for every $\ell = 1, \ldots, L$ \begin{align*}
    \E\Big[\nabla_{\weight_\ell} \big(Y - \bfX\tran \wtweight \big)^2 \bigmid
    \bftheta\Big] = \nabla_{\weight_\ell} \calL_R(\bftheta)
  \end{align*} with regularized loss $\calL_R = \calL + R$ given by the
  expectation \eqref{eq::sam_exp_loss}, evaluating to \begin{align*}
    \calL(\bftheta) &= \norm[\big]{\bfY - X \bfw}_2^2\\
    R(\bftheta) &= \bfone\tran \Diag\big(X\tran X\big) \left(\bigodot_{\ell = 1}^L
    \big(\weight_\ell^2 + \eta^2 \cdot \bfone\big) - \bigodot_{m = 1}^L
    \weight_m^2\right).
  \end{align*}
\end{lemma}

In case $L = 2$ and $\Diag(X\tran X) = I_d$, the regularizer reduces (up to an
additive constant) to $R(\bftheta) = \eta^2 \cdot (\norm{\weight_1}^2_2 +
\norm{\weight_2}^2_2)$, which matches the commonly used weight-decay method
\citep{krogh_hertz_1991, d_angelo_varre_et_al_2023}. This connection in the
two-layer case was already noted by \cite{orvieto_raj_et_al_2023}. For deep
networks, $R$ instead penalizes the squared norms of all possible partial
products of the vectors $\weight_\ell$, weighted by powers of $\eta^2$. Indeed,
expanding the product in the definition of $R$, commutativity of the
element-wise product and $\norm{\bfv}_2^2 = \bfone\tran \bfv^2$ yield
\begin{align}
  \label{eq::reg_poly}
  R(\bftheta) = \sum_{I \subsetneq \{1, \ldots, L\}} \eta^{2 (L - \# I)} \cdot
  \norm*{\sqrt{\Diag\big(X\tran X\big)} \bigodot_{m \in I} \weight_m}^2_2,
\end{align} where $\# I$ denotes the cardinality of $I$. This regularizer
provides convenient upper-bounds on polynomials in the variables $\weight_1^2,
\ldots, \weight_L^2$, which is one of the key insights underlying the results
presented in Section \ref{sec::conv_grad_desc}. The sum above always contains
the weight-decay term \begin{align}
  \label{eq::reg_wd}
  \bftheta \mapsto \eta^{2 L - 2} \cdot \sum_{\ell = 1}^L\
  \norm[\Big]{\sqrt{\Diag\big(X\tran X\big)} \weight_\ell}^2_2,
\end{align} but the training curves in Figure \ref{fig::depth_3_setting}
indicate that S-SAM and SGD with weight decay compute different solutions when
$L > 2$. From this point on, we will write \begin{equation*}
  S_{\avg}(\bftheta) = \E\left[\dfrac{1}{n} \cdot \sum_{i = 1}^n \big(Y_i -
  \bfX_i\tran \wtweight\big)^2\right] - \dfrac{1}{n} \cdot \sum_{i = 1}^n
  \big(Y_i - \bfX_i\tran \weight\big)^2
\end{equation*} for the average sharpness of the empirical loss
\eqref{eq::diag_reg_def}. Lemma \ref{lem::grad_exp_cond} further proves that
$S_{\avg} = \calL_R - \calL = R\geq 0$, so we expect $R$ to explicitly control
the average sharpness. For sufficiently small $\eta$, the average sharpness and
hence $R$ may be approximated by a constant multiple of $\eta^2 \cdot
\Tr(\nabla^2 \calL)$, giving a generalization measure that directly reflects the
local curvature of the loss. For a function with potentially unbounded third
derivative, such as the loss $\calL(\bftheta)$, the quality of this
approximation degrades with increasing parameter norm. See Appendix H of
\cite{tsuzuku_sato_et_al_2020} for a quantitative discussion.

In general, the S-SAM recursion \eqref{eq::sgd_sam_def} will not yield an exact
critical point of $\calL$, but rather a critical point of the regularized loss
$\calL_R$, as computed in Lemma \ref{lem::grad_exp_cond}. The recursion
\eqref{eq::sgd_sam_def} can be rewritten as \begin{equation}
  \label{eq::rec_split_noise}
  \weight_\ell(k + 1) = \weight_\ell(k) - \alpha_k \cdot
  \nabla_{\weight_\ell(k)} \calL_R(\bftheta_k) - \alpha_k \cdot
  \Bigg(\nabla_{\weight_\ell(k)} \Big(Y_k - \bfX\tran_k \wtweight(k)\Big)^2 -
  \nabla_{\weight_\ell(k)} \calL_R(\bftheta_k)\Bigg).
\end{equation} This separates each iteration into a deterministic part, governed
by the gradient of $\calL_R$, and a stochastic perturbation term. By Lemma
\ref{lem::grad_exp_cond}, these perturbations form a martingale difference
sequence. In analogy with \eqref{eq::sgd_gf_def}, we may then interpret the
S-SAM recursion \eqref{eq::sgd_sam_def} as a noisy discretization of the
gradient system \begin{align}
  \label{eq::sgd_sam_gf}
  \dfrac{\rmd}{\rmd t} \weight_\ell(t) = - \nabla_{\weight_\ell(t)}
  \calL_R(\bftheta_t).
\end{align} For small enough step-sizes $\alpha_k$, the S-SAM recursion
\eqref{eq::sgd_sam_def} should approximate the trajectories of
\eqref{eq::sgd_sam_gf} sufficiently well \citep{dereich_kassing_2024}, yielding
convergence to a critical point of $\calL_R$.

To the best of our knowledge, neither the dynamics \eqref{eq::rec_split_noise}
nor the regularized landscape $\calL_R$ have been analyzed in the literature. We
will first study the critical points of $\calL_R$ in Section
\ref{sec::land_no_noise}, before returning to the associated gradient descent
dynamics in Section \ref{sec::conv_grad_desc}.

\section{Landscape of the Marginalized Loss}
\label{sec::land_no_noise}

\begin{figure}[t]
  \centering
  \scalebox{1}{\begin{subfigure}{0.495\textwidth}
    \centering
    \includegraphics[width=\textwidth]{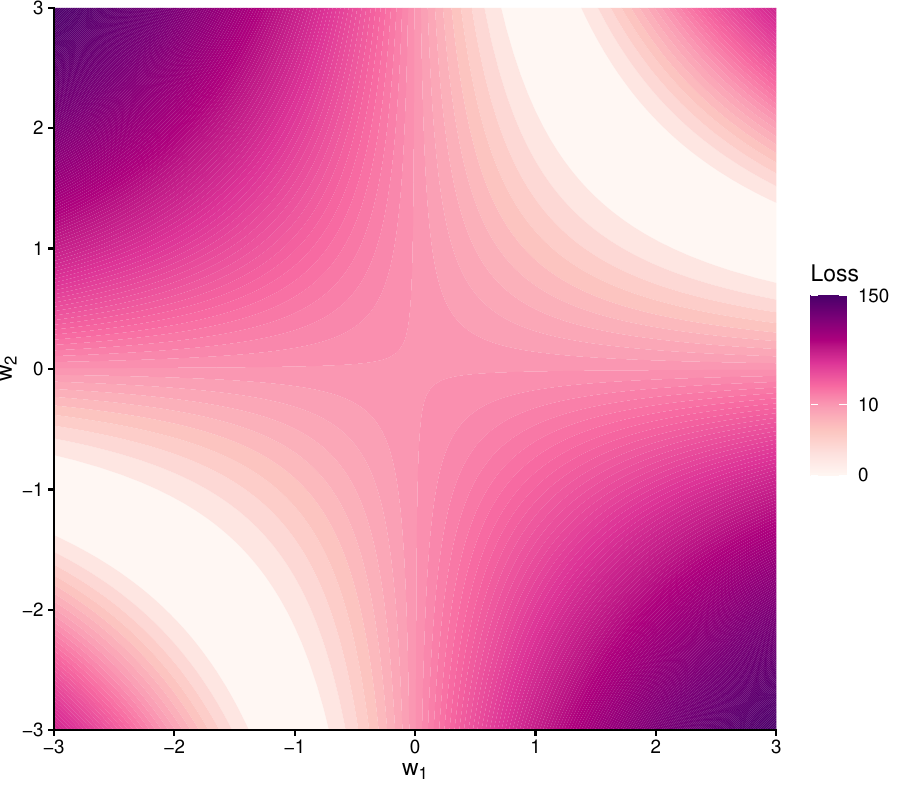}
    \caption{$\eta = 0$}
  \end{subfigure}
  \hfill
  \begin{subfigure}{0.495\textwidth}
    \centering
    \includegraphics[width=\textwidth]{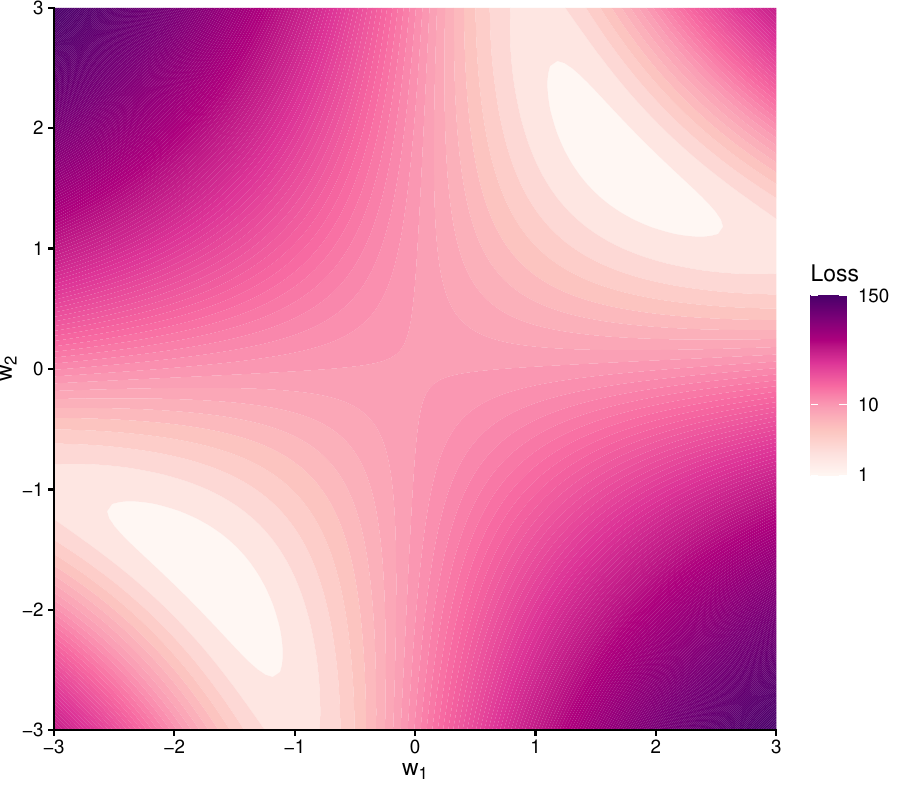}
    \caption{$\eta = 0.5$}
  \end{subfigure}}
  \hfill
  \scalebox{1}{\begin{subfigure}{0.495\textwidth}
    \centering
    \includegraphics[width=\textwidth]{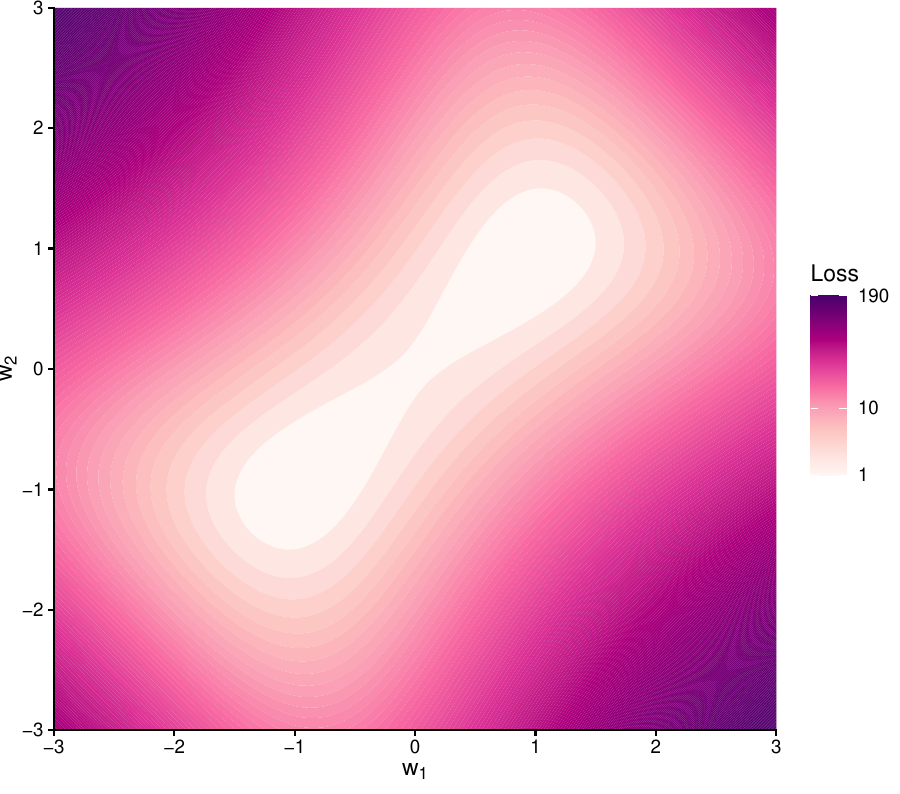}
    \caption{$\eta = 1.5$}
  \end{subfigure}
  \hfill
  \begin{subfigure}{0.495\textwidth}
    \centering
    \includegraphics[width=\textwidth]{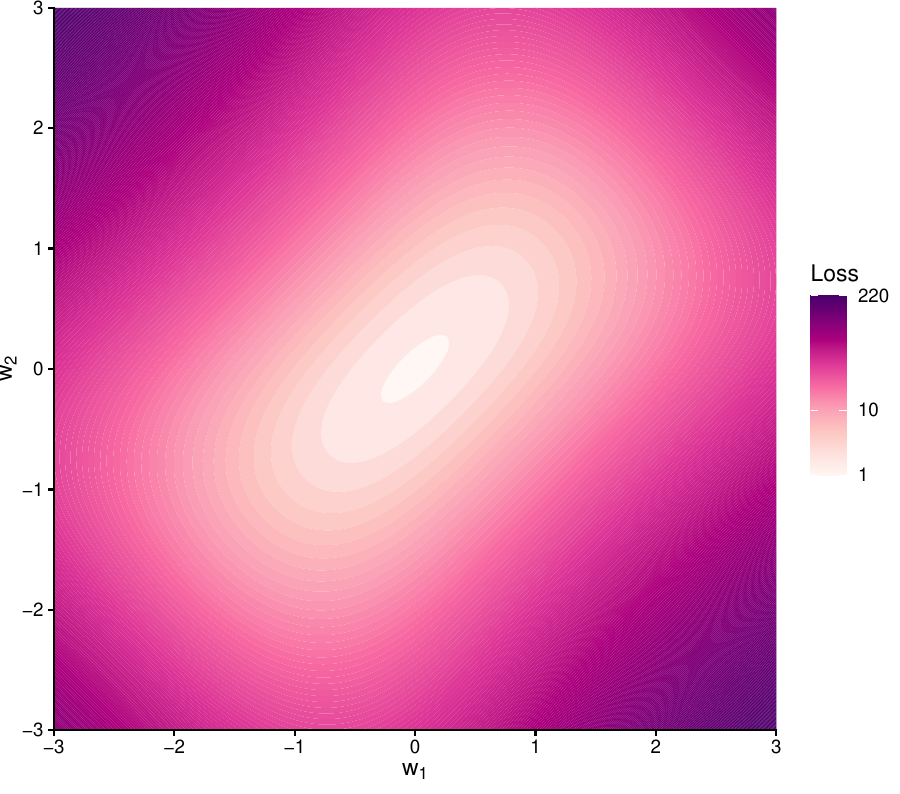}
    \caption{$\eta = 2$}
  \end{subfigure}}
  \caption{Contour plots of the loss $(\starw - w_2 w_1)^2 + \eta^2 (w_1^2 +
  w_2^2)$ with $\starw = 3.14159$ and different values of $\eta$. Large values
  of $\eta$ cause increasing shrinkage of the critical points towards zero.}
  \label{fig::cont_loss}
\end{figure}

We begin our analysis of S-SAM by characterizing the marginalized loss $\calL_R
= \calL + R$, as computed in Lemma \ref{lem::grad_exp_cond}. Since $R(\bftheta)
= S_{\avg}(\bftheta)$, we are particularly interested in how the addition of $R$
changes the overall landscape and its stationary points to control the average
sharpness. To illustrate the results we are aiming for, we briefly discuss
estimation of a single unknown $\starw \in \R$ by learning a factorization $w_2
w_1$. Up to additive and multiplicative constants, the expressions computed in
Lemma \ref{lem::grad_exp_cond} now reduce to $\calL(w_1, w_2) = (\starw - w_2
w_1)^2$ and $R(w_1, w_2) = \eta^2 \cdot (w_1^2 + w_2^2)$. Both the unregularized
loss $\calL$ and regularized loss $\calL_R$ are depicted in Figure
\ref{fig::cont_loss} for various values of $\eta$. The level curves of $\calL$
form hyperbolas, with the manifold of global minima $(\starw - w_2 w_1)^2 = 0$
emanating from the vertices $(\pm \sqrt{\starw}, \pm \sqrt{\starw})$ for $\starw
> 0$. In particular, if $w_2 w_1 = \starw$, then $(\gamma\inv \cdot w_2) (\gamma
\cdot w_1) = \starw$ for any non-zero scalar $\gamma$, meaning $\calL$ admits
global minima of arbitrary norm. Landscape and dynamics of higher-dimensional
versions of this unregularized loss are active research topics, see
\cite{achour_malgouyres_et_al_2024, arora_cohen_et_al_2018,
arora_cohen_et_al_2019a, bah_rauhut_et_al_2021, chitour_liao_et_al_2023,
chou_gieshoff_et_al_2024, kawaguchi_2016, nguegnang_rauhut_et_al_2024} for
further discussions and results.

For a well-chosen regularization parameter $\eta$, the addition of $R$ replaces
the manifold of minimizers with two single points, lying close to the vertices
of the hyperbola $(\starw - w_2 w_1)^2 = 0$. Intuitively, this is due to the
quadratic penalty ``locally coercifying'' the loss. Along each level set of
$\calL$, the quadratic $R$ is minimized when $w_2^2 = w_1^2$. Consequently, the
global minimizers of $\calL_R$ lie close to the intersection between $\argmin
\calL$ and the constraint set $w_2^2 - w_1^2 = 0$, subject to some additional
shrinkage depending on the regularization strength $\eta^2$. In the next
theorem, we prove equivalents of these properties that generalize from the
one-dimensional setting.

\begin{theorem}
  \label{thm::crit_set}
  The regularized loss $\calL_R$ in Lemma \ref{lem::grad_exp_cond} satisfies the
  following: \begin{enumerate}
    \item The critical set of $\calL_R$ is contained in the hyper-surface
      defined by the system of polynomial equations \begin{align*}
        \nabla_{\weight_{\ell + 1}} R(\bftheta) \odot \weight_{\ell + 1} =
        \nabla_{\weight_\ell} R(\bftheta) \odot \weight_\ell, \qquad \ell = 1,
        \ldots, L - 1.
      \end{align*} In particular, this entails $\weight_{\ell, j}^2 =
      \weight_{m, j}^2$ for all $\ell \neq m$ and each $j = 1, \ldots, d$ such
      that $\Diag(X\tran X)_{jj} > 0$. \label{thm::crit_set_A}
      \item Write $X_S = S X$ for a diagonal $(d \times d)$-matrix $S$ with
        $S_{ii} \in \{0, 1\}$ and define the reduced, regularized linear
        regression loss \begin{align*}
          \calL_S(\weight) = \norm[\big]{\bfY - X_S \weight}_2^2 + \bfone\tran
          \Diag\big(X_S\tran X_S\big) \bigg(\Big(\abs{\bfw}^{2 / L} + \eta^2
          \cdot \bfone\Big)^L - \abs{\bfw}^2\bigg).
        \end{align*} Suppose $\Diag(X\tran X)_{jj} > 0$ for every $j = 1,
        \ldots, d$, then the critical points of $\calL_R$ and $\calL_S$
        correspond in the following manner: If $\nabla \calL_R(\bftheta) =
        \bfzero$, then $\nabla \calL_S\big(\weight_L \odot \cdots \odot
        \weight_1\big) = \bfzero$ with $S_{ii} = 1$ for $\weight_{L, i} \cdots
        \weight_{1, i} \neq 0$. Conversely, for arbitrary $S$ and each critical
        point $\weight$ of $\calL_S$, there exists a critical point
        $(\bfw_1,\ldots,\bfw_L)$ of $\calL_R$ such that $\weight = \weight_L
        \odot \cdots \odot \weight_1$.
        \label{thm::crit_set_B}
      \item In the setting of the Part (b), suppose that $Y = X_S \starweight$
        with $X_S$ having orthonormal columns and $S_{ii} = 1$ if, and only if,
        $\weight_{*, i} \neq 0$. For $L = 2$, each $\weight_i$ admits the
        non-zero solution $\weight_i = \mathrm{sign}(\weight_{*, i}) \cdot
        \big(\abs{\weight_{*, i}} - \eta^2\big)$ whenever $\abs{\weight_{*, i}}
        > \eta^2$, otherwise $\weight_i = 0$. \label{thm::crit_set_C} For $L >
        2$, each $\weight_i$ admits at most two distinct solutions
        \begin{align*}
          \left(\dfrac{\eta^2}{\abs{\weight_{*, i}}^{2 / L}}\right)^{L (L - 1) /
          (L - 2)} \cdot \weight_{*, i} \leq \weight_i \leq \sqrt{1 -
          \dfrac{\eta^2}{\abs{\weight_{*, i}}^{2 / L}}}^L \cdot \weight_{*, i}
        \end{align*} whenever \begin{align*}
          \abs{\weight_{*, i}} \geq \left(1 + \dfrac{L}{L - 2}\right)^{L - 1}
          \left(\dfrac{L - 2}{L}\right)^{L / 2} \cdot \eta^L
        \end{align*} with $\weight_i = 0$ otherwise.
  \end{enumerate}
\end{theorem}

Theorem \ref{thm::crit_set_A} generalizes the constraint set $w_2^2 - w_1^2 = 0$
of the one-dimensional problem. In fact, the first part of Theorem
\ref{thm::crit_set_A} holds for any differentiable regularizer $R$ and a general
linear network, in which case the critical points are constrained to lie within
the hyper-surface $W_{\ell + 1}\tran \nabla_{W_{\ell + 1}} R(\bftheta) =
\nabla_{W_\ell} R(\bftheta) W_\ell\tran$. For non-singular $\Diag(X\tran X)$,
the resulting constraint $\weight_{\ell}^2 = \weight_{\ell + 1}^2$ gives an
instance of the \textit{balancing condition}, which reads $W_{\ell + 1}\tran
W_{\ell + 1} = W_{\ell} W_{\ell}\tran$ for general matrices and commonly appears
in the analysis of deep linear networks \citep{arora_cohen_et_al_2018,
arora_cohen_et_al_2019a, bah_rauhut_et_al_2021}. In Section \ref{sec::full_lnn},
we will show that applying S-SAM to linear neural networks yields a generalized
version of the balancing constraint. For the diagonal linear model, we conclude
that the average sharpness penalty forces any optimization algorithm yielding
stationary points to eventually become attracted to the set of balanced
parameters.

If $\Diag(X\tran X)_{jj} = 0$, then the $j$\textsuperscript{th} row of $X$ must
also equal zero. Ruling out this case, as in Theorem \ref{thm::crit_set_B} and
\ref{thm::crit_set_C}, equates to the removal of all features $j$ that are
uninformative for the linear regression model \eqref{eq::lin_reg_def}.
Pre-processing steps of this kind are common in regularized linear regression,
alongside centering and standardization of the data variables
\citep{zou_hastie_2005}.

Expanding the power in Theorem \ref{thm::crit_set_B}, the effective
regularization term acts as a composite penalty of various fractional norms of
$\weight$, weighted by the column norms of $X_S$ and with differing
regularization strengths. In particular, it obeys the general form
\begin{align*}
  \weight \mapsto \sum_{j = 1}^J \lambda_j \cdot \norm*{\sqrt{\Diag\big(X_S\tran
  X_S\big)} \weight}_{q_j}^{q_j}
\end{align*} with $\lambda_j > 0$, and $q_j < 2$. The exponents of the
fractional norms range from $2 / L$ to $2 - 1 / L$, meaning the regularizer
interpolates between various concave and convex penalties, similar to the
elastic net penalty interpolating between $\ell_1$ and $\ell_2$-regularization
\citep{zou_hastie_2005}. The concave norms feature infinite gradient at the
origin, causing hard thresholding of small parameters
\citep{mazumder_friedman_et_al_2011}. Unlike the well-known implicit bias
towards $\ell_1$-regularized solutions of SGD on the $2$-layer diagonal model
with small initialization \citep{pesme_pillaud-vivien_et_al_2021,
even_pesme_2023}, S-SAM fundamentally alters the loss landscape by directly
inducing a bias towards sparse solutions through the regularizer. Consequently,
the S-SAM algorithm \eqref{eq::sgd_sam_def} performs feature learning regardless
of the initialization scale, as already seen in Figure
\ref{fig::depth_3_setting}. The simultaneous presence of concave and convex
penalties for $L > 2$ also distinguishes the induced regularizer from standard
weight decay, which leads to a $2 / L$ fractional norm penalty when applied to
the diagonal linear model \citep{hoff_2017}.

Together, Theorem \ref{thm::crit_set_A} and \ref{thm::crit_set_B} illustrate how
the regularizer $R$ controls the parameter norm near stationary points of
$\calL_R$ for orthogonal data on a subset of the coordinates. As in the
one-dimensional case, $\gamma_\ell \cdot \weight_\ell$ with non-zero scalars
$\gamma_1 \cdots \gamma_L = 1$ also factorizes $\weight = \weight_L \odot \cdots
\odot \weight_1$. Taking $\gamma_1 \to \infty$, while sending the other scalars
to zero at a matching rate, then gives a factorization of $\weight$ with
arbitrarily large parameter norm. Forcing the weight matrices to be
approximately balanced near the stationary set rules out this escape to
infinity; the fit on the observed data and the norm of the linear predictor have
to lie in balance due to the added penalty, much in the same way as the squared
norm penalty causes shrinkage in ridge regression \citep[Chapter 3.4.1
of][]{hastie_tibshirani_et_al_2009}. A quantitative relation between parameter
norm and balancing is given in Proposition 3.2 of
\cite{nguegnang_rauhut_et_al_2024}.

Theorem \ref{thm::crit_set_C} describes the critical points of $\calL_R$ as a
\textit{shrinkage-thresholding} operator applied to the linear interpolator
$\starweight$ on a subset of orthogonal data. Such operators also appear in the
study of other algorithmic regularization techniques, for example in dropout
\citep{senen-cerda_sanders_2022}. The threshold in Theorem \ref{thm::crit_set_C}
depends both on the noise level $\eta$ and the number of layers $L$. If the
underlying signal $\weight_{*, i}$ is too weak in comparison with $\eta^L$, any
critical point will be forced towards $\weight_{\ell, i} = 0$. Hence, the
regularizer $R$ induces an explicit bias towards sparse linear predictors
$\weight$, with larger $\eta$ corresponding to a smaller set of active
coordinates by truncating the smaller entries of $\starweight$. As this bias
follows entirely from landscape regularization, it persists for any training
algorithm that returns an approximate critical point of $\calL_R$, which is
consistent with the empirical observation that SAM induces low-rank features
\citep{andriushchenko_bahri_et_al_2023}. This mechanism differs from the
implicit bias towards sparse solutions for the unregularized mirror flow
dynamics, which only holds for specific initializations and under regularity
conditions \citep{woodworth_gunasekar_et_al_2020}. The shrinkage also describes
the ``coercification'' of $\calL$ through $R$, as regulated by $\eta$, see the
change in landscape in Figure \ref{fig::cont_loss} for a depiction in the
one-dimensional model. As $\eta$ increases, the non-zero critical points are
gradually forced towards $0$, eventually merging into a single critical point at
$0$, once $\abs{\weight_{*, i}}$ lies below the threshold. When $L \to \infty$,
the threshold approximates $2^{L - 1} \cdot \eta^L$, asymptotically requiring
$\eta = O(2^{- 1 + 1 / L})$ to allow for a non-zero solution. The bounds on the
non-zero roots in Theorem \ref{thm::crit_set_C} may be refined to arbitrary
precision, see Appendix \ref{sec::approx_root}.

We conclude our study of the regularized landscape $\calL_R$ by directly
relating the balancing constraint in Theorem \ref{thm::crit_set_A} to the
average sharpness $S_{\avg}$ and the trace of the Hessian matrix, proving that
the balancing explicitly controls both of the latter.

\begin{lemma}
  \label{lem::bal_reg}
  Fix a linear predictor $\weight$ and define $\pi(\weight) = \{\bftheta \mid
  \weight = \weight_L \odot \cdots \odot \weight_1\}$, then the maps $\bftheta
  \mapsto S_\avg(\bftheta)$ and $\bftheta \mapsto \Tr\big(\nabla^2
  \calL(\bftheta)\big)$ both attain their minima over $\pi(\weight)$ if, and
  only if, $\weight_{\ell, i}^2 = \weight^2_{\ell + 1, i}$ for every $\ell = 1,
  \ldots, L - 1$ and every $i = 1, \ldots, d$ satisfying $\Diag(X\tran X)_{ii} >
  0$.
\end{lemma}

As will be shown in Lemma \ref{lem::pac_bound} for fully connected linear
networks, the average sharpness and parameter norm together act as a
generalization measure for linear neural networks training on the squared
Euclidean loss $\calL$. Combining Theorem \ref{thm::crit_set} and Lemma
\ref{lem::bal_reg} proves that in the diagonal linear model the balancing
constraint effectively controls both relevant quantities. Further, this
constraint decreases the model complexity by reducing the effective number of
parameters.

\section{Convergence of the S-SAM Recursion}
\label{sec::conv_grad_desc}

Having characterized the change in landscape induced by the average sharpness
regularizer, we now prove that the S-SAM recursion \eqref{eq::sgd_sam_def}
indeed produces a critical point of the expected loss $\calL_R$. Our analysis is
inspired by techniques found in \cite{bah_rauhut_et_al_2021},
\cite{chatterjee_2022}, and \cite{nguegnang_rauhut_et_al_2024}. Crucially, the
presence of the regularizer $R$ allows us to simplify certain arguments relating
to the boundedness of the trajectories.

\subsection{Gradient Flow Analysis}
\label{sec::grad_flow_anal}

For each $\ell = 1, \ldots, L$, the weight vectors $\bftheta_t =
\big(\weight_1(t), \ldots, \weight_L(t)\big)$, $t \geq 0$, are said to follow
the gradient flow of $\calL_R$ if they solve the system of ordinary differential
equations \eqref{eq::sgd_sam_gf} with given boundary conditions $\bftheta_0$,
meaning they evolve along the negative of the gradient vector field of
$\calL_R$.  Existence and uniqueness of the gradient flow trajectories follows
from standard theorems on ordinary differential equations, see for example
Chapter 2 of \cite{teschl_2012}. The continuous time index $t \in [0, \infty)$
may be interpreted as the limit of vanishing step-sizes in the corresponding
gradient descent recursion generated by $\calL_R$. 

As a consequence of the definition, the loss $\calL_R$ may never increase in $t$
when the weight vectors follow the gradient flow of $\calL_R$. Indeed, since the
time derivative of each $\weight_{\ell}(t)$, $\ell = 1, \ldots, L$ equals the
respective negative gradient of $\calL_R$, the chain rule implies \begin{align}
  \label{eq::grad_flow_monot}
  \dfrac{\rmd}{\rmd t} \calL_R(\bftheta_t) = \sum_{\ell = 1}^L
  \iprod*{\nabla_{\weight_\ell} \calL_R(\bftheta_t), \dfrac{\rmd}{\rmd t}
  \weight_\ell(t)} = - \sum_{\ell = 1}^L \norm[\big]{\nabla_{\weight_\ell}
  \calL_R(\bftheta_t)}^2_2,
\end{align} so the derivative of $\calL_R$ stays non-positive along the whole
trajectory. For the particular loss $\calL_R$, this already implies convergence
to a critical point as $t \to \infty$, as we will now show.

{\L}ojasiewicz's Theorem on gradient flows \citep{lojasiewicz_1984}, restated in
Appendix \ref{sec::gd_conv}, shows that the gradient flow of any analytic
function either diverges to infinity, or yields a critical point of said
function. If the trajectories of the $\weight_\ell(t)$ in \eqref{eq::sgd_sam_gf}
stay bounded, they must then necessarily converge to a critical point of
$\calL_R$ as $t \to \infty$. Ensuring boundedness of the trajectories can be difficult, see for example the proof of Theorem
3.2 in \cite{bah_rauhut_et_al_2021}, but we can profitably employ the
regularizer $R$.

\begin{lemma}
  \label{lem::gf_coerc}
  If $\min_{j = 1, \ldots, d} \Diag(X\tran X)_{jj} > 0$, then, for any $t\geq 0,$
  \begin{align*}
    \sum_{\ell = 1}^L \norm{\weight_\ell(t)}^2_2 \leq \dfrac{1}{\eta^{2 L - 2}
    \sigminp(X\tran X)} \cdot \calL_R(\bftheta_0).
  \end{align*}
\end{lemma}

The proof relies on coercivity of the regularized
loss $\calL_R$ through \eqref{eq::reg_wd}. The left-hand side in Lemma
\ref{lem::gf_coerc} denotes the squared Euclidean norm of the network parameter
$\bftheta_t$ as a vector in $\R^{Ld}$, and in turn Theorem \ref{thm::loj_cont}
ensures the existence of $\weight_\ell(\infty) = \lim_{t \to \infty}
\weight_\ell(t)$ for every $\ell$, which must be a critical point of $\calL_R$.
The effect of the data geometry on the upper-bound through $\sigminp(X\tran X)$
may be removed by standardizing the predictors to have $\Diag(X\tran X) = I_d$,
which also holds for any of the following results that reference the singular
values of $X\tran X$.

Further, since $\calL_R$ is locally Lipschitz continuous, we can integrate and
rearrange \eqref{eq::grad_flow_monot} to show that the average norm of the
gradient of $\calL_R$ along the gradient flow trajectory vanishes at a linear
rate in time \begin{equation}
  \label{eq::grad_flow_conv_avg}
  \dfrac{1}{t} \cdot \int_0^t \sum_{\ell = 1}^L
  \norm[\big]{\nabla_{\weight_\ell} \calL_R(\bftheta_s)}^2_2\ \rmd s =
  \dfrac{1}{t} \cdot \Big(\calL_R(\bftheta_0) - \calL_R(\bftheta_t)\Big) \leq
  \dfrac{1}{t} \cdot \calL_R(\bftheta_0).
\end{equation}

The arguments employed so far work for any analytic loss function that is
coercive, in the sense that it upper-bounds a suitable transformation of the
parameter norm. In our case, this transformation happens to be the quadratic
monomial \eqref{eq::reg_wd}.

Exploiting the specific structure of the loss $\calL_R$, we are moreover able to
give a quantitative version of the constraint in Theorem \ref{thm::crit_set_A}
along the trajectory of the gradient flow.

\begin{theorem}
  \label{thm::grad_flow_bal}
  Under the same assumptions as in Lemma \ref{lem::gf_coerc}, for every $\ell =
  1, \ldots, L$ and any $t \geq 0$, \begin{align*}
    \norm[\big]{\weight_\ell^2(t) - \weight_{\ell + 1}^2(t)}_2 \leq \exp\Big(- 4
    \eta^{2 L - 2} \sigminp\big(X\tran X\big) \cdot t\Big) \cdot
    \norm[\big]{\weight_\ell^2(0) - \weight_{\ell + 1}^2(0)}_2.
  \end{align*}
\end{theorem}

As mentioned in the discussion following Theorem \ref{thm::crit_set}, the
balancing constraint $\weight_\ell^2 - \weight_{\ell + 1}^2$ plays an
instrumental role in controlling the parameter norm at the critical points of
$\calL_R$ and induces the fractional norm penalty. Based on Theorem
\ref{thm::grad_flow_bal}, we conclude that the regularizer $R$ not only provides
this control asymptotically by changing the landscape at the stationary points,
but also forces the gradient flow of $\calL_R$ to rapidly balance the weight
matrices. This stands in stark contrast with the gradient flow of the
unregularized objective $\calL$, for which the balancing remains unchanged along
its trajectories \citep{arora_cohen_et_al_2018, bah_rauhut_et_al_2021}.
Combining Lemma \ref{lem::bal_reg} with Theorem \ref{thm::grad_flow_bal} also
proves that the gradient flow minimizes the average sharpness $S_{\avg}$ and the
trace of the Hessian matrix at an exponential rate along its trajectory.

\subsection{Generalizing to Gradient Descent}

The convergence result for gradient flows presented in Section
\ref{sec::grad_flow_anal} exploits the basic monotonicity property
\eqref{eq::grad_flow_monot}, which does not directly extend
to discrete time. As illustrated in
\cite{chatterjee_2022}, or the extension of Theorem 2.3 of
\cite{bah_rauhut_et_al_2021} on gradient flows to gradient descent in Theorem
2.4  of \cite{nguegnang_rauhut_et_al_2024}, the discretization must be chosen
carefully.

For a sequence of positive step-sizes $\alpha_k$ and initial values $\bftheta_0
= \big(\weight_1(0), \ldots, \weight_L(0)\big)$, the discrete-time analog of
\eqref{eq::sgd_sam_gf} is given by the gradient descent recursions \begin{align}
  \label{eq::gd_def_no_noise}
  \weight_\ell(k + 1) = \weight_\ell(k) - \alpha_k \cdot
  \nabla_{\weight_\ell(k)} \calL_R(\bftheta_k).
\end{align} We aim to apply the generic {\L}ojasiewicz-type theorem for
discrete-time iterations shown in \cite{absil_mahoney_et_al_2005}. See Theorem
\ref{thm::loj_disc} for a restatement and subsequently Lemma \ref{lem::gd_loj}
for a version adapted to gradient descent. Analogous to the result for gradient
flows, the theorem states that the gradient descent iterates of analytic
functions either diverge to infinity or converge to a critical point of the
function, given regularity conditions. The sole requirement for this result is
the so-called \textit{strong descent condition}. In our notation, the condition
may be stated as the existence of $\delta > 0$, such that \begin{equation}
  \label{eq::str_desc_main}
  \calL_R(\bftheta_k) - \calL_R(\bftheta_{k + 1}) \geq \delta \alpha_k \cdot
  \sum_{\ell = 1}^L \norm[\big]{\nabla_{\weight_\ell(k)} \calL_R(\bftheta_k)}^2
\end{equation} for every $k \geq 0$. This ensures a monotonically decreasing
loss along the trajectory of the gradient descent \eqref{eq::gd_def_no_noise},
giving a discrete-time equivalent of \eqref{eq::grad_flow_monot}. The condition
that $\delta$ can be chosen uniformly for all $k$ gives a quantitative version
of the classical descent lemma. For loss functions with globally Lipschitz
continuous gradients, it suffices that $\alpha_k$ never exceeds twice the
reciprocal of the Lipschitz constant \citep{garrigos_gower_2024}. When $L > 1$,
the loss $\calL_R$ is polynomial in the weights $\weight_{\ell, ii}$ with
leading terms of the form $\weight_{L, ii}^2 \cdots \weight_{1, ii}^2$.
Consequently, $\nabla \calL_R$ only admits local Lipschitz continuity and
verifying \eqref{eq::str_desc_main} along the gradient descent trajectory
requires a careful account of local Lipschitz constants. If the strong descent
condition can be proven for every $k$, then the bound in Lemma
\ref{lem::gf_coerc} holds with the continuous-time index $t$ replaced by $k$,
implying boundedness of the whole trajectory. This is the main idea behind the
next theorem.

\begin{theorem}
  \label{thm::conv_abs_disc}
  Suppose $\min_{j = 1, \ldots, d} \Diag(X\tran X)_{jj} > 0$ and fix $\delta \in
  (0, 1)$, then \eqref{eq::str_desc_main} holds with
  the chosen $\delta$ whenever \begin{align*}
    \alpha_k < 2 (1 - \delta) \cdot \left(\sqrt{L} \kappa\big(X\tran X\big)
    \cdot \dfrac{7 \sqrt{L} + 2}{\eta^2} \cdot \calL_R(\bftheta_k)\right)\inv.
  \end{align*} If the above condition is satisfied for all $k \geq K$ and some
  $K \geq 0$, then \begin{align*}
    \sup_{k \geq K} \, \sum_{\ell = 1}^L \norm{\weight_\ell(k)}^2_2 \leq
    \dfrac{1}{\eta^{2 L - 2} \sigminp(X\tran X)} \cdot \calL_R(\bftheta_K)
  \end{align*} In this case, the gradient descent iterates converge to a
  critical point of $\calL_R$ as $k \to \infty$, provided that $\sum_{k =
  K}^\infty \alpha_k = \infty$.
\end{theorem}

The difficulty in proving Theorem \ref{thm::conv_abs_disc} lies in
finding an efficient estimate for the largest singular value of $\nabla^2
\calL_R$. As shown in Lemma \ref{lem::grad_exp_cond}, the regularizer $R$ is
defined in terms of a product over vectors $\weight_\ell^2 + \eta^2 \cdot
\bfone$. Expanding this product and individually bounding the second derivatives
of each term would yield a bound for $\nabla^2 \calL_R$ of the order $O(2^L
\cdot \calL_R^L)$. Since the maximal step-size must eventually lie below twice
the reciprocal of the estimated operator norm, this would lead to slow
convergence for even moderately large $L$ and bad initializations. By exploiting
the specific form \eqref{eq::reg_poly} of the regularizer $R$, one can improve
this estimate to $O(L \cdot \calL_R)$, as shown in the assumption on $\alpha_k$
in Theorem \ref{thm::conv_abs_disc}. In comparison, the corresponding largest
singular value estimate for the unregularized descent scales like $O(L^{2 L}
\cdot \calL_R^2)$, unless the initial matrices are chosen close to being
balanced. See Theorem 2.4 and Remark 2.5 in \cite{nguegnang_rauhut_et_al_2024}
for further details.

In practice, training algorithms often feature distinct learning rate phases,
such as the popular warm-up/stable/decay (WSD) schedule. The stable phase
typically targets a constant learning rate exceeding the stability threshold,
thereby causing the loss to fluctuate wildly until the decay phase returns the
algorithm to stable descent \citep{wen_li_et_al_2025}. The stability threshold
for $\alpha_k$ in Theorem \ref{thm::conv_abs_disc} gives a sufficient condition
in terms of the loss $\calL_R(\bftheta_k)$ for stable descent to occur. 

For constant step-sizes $\alpha_k = \alpha$ satisfying the requirements of
Theorem \ref{thm::conv_abs_disc} after some $K \geq 0$, summing over
\eqref{eq::str_desc_main} yields the discrete-time equivalent of the convergence
rate \eqref{eq::grad_flow_conv_avg}, namely \begin{align*}
  \dfrac{\delta \alpha}{k - K} \cdot \sum_{j = K}^{k - 1} \sum_{\ell = 1}^L
  \norm[\big]{\nabla_{\weight_\ell(j)} \calL_R(\bftheta_j)}^2_2 \leq \dfrac{1}{k
  - K} \cdot \calL_R(\bftheta_k).
\end{align*}

To conclude our analysis of \eqref{eq::gd_def_no_noise}, we prove a
discrete-time version of Theorem \ref{thm::grad_flow_bal}. Consequently, the
addition of the regularizer $R$ changes the gradient descent dynamics in a
manner similar to the gradient flow, forcing balancing of the weight matrices.
As is the case with Theorem \ref{thm::grad_flow_bal}, this yields control over
both the sharpness and parameter norm near critical points of $\calL_R$.

\begin{theorem}
  \label{thm::grad_desc_bal}
  Let $\Diag(X\tran X)_{jj} > 0$ for every $j = 1, \ldots, d$ and suppose
  \begin{align*}
    \alpha_k < \min \left\{\dfrac{1}{\eta^{2 L - 2} \sigminp(X\tran X)},\
    \dfrac{\eta^2}{4 \kappa(X\tran X) \cdot \calL_R(\bftheta_k)},\ \dfrac{3
    \eta^{2 L + 2}}{16 \kappa(X\tran X)^2 \cdot \calL_R(\bftheta_k)^2}\right\},
  \end{align*} then 
  \begin{align*}
    \norm[\big]{\weight_\ell^2(k + 1) - \weight_{\ell + 1}^2(k + 1)}_2 \leq
    \Big(1 - \alpha_k \eta^{2 L - 2} \sigminp(X\tran X)\Big) \cdot
    \norm[\big]{\weight_\ell^2(k) - \weight_{\ell + 1}^2(k)}_2.
  \end{align*} for each $\ell = 1, \ldots, L - 1$.
\end{theorem}

\begin{figure}[t]
  \centering
  \scalebox{1}{\begin{subfigure}{0.495\textwidth}
    \centering
    \includegraphics[width=\textwidth]{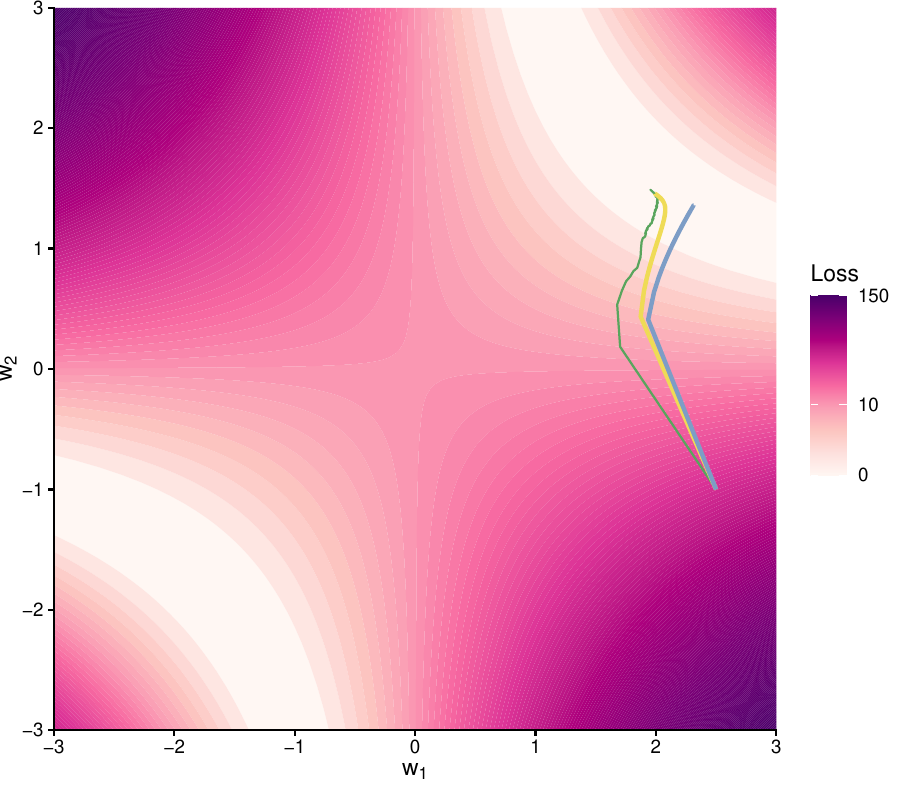}
    \caption{$\eta = 0.5$ and $\alpha_k = 0.1 / (k + 1)$}
  \end{subfigure}
  \hfill
  \begin{subfigure}{0.495\textwidth}
    \centering
    \includegraphics[width=\textwidth]{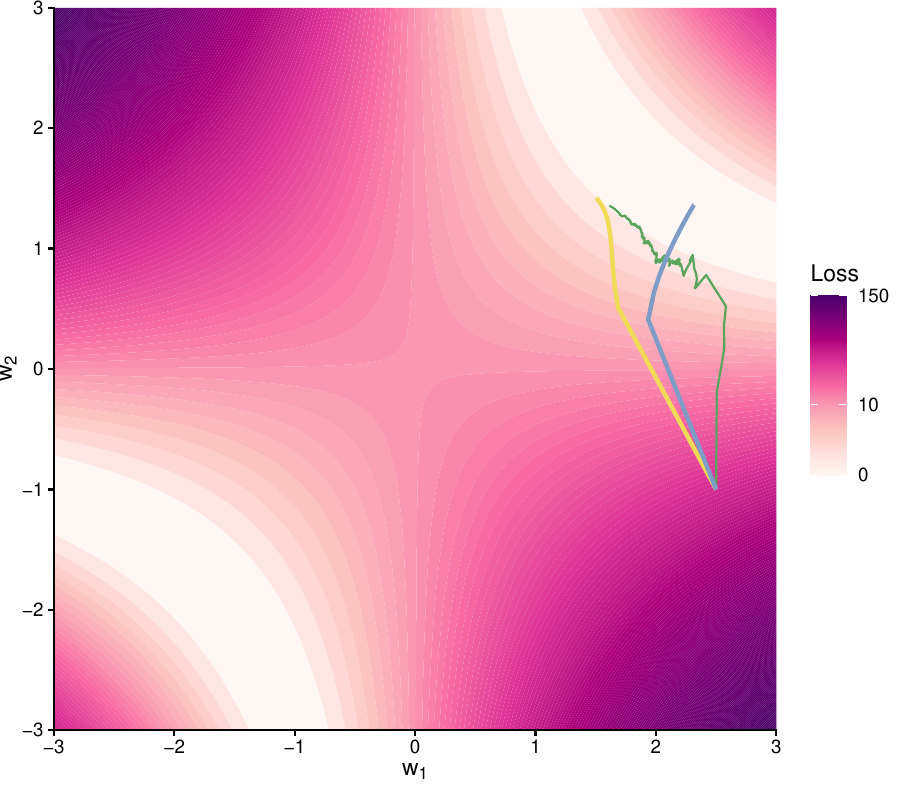}
    \caption{$\eta = 1$ and $\alpha_k = 0.1 / (k + 1)$}
  \end{subfigure}}
  \hfill
  \scalebox{1}{\begin{subfigure}{0.495\textwidth}
    \centering
    \includegraphics[width=\textwidth]{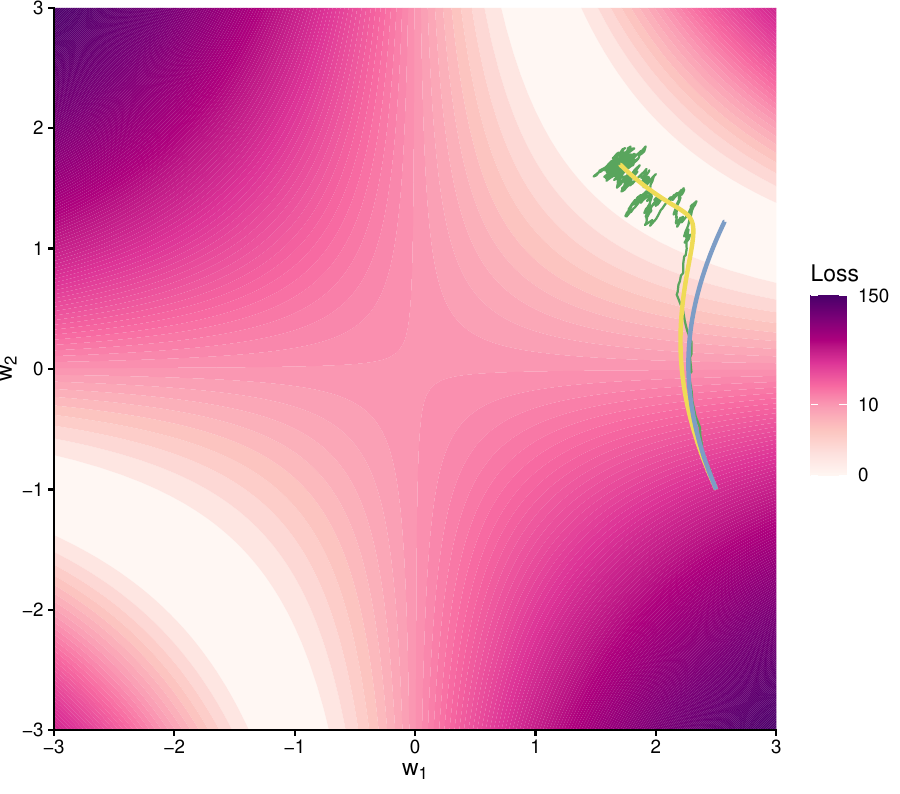}
    \caption{$\eta = 0.5$ and $\alpha_k = 0.01$}
  \end{subfigure}
  \hfill
  \begin{subfigure}{0.495\textwidth}
    \centering
    \includegraphics[width=\textwidth]{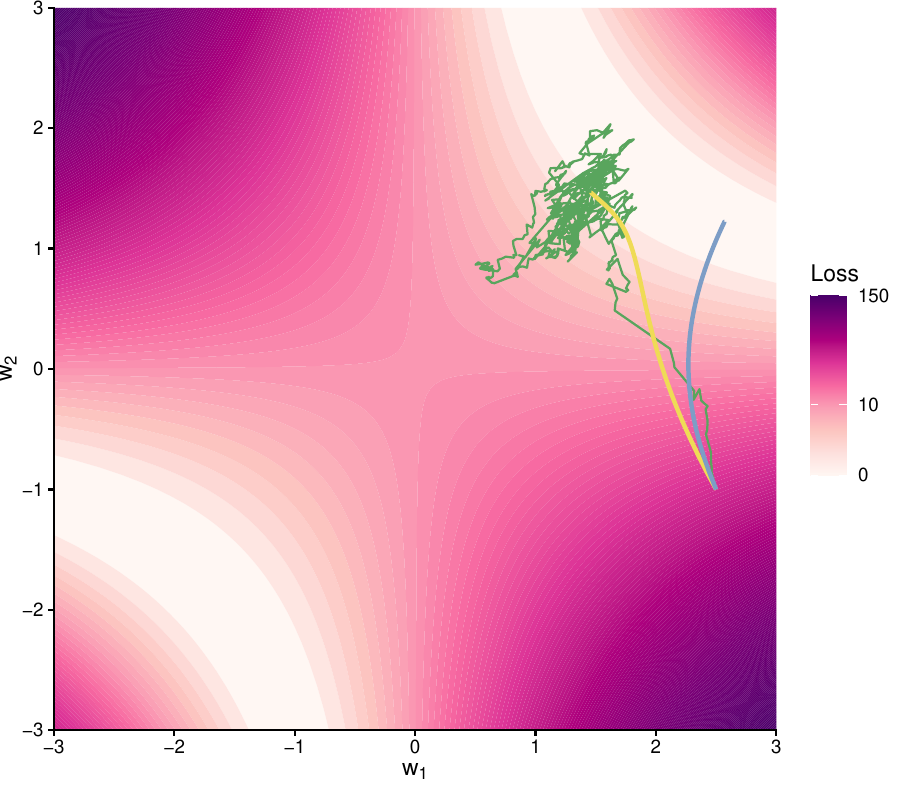}
    \caption{$\eta = 1$ and $\alpha_k = 0.01$}
  \end{subfigure}}
  \caption{Gradient descent trajectories overlaid onto a contour plot of the
  loss $(\starw - w_2 w_1)^2$ with $\starw = 3.14159$, started from the same
  initial point and run with different values of $\eta$ and $\alpha_k$. Shown
  are standard gradient descent following $- \nabla \calL$ (blue), explicitly
  regularized gradient descent following $- \nabla \calL_R$ (yellow), and
  gradient descent with S-SAM following $- \nabla \wtL_k$ (green).}
  \label{fig::gd_comp}
\end{figure}

For an illustration, see Figure
\ref{fig::gd_comp}. Once the unregularized gradient descent
\eqref{eq::sgd_reg_def} reaches a critical point of $\calL$, the algorithm
naturally terminates. In contrast, the regularizer $R$ continually forces the
iterates \eqref{eq::gd_def_no_noise} towards a balanced solution, once
$\calL_R(\bftheta_k)$ lies below some threshold in relation to $\alpha_k$, with
larger values of $\eta$ corresponding to faster balancing. The asymptotic speed
of convergence in Theorem \ref{thm::grad_desc_bal} further depends on the
step-sizes $\alpha_k$, with a constant step-size $\alpha_k = \alpha$ giving a
geometric rate. Choosing $\alpha_k = \alpha / (k + 1)$ for some suitable
constant $\alpha > 0$ leads to balancing at the rate $k^{- \alpha \eta^{2 L - 2}
\sigminp(X\tran X)}$. Indeed, since $\log(1 - x) \leq - x$ for $x \in (0, 1)$,
\begin{align*}
  \prod_{j = 0}^{k - 1} \big(1 - \alpha_j \eta^{2 L - 2} \sigminp(X\tran X)\big)
  &= \exp\left(\sum_{j = 0}^{k - 1} \log\Big(1 - \alpha_j \eta^{2 L - 2}
  \sigminp(X\tran X)\Big)\right)\\
  &\leq \exp\left(- \alpha \eta^{2 L - 2} \sigminp(X\tran X) \cdot \sum_{j =
  1}^k \dfrac{1}{j}\right)
\end{align*} and the rate now follows from $\sum_{j = 1}^k j\inv \approx
\log(k)$.

\subsection{Convergence of the Projected Stochastic Recursion}
\label{sec::conv_stoch}

On the event that the iterates stay bounded, almost sure convergence towards a
root of the expected gradient can be ensured for a large class of functions
\citep{davis_drusvyatskiy_et_al_2020, dereich_kassing_2024}, provided the
step-sizes $\alpha_k$ adequately control the variance of the gradient noise. To
ensure boundedness, we will consider a version of the S-SAM recursion
\eqref{eq::sgd_sam_def} projected onto a compact set.

Recall from Section \ref{sec::lin_nn_back} that we are given an i.i.d.\ sample
$(\bfX_i, Y_i)$, $i = 1, \ldots, n$. Suppose $(\bfX, Y) \sim
\mathrm{Unif}\big((\bfX_1, Y_1), \ldots, (\bfX_n, Y_n)\big)$ and $\bfxi \sim
\calN(0, \eta^2 \cdot I_{Ld})$, then the S-SAM recursion \eqref{eq::sgd_sam_def}
moves during every iteration along the negative gradient of an independent
sample from the random function \begin{align*}
  \wtL(\bftheta) = \big(Y - \bfX\tran \wtweight\big)^2,
\end{align*} with $\wtweight = \wtweight_L \odot \cdots \odot \wtweight_1$ and
$\wtweight_{\ell, i} = \weight_{\ell, i} + \bfxi_{\ell, i}$. To take an
independent sample $\wtL_k \sim \wtL$ of this randomized loss, we may
independently sample $\bfxi_k \sim \bfxi$ and $(\bfX_k, Y_k) \sim (\bfX, Y)$ and
simply plug in the outcomes. As shown in Lemma \ref{lem::grad_exp_cond}, the
discrete-time dynamical system generated by $- \nabla \wtL_k$, with an
independent sample $\wtL_k$ taken during every time-step, may be decomposed into
$- \nabla \calL_R + \bfzeta_k$ with $\bfzeta_k = \nabla \calL_R - \nabla \wtL_k$
forming a martingale difference sequence. The random vector $\bfxi$ has
unbounded support and so $\sup_{k \geq 0} \norm{\nabla \wtL_k} = \infty$. To
ensure that the algorithm produces almost surely bounded trajectories, we focus
on the case where the weights are projected onto a compact set after every
iteration. This makes the algorithm amenable to a convergence result in
\cite{davis_drusvyatskiy_et_al_2020}. Given $r > 0$, let $\Pi_r$ denote the
projection onto the Euclidean ball of radius $r$ in $\R^{Ld}$. Applied to the
stacked weight vectors $\bftheta = (\weight_1, \ldots, \weight_L)$, this takes
the form \begin{align*}
  \Pi_r\left(\begin{bmatrix}
    \weight_1\\
    \vdots\\
    \weight_L
  \end{bmatrix}\right) = \min \left\{1,\ \dfrac{r}{\sqrt{\sum_{\ell = 1}^L
  \norm{\weight_\ell}^2}} \right\} \cdot \begin{bmatrix}
    \weight_1\\
    \vdots\\
    \weight_L
  \end{bmatrix}
\end{align*} and the projected S-SAM iterates satisfy the recursion
\begin{align}
  \label{eq::proj_sam}
  \begin{bmatrix}
    \weight_1(k + 1)\\
    \vdots\\
    \weight_L(k + 1)
  \end{bmatrix} = \Pi_r\left(\begin{bmatrix}
    \weight_1(k)\\
    \vdots\\
    \weight_L(k)
  \end{bmatrix} - \alpha_k \cdot \begin{bmatrix}
    \nabla_{\weight_1(k)} \wtL_k(\bftheta_k)\\
    \vdots\\
    \nabla_{\weight_1(k)} \wtL_k(\bftheta_k)
  \end{bmatrix}\right).
\end{align} By definition, the resulting trajectories always stay confined to
the ball of radius $r$ and we may show that every limit point of the iterates
is a critical point of $\calL_R$ for sufficiently large $r$.

\begin{theorem}
  \label{thm::proj_sam_conv}
  Suppose the step-sizes $\alpha_k$ satisfy $\sum_{j = 0}^\infty \alpha_j =
  \infty$ and $\sum_{j = 0}^\infty \alpha_j^2 < \infty$. Let $\min_{j = 1,
  \ldots, d} \Diag(X\tran X)_{jj} > 0$ and pick a radius $r$ such that
  \begin{align*}
    r > \dfrac{\sqrt{L}}{2 \eta^{L - 1} \sqrt{\sigminp(X\tran X)}} \cdot
    \norm{\bfY}_2,
  \end{align*} then every limit point of the projected S-SAM recursion
  \eqref{eq::proj_sam} is a critical point of $\calL_R$ and the function values
  $\calL_R(\bftheta_k)$ converge as $k \to \infty$.
\end{theorem}

The assumption on the step-sizes $\alpha_k$ in Theorem \ref{thm::proj_sam_conv}
is standard in the stochastic approximation literature and known as the
Robbins-Monro condition after \cite{robbins_monro_1951}. Since $- \alpha_k \cdot
\nabla \wtL_k = - \alpha_k \cdot \nabla \calL_R + \alpha_k \cdot \bfzeta_k$, the
variance of the random gradients may be controlled by $\alpha_k^2 \cdot
\E\big[\norm{\bfzeta_k}^2\big]$, so picking square summable step-sizes ensures
that the random gradients converge to their expectations in squared mean at a
sufficiently fast rate.

The condition on the radius $r$ rules out cases where the iterates
asymptotically stick to the boundary of the ball. For smaller radii, the vector
field $- \nabla \calL_R$ may pull the iterates towards a stationary point
outside the ball of radius $r$. Since the regularized loss $\calL_R$ is roughly
quadratic at large scales by \eqref{eq::reg_wd}, we may pick $r$ large enough so
that all critical points have norm strictly less than $r$ and the vector field
$- \nabla \calL_R$ always transports the iterates back into the ball at the
boundary. If $\bfY = X \starweight$, then an alternative condition on $r$ in
Theorem \ref{thm::proj_sam_conv} reads \begin{align*}
  r > \dfrac{\sqrt{L \kappa(X\tran X)}}{2 \eta^{L - 1}} \cdot
  \norm{\starweight}_2.
\end{align*}

\section{Fully Connected Linear Neural Networks}
\label{sec::full_lnn}

We now consider the application of S-SAM to generic linear networks. Suppose we
observe $n$ i.i.d.\ data pairs $(\bfX_i, \bfY_i) \in \R^{d_0} \times \R^{d_L}$
and let $\bftheta = (W_L, \ldots, W_1)$, with each $W_\ell$ a $(d_\ell \times
d_{\ell - 1})$-matrix. We aim to fit the linear model $\bfY_i \approx W \bfX_i$,
with factorized parametrization $W = W_L \cdots W_1$, via minimization of the
empirical risk \begin{align}
  \calL(\bftheta) \mapsto \dfrac{1}{n} \cdot \sum_{i = 1}^n \norm[\big]{\bfY_i -
  W \bfX_i}_2^2 = \norm[\big]{Y - W X}^2.
\end{align} Here, $X$ denotes the $(d_0 \times n)$-matrix with columns $n^{-1 /
2} \cdot \bfX_i$ and $Y$ the $(d_L \times n)$-matrix with columns $n^{-1 / 2}
\cdot \bfY_i$. This loss features an intricate geometry that depends on the
depth $L$ \citep{kawaguchi_2016, achour_malgouyres_et_al_2024}. The equivalent
of the S-SAM loss \eqref{eq::sam_exp_loss} for this setting is given by the
expectation of $\wtL(\bftheta) = \norm{Y - \wtmweight X}^2$, where $\wtmweight =
\wtmweight_L \cdots \wtmweight_1$ and $\wtmweight_{\ell, ij} \sim
\calN\big(W_{\ell, ij}, \eta^2\big)$.

As mentioned in the introduction, in a classification task the generalization
gap predicted by the PAC Bayes framework depends both on the norm of the
parameter estimate and the sharpness of the loss, see \cite{dziugaite_roy_2017,
neyshabur_bhojanapalli_et_al_2017}. For unbounded losses, such as the squared
distance, the tail behavior of the loss requires further control. Adapting a
novel PAC Bayes bound in \cite{chugg_wang_et_al_2023} yields the following
generalization gap bound for arbitrary linear networks in terms of average
sharpness.

\begin{lemma}
  \label{lem::pac_bound}
  Suppose the $(\bfX_i, \bfY_i)$ are sampled from a distribution with finite
  fourth moment fourth moment. For any $\delta \in (0, 1)$ and any parameter
  estimate $\bftheta = (W_1, \ldots, W_L)$, the bound \begin{align*}
    \E\big[\calL(\bftheta)\big] - \calL(\bftheta) \leq S_{\avg}(\bftheta, \eta)
    + \dfrac{1}{\sqrt{n}} \cdot \left(\dfrac{1}{2 \eta^2} \cdot \sum_{\ell =
    1}^L \norm{W_\ell}^2 + \log \dfrac{1}{\delta} + \dfrac{1}{2} \cdot
    \E\big[\wtL(\bftheta)^2\big]\right)
  \end{align*} holds with probability $1 - \delta$ over the draw of the data.
\end{lemma}

The expression $\E\big[\wtL(\bftheta)^2\big]$ controls the total variance of
the empirical loss under both the data distribution and the PAC Bayes posterior
$\wtmweight_\ell$ for the parameter estimate $W_\ell$, with the condition on the
moments of $(\bfX, \bfY)$ ensuring that this expectation stays finite. Due to
the quadratic nature of the loss, the tail behavior of $\wtL(\bftheta)$ depends
strongly on the parameter estimate itself, requiring a fourth-order control.
This expectation uses the unknown distribution of $(\bfX, \bfY)$, so the bound
in Lemma \ref{lem::pac_bound} must be understood as an oracle bound, see Section
1.4 of \cite{alquier_2024} for additional discussion of this point. As in
\cite{haddouche_viallard_et_al_2025}, we may further connect the
variance-control expression with measures of sharpness by exploiting geometric
information contained in the posterior measure. Recall that a normal
distribution with non-singular covariance matrix $\Sigma$ satisfies a
Poincar\'{e} inequality with constant $\sigmax(\Sigma)$ (Corollary 2.1,
\cite{chafai_2004}), meaning \begin{align}
  \label{eq::loss_poinc}
  \begin{split}
    \E\big[\wtL(\bftheta)^2\big] \leq \E\big[\wtL(\bftheta)\big]^2 +
    \eta^2 \cdot \E\Big[\norm[\big]{\nabla \wtL(\bftheta)}_2^2\Big]
  \end{split}
\end{align} since $\Sigma = \eta^2 I$ in our setting. Write $p$ for the total
number of model parameters and let $\bfxi$ denote the length $p$ vector of
normal perturbations added to the weight matrices. Assuming sub-normal gradient
noise, the critical points of the empirical loss converge to critical points of
the population loss at an asymptotic rate $n^{- 1 / 2}$ inside any compact set
\citep{mei_bay_et_al_2018}. In the large data regime and near critical points of
the empirical loss, we may then use the triangle inequality and a first-order
approximation to the gradient at $\wtmweight_1, \ldots, \wtmweight_L$ to
estimate \begin{align*}
  \E\Big[\norm[\big]{\nabla \wtL(\bftheta)}_2^2\Big] \approx
  \E\Big[\norm[\big]{\nabla \wtL(\bftheta) - \nabla \calL(\bftheta)}_2^2\Big]
  &\leq \E\Big[\norm[\big]{\nabla^2 \calL(\bftheta)}_{\op}^2\Big] \cdot
  \E\big[\norm{\bfxi}_2^2\big] + O\Big(\E\big[\norm{\bfxi}_2^4\big]\Big)\\
  &= p \eta^2 \cdot \E\Big[\norm[\big]{\nabla^2 \calL(\bftheta)}_{\op}^2\Big] +
  O\big(p \eta^4\big).
\end{align*} Suppose $\eta = (\eta_0 p)^{- 1 / 2}$ with $\eta_0 > 0$, then $\xi$
approximates the uniform distribution on a sphere of radius $\eta_0$ as $p \to
\infty$. The computation in the previous display now affords control of the
gradient term in \eqref{eq::loss_poinc} at the level $O\big(\eta_0^4 p\inv \cdot
\norm{\nabla^2 \calL(\bftheta)}_{\op}^2\big)$ near critical points. For
empirical losses with sub-exponentially distributed second derivative, the
empirical Hessian converges to its population counterpart at the asymptotic rate
$(p / n)^{- 1 / 2}$ in operator norm, uniformly over any compact set
\citep{mei_bay_et_al_2018}. In this setting, it suffices to control the operator
norm of the empirical Hessian matrix, given enough data. A trajectory-wise sum
over the gradient sensitivity $\nabla \wtL(\bftheta) - \nabla \calL(\bftheta)$
also appears in some last-iterate generalization gap bounds for SGD
\citep{neu_dziugaite_et_al_2021}.

Similar to sharpness-based PAC Bayes bounds in a classification setting
\citep{dziugaite_roy_2017, neyshabur_bhojanapalli_et_al_2017}, Lemma
\ref{lem::pac_bound} demands control over both average sharpness and parameter
norm. The former determines the tightest possible upper bound as $n \to \infty$.
Since $\calL$ depends polynomially on the entries of the weight matrices, a
bound on the parameter norm implies a finite second moment
$\E\big[\wtL(\bftheta)^2\big]$. Then, the second part of the right-hand side
expression in Lemma \ref{lem::pac_bound} vanishes at the parametric rate $n^{-1
/ 2}$, leaving only the sharpness term.

In comparison with Lemma \ref{lem::grad_exp_cond}, for generic linear networks
the induced sharpness regularizer $R(\bftheta)$ features a more intricate
structure, due to correlation between the entries of $\wtmweight_\ell
\wtmweight_\ell\tran$. For a random matrix $Z$, define $\Cov(Z) = \E[Z Z\tran] -
\E[Z] \E[Z]\tran$, then an argument analogous to the proof of Lemma
\ref{lem::grad_exp_cond} yields \begin{align}
  \label{eq::exp_reg_lnn}
  \calL_R(\bftheta) = \E\big[\wtL(\bftheta)\big] =
  \calL(\bftheta) + \Tr\big(\Cov(\wtmweight X)\big)
\end{align} Our aim is to understand the penalty term $R(\bftheta) =
\Tr\big(\Cov(\wtmweight X)\big)$. Define $Q_0 = X X\tran$ and apply the
recursion \begin{align}
  \label{eq::Q_rec_def}
  Q_\ell = W_\ell Q_{\ell - 1} W_\ell\tran + \eta^2 \Tr(Q_{\ell - 1}) \cdot
  I_{d_\ell}, \qquad \ell=1, 2, \ldots, L.
\end{align} If $Z$ features independent normal entries with common variance
$\eta^2$ and $\E[Z] = M$, then $\E\big[Z A Z\tran\big] = M A M\tran + \eta^2
\Tr(A) \cdot I$ for any matrix $A$ of suitable size. Induction on $\ell$ now
proves the identity \begin{align}
  \label{eq::reg_rewrite_gen}
  R(\bftheta) = \Tr(Q_L) - \Tr\big(W X X\tran W\tran\big).
\end{align} Letting $L = 2$, we obtain the weighted ridge penalty \begin{align*}
  R(\bftheta) &= \eta^2 \norm{X}^2 \cdot \norm{W_2}^2 + \eta^2 d_2 \cdot
  \norm{W_1 X}^2,
\end{align*} see Section 2.4 of \cite{orvieto_raj_et_al_2023} for further
discussion of this case. When $L = 3$, a more intricate regularization term
emerges, involving products of the weight matrices and higher powers of the
noise level $\eta$, namely \begin{align*}
  R(\bftheta) &= \eta^4 \cdot \Big(d_1 \norm{X}^2 \cdot \norm{W_3}^2 + d_3
  \norm{X}^2 \cdot \norm{W_2}^2 + d_3 d_2 \cdot \norm{W_1 X}^2\Big)\\
  &\qquad + \eta^2 \cdot \Big(\norm{X}^2 \cdot \norm{W_3 W_2}^2 +
  \norm{\weight_3}^2 \norm{W_1 X}^2 + d_3 \cdot \norm{W_2 W_1 X}^2\Big) +
  C_{\eta, X},
\end{align*} where $C_{\eta, X}$ denotes a constant expression that does not
depend on the weight matrices. 

We now characterize the stationary points of any functional of the form
\begin{align}
  \label{eq::fctal_general}
  \bftheta \mapsto F\big(W_L \cdots W_1\big) + \Tr\big(\Cov(\wtmweight X)\big)
\end{align} with differentiable $F : \R^{d_L \times d_0} \to \R$, which includes
\eqref{eq::exp_reg_lnn} since $\calL(\bftheta)$ only depends on the product
matrix $W$. For a functional $f$ taking matrix arguments, we will write $\nabla
f(A)$ as a matrix with the same shape as $A$. In analogy with the forward
recursion \eqref{eq::Q_rec_def} for $Q_\ell$, we also define the backward
recursion \begin{align*}
  B_{\ell - 1} = W_\ell\tran B_\ell W_\ell + \eta^2 \Tr(B_\ell) \cdot I_{d_{\ell
  - 1}}, \qquad \ell = L, L - 1, \ldots, 2.
\end{align*} started at $B_L = I_{d_L}$.

\begin{theorem}
  \label{thm::gen_weight_bal}
  Let $(W_1, \ldots, W_L)$ be a stationary point of the regularized functional
  \eqref{eq::fctal_general}, then \begin{align}
    \label{eq::wb_thm_conc_1}
    \Tr(Q_{\ell - 2}) \cdot W_\ell\tran B_\ell W_\ell 
    = \Tr(B_\ell) \cdot W_{\ell - 1} Q_{\ell - 2} W_{\ell-1}\tran 
  \end{align} for each $\ell = 2, 3, \ldots, L$. Moreover, for any $\ell = 1,
  \ldots, L$, \begin{align*}
    B_\ell = \nabla_{Q_\ell} \Tr(Q_L).
  \end{align*}
\end{theorem}

In comparison with the classical weight balancing conditions $W_\ell\tran W_\ell
= W_{\ell - 1} W_{\ell - 1}\tran$ that occur for $\ell_2$-weight decay
(\cite{arora_cohen_et_al_2018, nguegnang_rauhut_et_al_2024}), we see that
\eqref{eq::wb_thm_conc_1} involves the forward and backward products $Q_{\ell -
2}$ and $B_\ell$, weighted by their respective traces. We may interpret
\eqref{eq::wb_thm_conc_1} as a generalized weight balancing condition.
Interestingly, the roles of $Q_{\ell - 2}$ and $B_\ell$ on the right and
left-hand side of the equation are interchanged.

Due to the recursive definitions, $B_\ell$ only depends on the weight matrices
$W_{\ell + 1}, \ldots, W_L$ and $Q_{\ell - 2}$ only on $X X\tran, W_1, \ldots
,W_{\ell - 2}$. For $L = 3$, the generalized weight balancing condition implies
the following structure:

\begin{lemma}
  \label{lem::comm_three}
  Let $L = 3$, then any stationary point $(W_1, W_2, W_3)$ of
  \eqref{eq::fctal_general} satisfies the commutator relationships
  \begin{align*}
    W_3\tran W_3 W_2 W_2\tran &= W_2 W_2\tran W_3\tran W_3 \\
    W_2\tran W_2 W_1 X X\tran W_1\tran &= W_1 X X\tran W_1\tran W_2\tran W_2.
  \end{align*}
\end{lemma}

The commutator relationships in the previous lemma give a weaker constraint than
the classical weight balancing. In particular, the first commutator relationship
implies simultaneous diagonalizability of $W_3\tran W_3$ and $W_2 W_2\tran$.
Further, the right eigenspaces of $W_3$ coincide with the left eigenspaces of
$W_2$.

\section{Conclusion and Outlook}
\label{sec::disc}

We may summarize our results for S-SAM as follows: For diagonal linear networks,
S-SAM induces balanced solutions that result in sparse predictors. In contrast
with the unregularized model, this bias persists for arbitrary initializations
and macroscopic step-sizes, due to the change in expected gradient vector field
directly regularizing the loss landscape (Theorem \ref{thm::crit_set_A}) and
forcing the underlying gradient flow/descent dynamics to balance the weight
matrices at a fast rate (Theorem \ref{thm::grad_flow_bal} and Theorem
\ref{thm::grad_desc_bal}). The balancing explicitly controls the sharpness
(Lemma \ref{lem::bal_reg}), so this also gives a rate in terms of the noise
level at which the underlying dynamics minimize sharpness. Further, the induced
regularizer controls the parameter norm of the gradient descent iterates by
forcing them towards a shrinkage-thresholding of the true parameter (Theorem
\ref{thm::crit_set_B} and Theorem \ref{thm::conv_abs_disc}), which also
illustrates how the noise level $\eta$ determines the signal strength needed for
a feature to become active. Together, the parameter norm and sharpness measure
the generalization performance via a PAC Bayes bound (Lemma
\ref{lem::pac_bound}). Lastly, convergence towards a critical point can be shown
for the projected gradient descent with S-SAM (Theorem
\ref{thm::proj_sam_conv}). We conclude that S-SAM has the intended effect in the
diagonal linear setup and efficiently regularizes the algorithm to control the
relevant generalization measures.

To complete the article, we now outline some possible directions for further
research. A natural follow-up question concerns the behavior of S-SAM as a
diffusion in the parameter space for constant step-sizes. Strong results in this
vein exist for stochastic gradient Langevin dynamics (SGLD)
\citep{raginsky_rakhlin_et_al_2017}. Unlike for SGLD, the gradient noise induced
by S-SAM highly depends on the current parameter location. Characterizing the
stationary distribution of gradient descent with some types of
location-dependent noise is possible, but requires a finely tuned setup
\citep{yoshida_nakakita_et_al_2024}. In general, given sufficient regularity
conditions, stochastic gradient descent with a constant step-size will converge
to a stationary distribution that reflects the local geometry of the loss
surface \citep{azizian_iutzeler_et_al_2024}. This is also illustrated in the
constant step-size S-SAM trajectories shown in Figure \ref{fig::gd_comp}, where
the S-SAM gradient slowly minimizes the balancing of the weight matrices and
then enters a stationary regime, concentrating around one of the global
minimzers of $\calL_R$.  Unlike the explicitly regularized $\nabla \calL_R$, the
S-SAM algorithm does not necessarily decrease the balancing during every
iteration. Rather, the balancing experiences a downward trend, due to the
alignment $\iprod[\big]{\nabla \calL_R, \nabla \wtL_k}$ between the expected and
random gradient seemingly being positive with higher probability than negative.
In this sense, we may expect S-SAM to act similar to a zeroth-order method, see
also \cite{bos_schmidt-hieber_2024, dexheimer_schmidt-hieber_2024}.

To conclude, we note that the choice of isotropic normal noise for the parameter
perturbations $\weight_{ij} + \xi_{ij}$ in Lemma \ref{lem::pac_bound} is
essentially arbitrary, and perhaps sub-optimal. PAC Bayes bounds hold for any
fixed prior distribution on the parameter space and any absolutely continuous
data-dependent posterior distribution. Choosing both normal priors and
posterior, as in Lemma \ref{lem::pac_bound}, gives a convenient setup and the
average sharpness $S_{\avg}(\ \cdot\ , \eta)$ seems well-motivated from an
intuitive standpoint, but there is no guarantee that this choice gives the
optimal generalization gap. In particular, one may try to choose the optimal
posterior for a given prior to minimize the gap, as in
\cite{dziugaite_roy_2017}. Even in a diagonal linear network, various
generalization measures capture different aspects of the relationship between
the empirical and population loss and may perform differently depending on the
observed data \citep{andriushchenko_croce_et_al_2023}. Generalization measures
specialized to gradient descent either require complexity bounds along the
trajectory \citep{sachs_van_erven_et_al_2023, park_simsekli_et_al_2022} or
trajectory-wise bounds on the sensitivity of the loss and its gradient to slight
perturbations \citep{neu_dziugaite_et_al_2021}. By adapting to the change in
local geometry along the trajectory, one would hope to strike the correct
balance between sufficient exploration of the parameter space and refinement of
the solution once the algorithm enters the basin of attraction of a good
local/global minimum. This could be incorporated into the PAC Bayes bound of
Lemma \ref{lem::pac_bound} by choosing a posterior that incorporates the
geometry of the loss around the learned parameter value. To optimize the bound,
the gradient descent iterates may then be trained with adaptive injected noise
that makes them robust towards perturbations sampled from this posterior, in
analogy with robustness towards isotropic normal perturbations optimizing the
average sharpness. As shown in Section 5 of \cite{bartlett_et_al_2023}, the
deterministic variant of SAM incorporates third-order information into the
gradient descent update to find wide valleys in the loss landscape. Hence, we
expect that a posterior distribution that penalizes higher-order geometric
information may yield stronger generalization guarantees. We leave the details
to future work.


\acks{The authors thank Hermen Jan Hupkes, Scott Pesme, Taiji Suzuki, Matus
Telgarsky, and Harro Walk for productive conversations during the research
phase. Further, the authors thank three anonymous referees and an editor for
their time and comments, as well as Bin Yu for suggesting the connection with
fractional penalties in Theorem \ref{thm::crit_set}.

\vspace{10pt}

\noindent \begin{minipage}{0.05\textwidth}
  \includegraphics[height = 2.5\baselineskip]{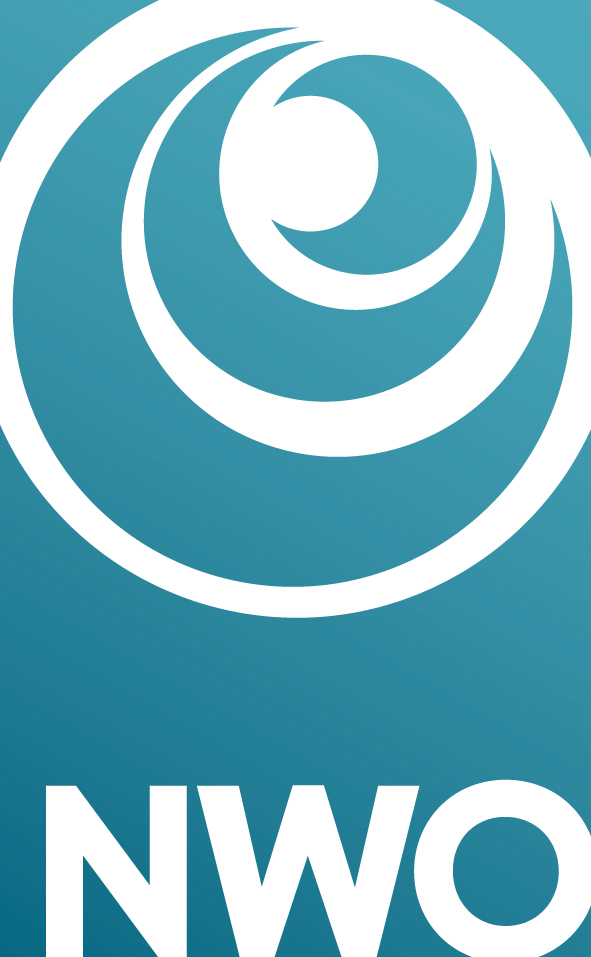}
\end{minipage} \hfill \begin{minipage}{0.935\textwidth}
  This publication is part of the project \textit{Statistical Foundation for
  Multilayer Neural Networks} (project number VI.Vidi.192.021 of the Vidi ENW
  programme) financed by the Dutch Research Council (NWO).
\end{minipage}}


\appendix

\section{Proofs for Section \ref{sec::lin_nn_back}}
\label{sec::lin_nn_back_proof}

\subsection{Proof of Lemma \ref{lem::grad_exp_cond}}

We start by proving that the expectation of the randomized loss in
\eqref{eq::sam_exp_loss} indeed evaluates to $\calL_R$. By definition, the
random weight vectors $\wtweight_\ell$ are independent, given their noiseless
versions $\weight_\ell$. Each entry of $\eta^{- 2} \cdot \wtweight_\ell^2$
follows a non-central $\chi^2$-distribution with a single degree of freedom,
meaning $\E\big[\wtweight_\ell^2 \mid \weight_\ell\big] = \weight_\ell^2 +
\eta^2 \cdot \bfone$. Combining these facts, \begin{align}
  \E\Big[\wtweight_r \odot \cdots \odot \wtweight_s \bigmid \bftheta\Big] &=
  \bigodot_{m = s}^r \weight_{m} \label{eq::prod_exp} \\
  \E\Big[\wtweight_r^2 \odot \cdots \odot \wtweight_s^2 \bigmid \bftheta\Big] &=
  \bigodot_{m = s}^r \E\big[\wtweight_m^2 \mid \weight_m\big] = \bigodot_{m =
  s}^r \big(\weight_{m}^2 + \eta^2 \cdot \bfone\big) \label{eq::prod_exp_sq}
\end{align} whenever $r \geq s$. Fixing deterministic weight matrices
$\weight_1, \ldots, \weight_L$ and applying \eqref{eq::prod_exp}, we now compute
\begin{align*}
  \E\Big[\norm[\big]{\bfY - X \wtweight}_2^2\Big] = \norm[\big]{\bfY - X
  \weight}_2^2 + \Tr\Big(\Cov\big(\bfY - X \wtweight\big)\Big) = \calL(\bftheta)
  + \Tr\Big(\Cov\big(X \wtweight\big)\Big)
\end{align*} where the last equality follows from independence of $\bfY$ and the
random perturbations $\bfxi_\ell$. It now suffices to show that the trace of the
covariance matches $R$. Given a random vector $\mathbf{z}$ with uncorrelated
entries and any matrix $A$, note that $\Tr\big(\Cov(A \mathbf{z})\big) =
\Tr\big(A \Cov(\mathbf{z}) A\tran\big) = \Tr\big(\Diag(A\tran A)
\Cov(\mathbf{z})\big)$. For fixed weights, the random vector $\wtweight$
features independent entries, so together with \eqref{eq::prod_exp} and
\eqref{eq::prod_exp_sq} we find that \begin{align*}
  \Tr\Big(\Cov\big(X \wtweight\big)\Big) &= \Tr\Big(\Diag\big(X\tran X\big)
  \Cov(\wtweight)\Big)\\
  &= \bfone\tran \Diag\big(X\tran X\big) \left(\E\left[\bigodot_{\ell = 1}^L
  \wtweight_\ell^2\right] - \E\left[\bigodot_{m = 1}^L
  \wtweight_m\right]^2\right)\\
  &= \bfone\tran \Diag\big(X\tran X\big) \left(\bigodot_{\ell = 1}^L
  \big(\weight_\ell^2 + \eta^2 \cdot \bfone\big) - \bigodot_{m = 1}^L
  \weight_m^2\right).
\end{align*} The latter matches the definition of $R$, thereby completing the
first part of the proof.

We now turn our attention to the conditional expectation of the randomized
gradients in \eqref{eq::backprop_noise}. Integrating over the empirical measure
on the observed data yields \begin{align*}
  - \dfrac{2}{n} \cdot \sum_{i = 1}^n \bfX_i \big(Y_i - \bfX_i\tran
  \wtweight\big) \bigodot_{\substack{m = 1\\ m \neq \ell}}^L \wtweight_{m} &= -
  2 \cdot \left(\dfrac{1}{n} \cdot \sum_{i = 1}^n \bfX_i Y_i - \dfrac{1}{n}
  \cdot \sum_{i = 1}^n \bfX_i \bfX_i\tran \wtweight\right) \bigodot_{\substack{m
  = 1\\ m \neq \ell}}^L \wtweight_{m}\\
  &= - 2 \cdot \big(X\tran \bfY - X\tran X \wtweight\big) \bigodot_{\substack{m
  = 1\\ m \neq \ell}}^L \wtweight_{m} = - \nabla_{\weight_\ell} \norm[\big]{\bfY
  - X \wtweight}_2^2.
\end{align*} By construction, the norm is polynomial in the entries of the
randomized weight vectors, so for any fixed $\weight_1, \ldots, \weight_L$ we
may exchange the expectation over the normal noise with the gradient. This would
complete the proof, but as we require the actual expectations for latter results
we shall take the scenic route.

Due to independence of the normal perturbations, entries of $\bfw$ and any
sub-product of $\wtw$ with different indices are uncorrelated. Further,
\eqref{eq::prod_exp} and \eqref{eq::prod_exp_sq} imply \begin{align*}
  &\left(\E\left[\wtweight \left(\bigodot_{\substack{r = 1\\ r \neq \ell}}^L
  \wtweight_{r}\right)\tran\right] - \weight \left(\bigodot_{\substack{s = 1\\ s
  \neq \ell}}^L \weight_{s}\right)\tran\right)_{ij} \\
  =\ &\begin{cases}
    \weight_{\ell, i} \cdot \prod_{\substack{r = 1\\ r \neq \ell}}^L
    \big(\weight_{r, i}^2 + \eta^2\big) - \weight_{\ell, i} \cdot
    \prod_{\substack{s = 1\\ s \neq \ell}}^L \weight_{s, i}^2 &\qquad \mbox{ if
    } i = j, \\
    0 &\qquad \mbox{ otherwise.}
  \end{cases}
\end{align*} For any fixed matrix $A$ and random vectors $\bfu$ and $\bfv$, the
$i$\textsuperscript{th} entry of $A \bfu \odot \bfv$ matches $(A \bfu
\bfv\tran)_{ii}$. In particular, $A \bfu \odot \bfv = \Diag(A) \E[\bfu \odot
\bfv]$ whenever the latter expectation is diagonal. Taking $A = X\tran X$, $\bfu
= \wtweight$, and $\bfv = \bigodot_{\substack{m = 1\\ m \neq \ell}}^L
\weight_{m}$ now leads to \begin{align*}
  \E\left[X\tran X \wtweight \bigodot_{\substack{m = 1\\ m \neq \ell}}^L
  \wtweight_{m}\right] &= X\tran X \weight \bigodot_{\substack{m = 1\\ m \neq
  \ell}}^L \weight_{m} + \Diag\big(X\tran X\big) \weight_\ell \odot
  \left(\bigodot_{\substack{r = 1\\ r \neq \ell}}^L \big(\weight_{r}^2 + \eta^2
  \cdot \bfone\big) - \bigodot_{\substack{s = 1\\ s \neq \ell}}^L
  \weight_{s}^2\right),
\end{align*} which in turn implies \begin{align*}
  - 2 \cdot \E\left[\big(X\tran \bfY - X\tran X \wtweight\big)
  \bigodot_{\substack{m = 1\\ m \neq \ell}}^L \wtweight_{m}\right] &= - 2 \cdot
  \big(X\tran \bfY - X\tran X \weight\big) \bigodot_{\substack{m = 1\\ m \neq
  \ell}}^L \weight_{m}\\
  & + 2 \cdot \Diag\big(X\tran X\big) \weight_\ell \odot
  \left(\bigodot_{\substack{r = 1\\ r \neq \ell}}^L \big(\weight_{r}^2 + \eta^2
  \cdot \bfone\big) - \bigodot_{\substack{s = 1\\ s \neq \ell}}^L
  \weight_{s}^2\right).
\end{align*} The latter indeed matches the gradient of $\calL_R$; applying
\eqref{eq::backprop_noise} with $\eta = 0$ to compute the gradient of $\calL$
and the chain rule to compute the gradient of $R$ yields \begin{align}
  \nabla_{\weight_\ell} \calL(\bftheta) &= - 2 \cdot
  \big(X\tran \bfY - X\tran X \weight\big) \bigodot_{\substack{m = 1\\ m \neq
  \ell}}^L \weight_{m}\label{eq::diff_loss} \\
  \nabla_{\weight_\ell} R(\bftheta) &= 2 \cdot \Diag\big(X\tran X\big)
  \weight_\ell \odot \left(\bigodot_{\substack{r = 1\\ r \neq \ell}}^L
  \big(\weight_{r}^2 + \eta^2 \cdot \bfone\big) - \bigodot_{\substack{s = 1\\ s
  \neq \ell}}^L \weight_{s}^2\right). \label{eq::diff_reg}
\end{align}

\section{Proofs for Section \ref{sec::land_no_noise}}  
\label{sec::land_no_noise_proof}

\subsection{Proof of Theorem \ref{thm::crit_set}}

\begin{enumerate}
  \item Any critical point $\bftheta$ of $\calL_R$ must automatically satisfy
    the constraint $\nabla_{\weight_\ell} \calL_R(\bftheta) \odot \weight_\ell =
    \bfzero = \nabla_{\weight_{\ell + 1}} \calL_R(\bftheta) \odot \weight_{\ell
    + 1}$ for each $\ell = 1, \ldots, L - 1$. Using \eqref{eq::diff_loss}, we
    find that $\nabla_{\weight_\ell} \calL_R(\bftheta) \odot \weight_\ell = - 2
    \cdot \big(X\tran \bfY - X\tran X \weight\big) \odot \weight$ does not
    depend on $\ell$ anymore, which implies the desired constraint \begin{align}
      \label{eq::reg_cons_eq}
      \nabla_{\weight_\ell} R(\bftheta) \odot \weight_\ell =
      \nabla_{\weight_{\ell + 1}} R(\bftheta) \odot \weight_{\ell + 1}.
    \end{align}

    For the second claim, the expression \eqref{eq::diff_reg} for the gradients
    of $R$ yields \begin{align*}
      \nabla_{\weight_\ell} R(\bftheta) \odot \weight_\ell &= 2 \cdot
      \Diag\big(X\tran X\big) \weight_\ell^2 \odot \left(\bigodot_{\substack{r =
      1\\ r \neq \ell}}^L \big(\weight_{r}^2 + \eta^2 \cdot \bfone\big) -
      \bigodot_{\substack{s = 1\\ s \neq \ell}}^L \weight_{s}^2\right)\\
      &= 2 \cdot \Diag\big(X\tran X\big) \weight_\ell^2 \bigodot_{\substack{r =
      1\\ r \neq \ell}}^L \big(\weight_{r}^2 + \eta^2 \cdot \bfone\big) - 2
      \cdot \Diag\big(X\tran X\big) \weight^2
    \end{align*} with an analogous identity for $\nabla_{\weight_{\ell + 1}}
    R(\bftheta) \odot \weight_{\ell + 1}$ following by symmetry. The second term
    $\Diag(X\tran X) \weight^2$ does not depend on $\ell$ anymore, so
    \eqref{eq::reg_cons_eq} turns into \begin{align}
      \bfzero &= \nabla_{\weight_\ell} R(\bftheta) \odot \weight_\ell -
      \nabla_{\weight_{\ell + 1}} R(\bftheta) \odot \weight_{\ell + 1}
      \nonumber\\
      &= \Diag\big(X\tran X\big) \weight_\ell^2 \bigodot_{\substack{r = 1\\ r
      \neq \ell}}^L \big(\weight_{r}^2 + \eta^2 \cdot \bfone\big) -
      \Diag\big(X\tran X\big) \weight_{\ell + 1}^2 \bigodot_{\substack{s = 1\\ s
      \neq \ell + 1}}^L \big(\weight_{s}^2 + \eta^2 \cdot \bfone\big)
      \nonumber\\
      &= \eta^2 \cdot \Diag\big(X\tran X\big) \left(\bigodot_{\substack{r = 1\\
      r \neq \ell, \ell + 1}}^L \big(\weight_{r}^2 + \eta^2 \cdot
      \bfone\big)\right) \odot \Big(\weight_\ell^2 - \weight_{\ell + 1}^2\Big)
      \label{eq::reg_diff_sq}
    \end{align} Each vector $\weight_r^2 + \eta^2 \cdot \bfone$ features
    strictly positive entries. In turn, \eqref{eq::reg_diff_sq} only vanishes if
    the entries of $\weight_\ell^2 - \weight_{\ell + 1}^2$ equal zero wherever
    $\Diag(X\tran X)$ has non-zero diagonal elements. This constraint holds for
    each $\ell = 1, \ldots, L - 1$ simultaneously, which completes the proof.
  \item Fix a critical point of $\calL_R$ and let $S_{\weight}$ and
    $S_{\weight_\ell}$ denote the matrices with respective diagonal entries
    $\mathrm{sign}(\weight)$ and $\mathrm{sign}(\weight_\ell)$. Write $S =
    \abs{S_{\weight}}$ for the element-wise absolute value of $S_{\weight}$,
    then $S = S_{\weight_\ell}^2$ for any $\ell$. As $\Diag(X\tran X)$ is
    non-singular, the previous part of the theorem yields $\weight_\ell^2 =
    \weight_{\ell + 1}^2$ and in turn $\weight_\ell = S_{\weight_\ell}
    \abs{\weight_\ell}^{1 / L}$. In particular, $\weight_{j} = 0$ implies
    $\weight_{1, j} = \cdots = \weight_{L, j} = 0$ and so $\prod_{m \in I
    \setminus \{\ell\}} S_{\weight_m} = S_{\weight_\ell} \prod_{m \in I}
    S_{\weight_m}$. The respective gradients in \eqref{eq::diff_loss} and
    \eqref{eq::diff_reg} now take the form \begin{align*}
      \nabla_{\weight_\ell} \calL(\bftheta) &= - 2 \cdot X\tran \big(\bfY - X
      \weight\big) \odot S_{\weight_\ell} S_{\weight} \abs{\weight}^{(L - 1) /
      L} \\
      \nabla_{\weight_\ell} R(\bftheta) &= 2 \cdot \Diag\big(X\tran X\big)
      \bigg(\big(\abs{\weight}^{2 / L} + \eta^2 \cdot \bfone\big)^{L - 1} -
      \abs{\weight}^{2 (L - 1) / L}\bigg) \odot S_{\weight_\ell}
      \abs{\weight}^{1 / L}
    \end{align*} Note that $\bfu \odot S_{\weight_\ell} \bfv = S_{\weight_\ell}
    (\bfu \odot \bfv)$ and $\weight = S \weight$. Adding the gradients,
    multiplying by $S_{\weight_\ell}$, and dividing element-wise by $S_{\bfw}
    \abs{\weight}^{(L - 1) / L}$ (with the convention that $0 / 0 = 0$) then
    implies \begin{equation}
      \label{eq::grad_pred_solve}
      \begin{split}
      \bfzero &= - 2 \cdot S X\tran \big(\bfY - X S \weight\big)\\
      &\qquad + 2 \cdot S \Diag\big(X\tran X\big) \bigg(\big(\abs{\weight}^{2 /
      L} + \eta^2 \cdot \bfone\big)^{L - 1} - \abs{\weight}^{2 (L - 1) /
      L}\bigg) \odot S_{\bfw} \abs{\weight}^{- (L - 2) / L}. 
      \end{split}
    \end{equation} The latter expression does not depend on $\ell$ anymore and
    $S$ pre-multiplies both terms. If $S_{ii} = 0$, then the corresponding entry
    of \eqref{eq::grad_pred_solve} vanishes. By construction, the diagonal
    entries of $S$ encode the sparsity pattern of $\weight$, meaning $S_{ii} =
    0$ removes $\weight_i$ from the model entirely. In turn, the linear
    predictor $\weight$ at critical points of $\calL_R$ correspond with
    solutions of \eqref{eq::grad_pred_solve} for arbitrary choices of $S$.
    Writing $X_S = S X$, we recognize $- 2 \cdot S X\tran \big(\bfY - X S
    \weight\big)$ as the gradient of $\norm{\bfY - X_S \weight}_2^2$. Similarly,
    the second expression in \eqref{eq::grad_pred_solve} matches \begin{align*}
      &\nabla_{\weight} \Bigg(\bfone\tran \Diag\big(X_S\tran X_S\big)
      \bigg(\Big(\abs{\bfw}^{2 / L} + \eta^2 \cdot \bfone\Big)^L -
      \abs{\bfw}^2\bigg)\Bigg)\\
      =\ &L \cdot \Diag\big(X_S\tran X_S\big) \Big(\abs{\bfw}^{2 / L} + \eta^2
      \cdot \bfone\Big)^{L - 1} \odot \begin{bmatrix}
        \tfrac{\mathrm{d}}{\mathrm{d} \weight_1} \abs{\weight}^{2 / L}_1\\
        \vdots \\
        \tfrac{\mathrm{d}}{\mathrm{d} \weight_d} \abs{\weight}^{2 / L}_d
      \end{bmatrix} + 2 \cdot S_{\bfw} \abs{\weight}\\
      &= 2 \cdot \Diag\big(X\tran_S X_S\big) \bigg(\big(\abs{\weight}^{2 / L} +
      \eta^2 \cdot \bfone\big)^{L - 1} - \abs{\weight}^{2 (L - 1) / L}\bigg)
      \odot S_{\bfw} \abs{\weight}^{- (L - 2) / L} 
    \end{align*} since $\Diag(X_S\tran X_S) = S \Diag(X\tran X)$. Combining
    these computations rewrites \eqref{eq::grad_pred_solve} into $\bfzero =
    \nabla \calL_S(\bfw)$, which completes the proof.
  \item Without loss of generality, we may assume $S = I_d$ and $\weight_{*, i}
    \neq 0$ for each $i = 1, \ldots, d$. Indeed, for $S_{ii} = 0$ both the
    $i$\textsuperscript{th} column of $X$ and $i$\textsuperscript{th} entry of
    $\starweight$ can be safely deleted until $S$ equals the identity matrix of
    some size $d' \leq d$, all without changing the loss in Theorem
    \ref{thm::crit_set_B}. Taking $\weight = \weight(\bft) = \bft \odot
    \starweight$ and $X\tran X = I_d$ in \eqref{eq::grad_pred_solve} now yields
    \begin{align*}
      \bfzero &= - \big(\starweight - \bft \odot \starweight\big) + S_{\bfw}
      \bigg(\big(\abs{\bft \odot \starweight}^{2 / L}  + \eta^2 \bfone\big)^{L -
      1} \odot \abs{\bft \odot \starweight}^{- (L - 2) / L} - \abs{\bft \odot
      \starweight}\bigg).
    \end{align*} The vector pre-multiplied by $S_{\weight}$ features
    non-negative entries since the leading power of the first term cancels with
    $\abs{\bft \odot \starweight}$. If $\bft_i < 0$, then $- \weight_{*, i} +
    \bft_i \cdot \weight_{*, i} < 0$ and $S_{\bfw, ii} \neq \mathrm{sign}(\bft_i
    \cdot \weight_{*, i})$ would cause the $i$\textsuperscript{th} right-hand
    side entry in the previous display to be strictly negative. This creates a
    contradiction, implying that $\bft$ must have non-negative entries. In turn,
    $S_{\bfw, ii} = \mathrm{sign}(\weight_{*, i})$ and we may divide out the
    signs and cancel like terms to arrive at \begin{align*}
      \bfzero &= - \abs{\starweight} + \Big(\bft^{2 / L} \odot
      \abs{\starweight}^{2 / L}  + \eta^2 \cdot \bfone\Big)^{L - 1} \odot
      \bft^{- (L - 2) / L} \odot \abs{\starweight}^{- (L - 2) / L} \\
      &= \abs{\starweight} \odot \Bigg(- \bfone + \Big(\bft^{2 / L} + \eta^2
      \cdot \abs{\starweight}^{- 2 / L} \Big)^{L - 1} \odot \bft^{- (L - 2) /
      L}\Bigg)
    \end{align*} Dividing by $\abs{\starweight}$ and performing the change of
    variables $\bfz = \bft^{1 / L}$, any non-zero solutions $\bft_i$ for the
    previous display correspond with the element-wise roots of \begin{align*}
      p(\bfz) = \Big(\bfz^2 + \eta^2 \cdot \abs{\starweight}^{- 2 / L}\Big)^{L -
      1} \odot \bfz^{- (L - 2)} - \bfone.
    \end{align*} When $L = 2$, the latter reduces to $p(\bfz) = \bfz^2 + \eta^2
    \cdot \abs{\starweight} - \bfone$, which admits the unique non-zero roots
    \begin{align*}
      \bfz_i = \sqrt{1 - \dfrac{\eta^2}{\abs{\weight_{*, i}}}}
    \end{align*} for coordinates $i$ satisfying $\abs{\weight_{*, i}} > \eta^2$.
    This corresponds to the solution $\weight_i = \bft_i \cdot \weight_{*, i} =
    \mathrm{sign}(\weight_{*, i}) \cdot \big(\abs{\weight_{*, i}} -
    \eta^2\big)$, which concludes the proof for $L = 2$.

    Fix $L > 2$, then any non-zero element-wise roots of $p(\bfz)$ exactly match
    those of \begin{align*}
      \bfz^{L - 1} \odot p(\bfz) = \Big(\bfz^2 + \eta^2 \cdot
      \abs{\starweight}^{- 2 / L}\Big)^{L - 1} \odot \bfz - \bfz^{L - 1}
    \end{align*} The coefficient sequence of the latter polynomial features at
    most two sign changes, and so Descartes' rule of signs ensures that $p$
    admits at most two positive roots while counting multiplicity, see
    Proposition 1.2.14 of \cite{bochnak_coste_et_al_1998}. Write $\partial /
    \partial \bfz$ for the element-wise differentiation operator, then
    \begin{align*}
      \dfrac{\partial p(\bfz)}{\partial \bfz} &= 2 (L - 1) \cdot \Big(\bfz^2 +
      \eta^2 \cdot \abs{\starweight}^{- 2 / L}\Big)^{L - 2} \odot \bfz^{- (L -
      3)}\\
      &\qquad - (L - 2) \cdot \Big(\bfz^2 + \eta^2 \cdot \abs{\starweight}^{- 2
      / L}\Big)^{L - 1} \odot \bfz^{- (L - 1)}.
    \end{align*} Any non-zero element-wise roots of the latter exactly match
    those of \begin{align*}
      \Big(\bfz^2 + \eta^2 \cdot \abs{\starweight}^{- 2 / L}\Big)^{- (L - 2)}
      \odot \bfz^{L - 1} \odot \dfrac{\partial p(\bfz)}{\partial \bfz} = L \cdot
      \bfz^2 - (L - 2) \eta^2 \cdot \abs{\starweight}^{- 2 / L},
    \end{align*} which features the unique solution \begin{align*}
      \bfz_* = \sqrt{\dfrac{L - 2}{L}} \cdot \eta \cdot \abs{\starweight}^{- 1 /
      L}.
    \end{align*} The binomial theorem expresses $p(\bfz)$ as a sum of functions
    that are either strictly convex for positive inputs, or affine via
    \begin{align*}
      p(\bfz) = \sum_{k = 0}^{L - 1} \dfrac{(L - 1)!}{k ! (L - 1 - k) !} \cdot
      \big(\eta^2 \cdot \abs{\starweight}^{- 2 / L}\big)^{L - 1 - k} \cdot
      \bfz^{2 k - L + 2} - \bfone,
    \end{align*} thereby proving strict convexity of $p$ over $\ggR^d$ and in
    turn $\bfz_* = \argmin_{\bfz \in \ggR^{d}} p(\bfz)$. This yields the minimal
    values \begin{align*}
      p(\bfz_*) &= \left(\left(1 + \dfrac{L - 2}{L}\right) \cdot \eta^2 \cdot
      \abs{\starweight}^{- 2 / L}\right)^{L - 1} \odot \left(\dfrac{L - 2}{L}
      \cdot \eta^2 \cdot \abs{\starweight}^{- 2 / L}\right)^{- L / 2 + 1} -
      \bfone \\
      &= \left(1 + \dfrac{L - 2}{L}\right)^{L - 1} \left(\dfrac{L -
      2}{L}\right)^{- L / 2 + 1} \cdot \eta^L \cdot \abs{\starweight}\inv -
      \bfone \\ 
      &= \left(1 + \dfrac{L}{L - 2}\right)^{L - 1} \left(\dfrac{L -
      2}{L}\right)^{L / 2} \cdot \eta^L \cdot \abs{\starweight}\inv - \bfone.
    \end{align*} By construction, $p$ can only admit a non-zero root $\bfz_i$ if
    the corresponding minimal value is non-positive, which requires
    \begin{align}
      \label{eq::star_thresh}
      \abs{\weight_{*, i}} \geq \left(1 + \dfrac{L}{L - 2}\right)^{L - 1}
      \left(\dfrac{L - 2}{L}\right)^{L / 2} \cdot \eta^L.
    \end{align}

    Next, we observe that the values of $\bfz_i$ for which \begin{align*}
      \big(p(\bfz_i) + \bfone\big)^{1 / (L - 1)} = \left(\bfz_i^2 +
      \dfrac{\eta^2}{\abs{\weight_{*, i}}^{2 / L}}\right) \cdot \bfz_i^{- (L -
      2) / (L - 1)}
    \end{align*} can equal $1$ must lie in $(0, 1)$. Further, the latter exceeds
    both $f(\bfz_i) = \bfz_i^2 + \eta^2 / \abs{\weight_{*, i}}^{2 / L}$ and
    $g(\bfz_i) = \bfz_i^{- (L - 2) / (L - 1)} \cdot \eta^2 / \abs{\weight_{*,
    i}}^{2 / L}$ over $(0, 1)$. Using \eqref{eq::star_thresh} to bound
    \begin{align*}
      \dfrac{\eta^2}{\abs{\weight_{*, i}}^{2 / L}} &\leq \dfrac{3}{4^{4 /
      3}} < 1,
    \end{align*} we find that $\lim_{\bfz_i \to 0} f(\bfz_i) \leq 1$ and
    $\lim_{\bfz_i \to 1} g(\bfz_i) \leq 1$. Consequently, any non-zero solution
    $\bfz_i$ must lie between the unique positive roots of
    $f(\bfz_i) = 1$ and $g(\bfz_i) = 1$, which yields \begin{align*}
      \left(\dfrac{\eta^2}{\abs{\weight_{*, i}}^{2 / L}}\right)^{(L - 1) / (L -
      2)} \leq \bfz_i \leq \sqrt{1 - \dfrac{\eta^2}{\abs{\weight_{*, i}}^{2 /
      L}}}.
    \end{align*} This corresponds to solutions \begin{align*}
      \left(\dfrac{\eta^2}{\abs{\weight_{*, i}}^{2 / L}}\right)^{L (L - 1) / (L
      - 2)} \cdot \weight_{*, i} \leq \weight_i \leq \sqrt{1 -
      \dfrac{\eta^2}{\abs{\weight_{*, i}}^{2 / L}}}^L \cdot \weight_{*, i},
    \end{align*} thereby completing the proof.
\end{enumerate}

\subsection{Proof of Lemma \ref{lem::bal_reg}}

It suffices to prove the result for $d = 1$, in which case $\weight_\ell$ and
$X\tran X$ are real numbers. Indeed, recall from Lemma \ref{lem::grad_exp_cond}
that $S_{\avg}(\ \cdot\ , \eta) = R(\ \cdot \ )$ and apply \eqref{eq::reg_poly}
to show that \begin{align*}
  R(\bftheta) &= \sum_{I \subsetneq \{1, \ldots, d\}} \eta^{2 (L - \# I)} \cdot
  \norm*{\sqrt{\Diag\big(X\tran X\big)} \bigodot_{m \in I} \weight_m}^2_2\\
  &= \sum_{h = 1}^d \sum_{I \subsetneq \{1, \ldots, d\}} \eta^{2 (L - \# I)}
  \Diag\big(X\tran X\big)_{hh} \cdot \prod_{m \in I} \weight_{m, hh}^2,
\end{align*} which may be minimized separately for each $h$. We will drop the
index $h$ from here on and work in the setting $d = 1$. If $X\tran X = 0$, the
conclusion holds vacuously, otherwise we may drop it as it does not affect
minimization of $R$. Now, $\pi(\weight)$ is given by all scalars
$\weight_{\ell}$, $\ell = 1, \ldots, L$ with product $\weight$. In case,
$\weight = 0$ there is nothing left to prove since $R$ is non-negative and takes
the value $0$ for $\weight_1 = \cdots = \weight_L = 0$. Suppose $\weight \neq 0$
and let $\bftheta$ in $\pi(\weight)$ be balanced, then $\abs{\weight_\ell} =
\abs{\weight}^{1 / L}$ for every $\ell$ and hence \begin{align}
  R(\bftheta) &= \sum_{I \subsetneq \{1, \ldots, d\}} \eta^{2 (L - \# I)} \cdot
  \prod_{m \in I} \weight_{m}^2 \nonumber\\
  &= \sum_{I \subsetneq \{1, \ldots, d\}} \eta^{2 (L - \# I)} \cdot \weight^{2
  \cdot \# I / L} = \sum_{m = 0}^{L - 1} \binom{L}{m} \cdot \eta^{2 (L - m)}
  \cdot \weight^{2 m / L} \label{eq::sharp_bal}
\end{align} Given any other $\bfvartheta = (\bfv_L \cdots \bfv_1)
\in \pi(\weight)$, note that $\prod_{\ell = 1}^L \bfv_\ell / \weight_\ell = 1$.
Consequently, the arithmetic/geometric mean (AM/GM) inequality implies
\begin{align}
  R(\bfvartheta) &= \sum_{I \subsetneq \{1, \ldots, d\}} \eta^{2 (L - \# I)}
  \cdot \prod_{m \in I} V_{m}^2 \nonumber\\
  &= \sum_{I \subsetneq \{1, \ldots, d\}} \eta^{2 (L - \# I)} \cdot
  \weight^{2 \cdot \# I / L} \cdot \prod_{m \in I}
  \left(\dfrac{\bfv_{m}}{\weight_m}\right)^2 \nonumber\\
  &\geq \sum_{m = 0}^{L - 1} \binom{L}{m} \cdot \eta^{2 (L - m)} \cdot
  \weight^{2 m / L} \cdot \left(\prod_{\substack{I \subset \{1, \ldots, d\}\\ \#
  I = m}} \prod_{m \in I} \left(\dfrac{\bfv_{m}}{\weight_m}\right)^2\right)^{1 /
  \binom{L}{m}} \nonumber\\
  &= \sum_{m = 0}^{L - 1} \binom{L}{m} \cdot \eta^{2 (L - m)} \cdot \weight^{2 m
  / L}, \label{eq::sharp_bal_lower}
\end{align} where the last equality follows from each ratio $\bfv_m / \weight_m$
appearing the same number of times in the product. The expressions
\eqref{eq::sharp_bal} and \eqref{eq::sharp_bal_lower} match, proving that
$R(\bfvartheta) \geq R(\bftheta)$. For the reverse implication, it suffices to
note that the AM/GM inequality turns into an equality if, and only if, all
summands are equal. This proves the first part of the lemma.

The result for the trace of the Hessian matrix follows via an analogous
argument. Let $\weight(\bftheta) = \weight_L \odot \cdots \odot \weight_1$
denote the product map, then $\calL(\bftheta)$ results from composing
$\weight(\bftheta)$ with the linear regression loss \eqref{eq::lin_reg_def}. As
a function of $\bfw$, the latter has gradient $- 2 \cdot X\tran (\bfY - X \bfw)$
and Hessian $2 \cdot X\tran X$. Consequently, the chain and product rule
together imply \begin{align}
  \label{eq::hess_loss}
  \nabla^2 \calL(\bftheta) = 2 \cdot D \bfw(\bftheta)\tran X\tran X D
  \bfw(\bftheta) - 2 \cdot \sum_{j = 1}^d \big(X\tran \bfY - X\tran X
  \bfw\big)_j \cdot \nabla^2 \bfw(\bftheta)_j
\end{align} with $D \bfw(\bftheta)$ the Jacobian of the product map
$\bfw(\bftheta)$ and $\nabla^2 \bfw(\bftheta)_j$ the Hessian of the coordinate
map $\bfw(\bftheta)_j = \weight_{L, j} \cdots \weight_{1, j}$. As
$\bfw(\bftheta)_j$ is linear in the entries of the weight vectors, $\nabla^2
\bfw(\bftheta)_j$ features zeros on its main diagonal. Further, $D
\bfw(\bftheta)$ takes the form $\big[D_1, \ldots, D_L\big]$, with diagonal
blocks \begin{align}
  \label{eq::jac_diag}
  D_\ell = \begin{bmatrix}
    \prod_{\substack{r = 1\\ r \neq \ell}}^L \weight_{r, 1} & & \\
    & \ddots & \\
    & & \prod_{\substack{r = 1\\ r \neq \ell}}^L \weight_{r, d}
  \end{bmatrix}.
\end{align} Accordingly, the diagonal entries of $\nabla^2 \calL(\bftheta)$
match those of $D \bfw(\bftheta)\tran X\tran X D \bfw(\bftheta)$, which implies
\begin{align*}
  \Tr\big(\nabla^2 \calL(\bftheta)\big) &= 2 \cdot \sum_{h = 1}^d
  \Diag\big(X\tran X\big)_{hh} \cdot \sum_{\ell = 1}^L \prod_{\substack{m = 1\\
  m \neq \ell}}^L \weight_{m, h}^2.
\end{align*} The optimization again separates over the indices $h$, so it
suffices to prove the result for $d = 1$ and we may drop constant multipliers.
Let $\bftheta$ and $\bfvartheta$ be as in the first part of the proof, then the
AM/GM inequality implies \begin{align*}
  \Tr\big(\nabla^2 \calL(\bfvartheta)\big) &= \sum_{\ell = 1}^L
  \prod_{\substack{m = 1\\ m \neq \ell}}^L \bfv_{m}^2\\
  &= \sum_{\ell = 1}^L \weight^{2 - 2 / L} \cdot \prod_{\substack{m = 1\\ m \neq
  \ell}}^L \left(\dfrac{\bfv_m}{\weight_m}\right)^2 = \weight^{2 - 2 / L} \cdot
  \sum_{\ell = 1}^L \left(\dfrac{\bfv_\ell}{\weight_\ell}\right)^2 \geq L \cdot
  \weight^{2 - 2 / L}.
\end{align*} This matches the expression for balanced weights and hence proves
the second part of the lemma, with the ``only if'' direction again following
from tightness in the AM/GM inequality for equal scalars.

\section{Proofs for Section \ref{sec::conv_grad_desc}}
\label{sec::conv_grad_desc_proof}

\subsection{Proof of Lemma \ref{lem::gf_coerc}}

Using the weight-decay term \eqref{eq::reg_wd} and monotonicity of the gradient
flows due to \eqref{eq::grad_flow_monot}, we deduce the bound \begin{align*}
  \sum_{\ell = 1}^L \norm[\Big]{\sqrt{\Diag\big(X\tran X\big)}
  \weight_\ell(t)}^2_2 \leq \dfrac{1}{\eta^{2 L - 2}} \cdot \calL_R(\bftheta_t)
  \leq \dfrac{1}{\eta^{2 L - 2}} \cdot \calL_R(\bftheta_0)
\end{align*} so it suffices to show that the left-hand side bounds the squared
norm of the network parameter, which follows directly from the assumed
non-singularity of $\Diag(X\tran X)$ and $\min_{i = 1, \ldots, d} \Diag(X\tran
X)_{ii} \geq \sigminp(X\tran X)$.

\subsection{Proof of Theorem \ref{thm::grad_flow_bal}}

Applying the product rule and commutativity of diagonal matrices, as well as
recalling the definition of the gradient flows \eqref{eq::sgd_sam_gf}, the
time-derivative of $\weight_\ell^2(t) - \weight_{\ell + 1}^2(t)$ satisfies
\begin{align*}
  \dfrac{\rmd}{\rmd t} \Big(\weight_\ell^2(t) - \weight_{\ell + 1}^2(t)\Big) = -
  2 \cdot \bigg(\nabla_{\weight_\ell(t)} \calL_R(\bftheta_t) \odot
  \weight_\ell(t) - \nabla_{\weight_{\ell + 1}(t)} \calL_R(\bftheta_t) \odot
  \weight_{\ell + 1}(t)\bigg)
\end{align*} Recall from the proof of Theorem \ref{thm::crit_set_A} that
\begin{align*}
  \nabla_{\weight_\ell} \calL_R(\bftheta_t) \odot \weight_\ell -
  \nabla_{\weight_{\ell + 1}} \calL_R(\bftheta_t) \odot \weight_{\ell + 1} =
  \nabla_{\weight_\ell} R(\bftheta_t) \odot \weight_\ell - \nabla_{\weight_{\ell
  + 1}} R(\bftheta_t) \odot \weight_{\ell + 1}
\end{align*} for all parameter values, due to the cyclic property of $\calL$.
Using the expression \eqref{eq::reg_diff_sq} for the latter, the previous
display turns into \begin{align}
  \label{eq::time_diff_bal}
  \dfrac{\rmd}{\rmd t} \Big(\weight_\ell^2(t) - \weight_{\ell + 1}^2(t)\Big) = -
  4 \eta^2 \Diag\big(X\tran X\big) \left(\bigodot_{\substack{r = 1\\ r \neq
  \ell, \ell + 1}}^L \big(\weight_r^2(t) + \eta^2 \cdot \bfone\big)\right) \odot
  \Big(\weight_\ell^2(t) - \weight_{\ell + 1}^2(t)\Big)
\end{align} The vector multiplying the difference of squares $\weight_\ell^2(t)
- \weight_{\ell + 1}^2(t)$ features entirely non-negative entries, being a
product of positive factors. In case $\weight_r(t) = 0$ for every $r \neq \ell,
\ell + 1$, its entries take on their minimal achievable value
\begin{align}
  \label{eq::reg_diff_lower}
  \min_{h = 1, \ldots, d}\ \left(\eta^2 \cdot \Diag\big(X\tran X\big)
  \prod_{\substack{r = 1\\ r \neq \ell, \ell + 1}}^L \big(\weight_r^2(t) +
  \eta^2 \cdot \bfone\big)\right)_{h} \geq \eta^{2 L - 2} \min_{j = 1, \ldots,
  d} \Diag\big(X\tran X\big)_{jj} > 0.
\end{align} Since $\weight_{r, h}(t)^2 \geq 0$, the latter bound holds for all
$t \geq 0$. Together with \eqref{eq::time_diff_bal}, the chain rule, and
$\Diag(X\tran X)_{jj} \geq \sigminp(X\tran X)$ this now implies \begin{align*}
  &\dfrac{\rmd}{\rmd t} \norm[\big]{\weight_\ell^2(t) - \weight_{\ell +
  1}^2(t)}^2_2\\
  =\ &\bfone\tran \left(\dfrac{\rmd}{\rmd t} \Big(\weight_\ell^2(t) -
  \weight_{\ell + 1}^2(t)\Big)^2\right) \\
  =\ &2 \cdot \bfone\tran \left(\Big(\weight_\ell^2(t) - \weight_{\ell +
  1}^2(t)\Big) \odot \dfrac{\rmd}{\rmd t} \Big(\weight_\ell^2(t) - \weight_{\ell
  + 1}^2(t)\Big)\right)\\
  =\ &- 8 \cdot \bfone\tran \left(\left(\eta^2 \cdot \Diag\big(X\tran X\big)
  \bigodot_{\substack{r = 1\\ r \neq \ell, \ell + 1}}^L \big(\weight_r^2(t) +
  \eta^2 \cdot \bfone\big)\right) \odot \Big(\weight_\ell^2(t) - \weight_{\ell +
  1}^2(t)\Big)^2\right)\\
  \leq\ &- 8 \eta^{2 L - 2} \sigminp\big(X\tran X\big) \cdot
  \norm[\big]{\weight_\ell^2(t) - \weight_{\ell + 1}^2(t)}^2_2.
\end{align*} Using Gr\"{o}nwall's Lemma \citep{gronwall_1919}, this differential
inequality admits the estimate \begin{align*}
  \norm[\big]{\weight_\ell^2(t) - \weight_{\ell + 1}^2(t)}^2_2 \leq \exp\Big(- 8
  \eta^{2 L - 2} \sigminp\big(X\tran X\big) \cdot t\Big) \cdot
  \norm[\big]{\weight_\ell^2(0) - \weight_{\ell + 1}^2(0)}^2_2
\end{align*} for all $t \geq 0$. Taking the square root on both sides of the
previous display now completes the proof.

\subsection{Proof of Theorem \ref{thm::conv_abs_disc}}

Throughout this proof, we will write $\nabla \calL_R$ and $\nabla^2 \calL_R$ for
the gradient and Hessian of the loss with respect to the whole parameter
$\bftheta$. By definition, $\bftheta_{k + 1} - \bftheta_k = - \alpha_k \cdot
\nabla \calL(\bftheta_k)$ for every $k$. Using a first order approximation,
there exists $\bfvartheta_k$ on the line segment between $\bftheta_{k + 1}$ and
$\bftheta_k$, such that \begin{equation}
  \label{eq::hess_exp}
  \begin{split}
    \calL_R(\bftheta_{k + 1}) - \calL_R(\bftheta_k) &= \nabla
    \calL_R(\bftheta_k)\tran \big(\bftheta_{k + 1} - \bftheta_k\big) +
    \dfrac{1}{2} \cdot \big(\bftheta_{k + 1} - \bftheta_k\big)\tran \nabla^2
    \calL_R(\bfvartheta_k) \big(\bftheta_{k + 1} - \bftheta_k\big)\\
    &\leq \left(- \alpha_k + \dfrac{1}{2} \alpha_k^2 \cdot \sigmax\big(\nabla^2
    \calL_R(\bfvartheta_k)\big)\right) \cdot \sum_{\ell = 1}^L
    \norm[\big]{\nabla_{\weight_\ell(k)} \calL_R(\bftheta_k)}^2_2.
  \end{split}
\end{equation} Flipping the sign in \eqref{eq::hess_exp}, the strong descent
condition \eqref{eq::str_desc_main} holds if $1 - \alpha_k \cdot
\sigmax\big(\nabla^2 \calL_R(\bftheta_k)\big) / 2$ can be uniformly lower
bounded by some $\delta > 0$. To do so, we require an upper bound on the largest
singular value of $\nabla^2 \calL_R(\bfvartheta_k)$.

\begin{lemma}
  \label{lem::hess_frech_bound}
  Fix $\bftheta = (\weight_1, \ldots, \weight_L)$, then \begin{align*}
    \sigmax\big(\nabla^2 \calL_R(\bftheta)\big) \leq \sqrt{L} \kappa\big(X\tran
    X\big) \cdot \dfrac{7 \sqrt{L} - 2}{\eta^2} \cdot \calL_R(\bftheta).
  \end{align*} 
\end{lemma}

\begin{CustomProof}
  Recall that $\sigmax(\ \cdot\ )$ is both sub-multiplicative and sub-additive
  \citep[Problem 7.3.P16 of][]{horn_johnson_2013}, so it suffices to bound
  $\sigmax(\nabla^2 \calL)$ and $\sigmax(\nabla^2 R)$ separately. Using the
  expression \eqref{eq::hess_loss} for $\nabla^2 \calL$, we compute
  \begin{align}
    \label{eq::hess_loss_sig}
    \nabla^2 \calL(\bftheta) \leq 2 \sigmax\big(X\tran X) \cdot \sigmax^2\big(D
    \bfw(\bftheta)\big) + 2 \cdot \sigmax\left(\sum_{j = 1}^d \big(X\tran \bfY -
    X\tran X \bfw\big)_j \cdot \nabla^2 \bfw(\bftheta)_j\right).
  \end{align} The singular values of a matrix cannot exceed $\norm{A} = \Tr(A
  A\tran)^{1 / 2}$ \citep[Problem 5.6.P56 of][]{horn_johnson_2013}. Together
  with the block structure of $D \weight(\bftheta)$, this implies \begin{align*}
    \sigmax^2\big(D \bfw(\bftheta)\big) \leq \sum_{\ell = 1}^L \norm{D_\ell}^2
    \leq \sum_{\ell = 1}^L\ \norm*{\bigodot_{\substack{m = 1\\ m \neq \ell}}^L
    \weight_m}_2^2 \leq L \cdot \max_{\ell = 1, \ldots, L}\
    \norm*{\bigodot_{\substack{m = 1\\ m \neq \ell}}^L \weight_m}_2^2
  \end{align*} with $D_\ell$ defined in \eqref{eq::jac_diag}. Each map
  $\weight(\bftheta)_j = \weight_{L, j} \cdots \weight_{1, j}$ is linear in the
  weight vector entries and only depends on the $j$\textsuperscript{th}
  coordinate, so the matrix inside the second term of \eqref{eq::hess_loss_sig}
  admits the block-diagonal structure \begin{align*}
    &\sum_{j = 1}^d \big(X\tran \bfY - X\tran X \bfw\big)_j \cdot \nabla^2
    \bfw(\bftheta)_j \\
    =\ &\begin{bmatrix}
      \big(X\tran \bfY - X\tran X \bfw\big)_1 \cdot H_{11}(\bftheta) & &
      \\
      & \ddots & \\
      & & \big(X\tran \bfY - X\tran X \bfw\big)_d \cdot H_{dd}(\bftheta)
    \end{bmatrix},
  \end{align*} with each $(L \times L)$ block $H_{jj}(\bftheta)$ taking the form
  \begin{align*}
    H_{ii}(\bftheta)_{r s} = \begin{cases}
      \prod_{\substack{\ell = 1\\ \ell \neq r, s}}^L \weight_{\ell, i} & \mbox{
      if } r \neq s, \\
      0 & \mbox{ otherwise.}
    \end{cases}
  \end{align*} As $\sigmax(X) \leq \sigmax^{1 / 2}(X\tran X)$ and $\abs{\bfv_i}
  \leq \norm{\bfv}_2$ for any vector, together with the bound $\sigmax(\ \cdot\
  ) \leq \norm{\ \cdot\ }$, we now conclude that \begin{align*}
    \sigmax\left(\sum_{j = 1}^d \big(X\tran \bfY - X\tran X \bfw\big)_j \cdot
    \nabla^2 \bfw(\bftheta)_j\right) &\leq \max_{j = 1, \ldots, d}\
    \abs[\Big]{\big(X\tran \bfY - X\tran X \bfw\big)_j} \cdot
    \norm[\big]{H_{jj}(\bftheta)}\\
    &\leq \sqrt{\sigmax\big(X\tran X\big) \cdot \calL(\bftheta) \cdot
    \sum_{\substack{r, s = 1\\ r\neq s}}^L\ \norm*{\bigodot_{\substack{m = 1\\ m
    \neq r, s}}^L \weight_m}^2_2} \\
    &\leq \sqrt{L^2 \sigmax\big(X\tran X\big) \cdot \calL(\bftheta) \cdot
    \max_{\substack{r, s = 1, \ldots, L\\ r\neq s}}\
    \norm*{\bigodot_{\substack{m = 1\\ m \neq r, s}}^L \weight_m}^2_2}
  \end{align*} Inserting these computations into \eqref{eq::hess_loss_sig}
  yields \begin{equation}
    \label{eq::loss_hess_op_bound}
    \begin{split}
      \sigmax\big(\nabla^2 \calL(\bftheta)\big) &\leq 2 L
      \left(\sigmax\big(X\tran X\big) \cdot \max_{\ell = 1, \ldots, L}\
      \norm*{\bigodot_{\substack{m = 1\\ m \neq \ell}}^L \weight_m}_2^2 \right.
      \\
      &\qquad \qquad \left. + \sqrt{\sigmax\big(X\tran X\big) \cdot
      \calL(\bftheta) \cdot \max_{\substack{r, s = 1, \ldots, L\\ r\neq s}}\
      \norm*{\bigodot_{\substack{m = 1\\ m \neq r, s}}^L \weight_m}^2_2}\right).
    \end{split}
  \end{equation} To proceed, we require a bound on the norm of products over
  subsets of the weight vectors. The specific form of the regularizer $R$ leads
  to generic estimates in terms of $R(\bftheta)$ and $\calL_R(\bftheta)$ that
  will come into use in various situations over the course of the remaining
  proofs. 

  \begin{lemma}
    \label{lem::prod_bound}
    Let $I \subsetneq \{1, \ldots, d\}$ and assume $\min_{i = 1, \ldots, d}
    \Diag(X\tran X)_{ii} > 0$, then the following inequalities hold
    \begin{align*}
      \norm*{\bigodot_{m \in I} \weight_m^2}_2 &\leq
      \max\left\{\norm*{\bigodot_{m \in I} \weight_m}^2_2,\ \norm*{\bigodot_{m
      \in I} \big(\weight_m^2 + \eta^2 \cdot \bfone\big)}_2\right\} \leq
      \dfrac{\sigminp(X\tran X)\inv}{\eta^{2 (L - \# I)}} \cdot R(\bftheta).
    \end{align*} Further, $R$ may be replaced by $\calL_R$ in these bounds.
  \end{lemma}
  
  \begin{CustomProof}
    Sub-multiplicativity directly proves $\norm{\bfv^2}_2 \leq \norm{\bfv}^2_2$
    for any vector $\bfv$. Non-negativity of each squared entry $\weight_{m,
    h}^2$ and $\eta^2$ implies \begin{align*}
      \norm*{\bigodot_{m \in I} \weight_m^2}_2 = \sqrt{\bfone\tran \bigodot_{m
      \in I} \weight_m^4} \leq \sqrt{\bfone\tran \bigodot_{m \in I}
      \big(\weight_m^2 + \eta^2 \cdot \bfone\big)^2} = \norm*{\bigodot_{m \in I}
      \big(\weight_m^2 + \eta^2 \cdot \bfone\big)}_2.
    \end{align*} Expanding the product over $I$, using the triangle inequality,
    and applying the bound $\norm{\Diag(A) \bfv}_2 \geq \sigminp(A) \cdot
    \norm{\bfv}_2$, valid whenever $\min_{j = 1, \ldots, d} A_{jj} > 0$, leads
    to \begin{align*}
      \norm*{\bigodot_{m \in I} \big(\weight_m^2 + \eta^2 \cdot I_d\big)}_2
      &\leq \sum_{J \subset I} \eta^{2 (\# I - \# J)} \cdot \norm*{\bigodot_{m
      \in J} \weight_m}^2_2\\
      &= \dfrac{1}{\eta^{2 (L - \# I)}} \sum_{J \subset I} \eta^{2 (L - \# J)}
      \norm*{\bigodot_{m \in J} \weight_m}^2_2 \leq \dfrac{\sigminp(X\tran
      X)\inv}{\eta^{2 (L - \# I)}} \cdot R(\bftheta),
    \end{align*} where the last inequality follows by comparison with the
    representation \eqref{eq::reg_poly} for $R$. An analogous argument also
    proves \begin{align*}
      \norm*{\bigodot_{m \in I} \weight_m}^2_2 = \bfone\tran \bigodot_{m \in I}
      \weight_m^2 \leq \bfone\tran \bigodot_{m \in I} \big(\weight_m^2 + \eta^2
      \cdot \bfone\big) \leq \dfrac{\sigminp(X\tran X)\inv}{\eta^{2 (L - \# I)}}
      \cdot R(\bftheta).
    \end{align*} For the final statement recall that $\calL_R = \calL + R$, with
    $\calL$ given by a norm and hence non-negative. 
  \end{CustomProof}

  Applying Lemma \ref{lem::prod_bound} to each of the products in
  \eqref{eq::loss_hess_op_bound} and using the AM/GM inequality to bound
  $\sqrt{\calL \cdot R} \leq 2\inv \cdot (\calL + R) = \calL_R / 2$ now leads to
  \begin{align}
    \sigmax\big(\nabla^2 \calL(\bftheta)\big) &\leq \dfrac{2 L \cdot
    \sigmax(X\tran X)}{\eta^2 \cdot \sigminp(X\tran X)} \cdot \calL_R(\bftheta)
    + 2 L \cdot \sqrt{\dfrac{\sigmax(X\tran X)}{\eta^4 \sigminp(X\tran X)} \cdot
    \calL(\bftheta) \cdot R(\bftheta)} \nonumber\\
    &\leq \dfrac{3 L \cdot \kappa(X\tran X)}{\eta^2} \cdot \calL_R(\bftheta),
    \label{eq::loss_diff_op}
  \end{align} where the second inequality follows from $\kappa(X\tran X) \geq
  1$.
  
  Finding an analogous bound for the largest singular value of $\nabla^2 R$
  necessitates some additional preparation. Recall from Lemma
  \ref{lem::grad_exp_cond} that \begin{align*}
    R(\bftheta) &= \bfone\tran \Diag\big(X\tran X\big) \left(\bigodot_{\ell =
    1}^L \big(\weight_\ell^2 + \eta^2 \cdot \bfone\big) - \bigodot_{m = 1}^L
    \weight_m^2\right) = \bfone\tran \Diag\big(X\tran X\big) \bfz(\bftheta),
  \end{align*} where we define the variable transformation $\bfz(\bftheta) =
  \bigodot_{\ell = 1}^L \big(\weight_m^2 + \eta^2 \cdot \bfone\big) -
  \bigodot_{m = 1}^L \weight_m^2$. Left-multiplication by $\bfone\tran
  \Diag(X\tran X)$ computes a linear functional of $\bfz(\bftheta)$, with
  corresponding gradient $\Diag(X\tran X)$ and vanishing Hessian. Write
  $\nabla^2 \bfz(\bftheta)_j$ for the Hessian of the map from $\bftheta$ to the
  $j$\textsuperscript{th} entry of $\bfz(\bftheta)$, then the chain rule implies
  \begin{align*}
    \nabla^2 R(\bftheta) = \sum_{j = 1}^d \Diag\big(X\tran X\big)_{jj} \cdot
    \nabla^2 \bfz(\bftheta)_j.
  \end{align*} By construction, the $j$\textsuperscript{th} entry of
  $\bfz(\bftheta)$ only depends on $(\weight_{1, j}, \ldots, \weight_{L, j})$
  and so \begin{align*}
    \sum_{j = 1}^d \Diag\big(X\tran X\big)_{jj} \cdot \nabla^2 \bfz(\bftheta)_j
    = \begin{bmatrix}
      \Diag\big(X\tran X\big)_{11} \cdot H_{11}(\bfz) & & \\
      & \ddots & \\
      & & \Diag\big(X\tran X\big)_{dd} \cdot H_{dd}(\bfz)
    \end{bmatrix},
  \end{align*} where each $(L \times L)$ block $H_{jj}(\bfz)$ has entries
  \begin{align*}
    H_{jj}(\bftheta)_{r s} = \begin{cases}
      2 \cdot \left(\prod_{\substack{\ell = 1\\ \ell \neq r}}^L
      \big(\weight_{\ell, j}^2 + \eta^2\big) - \prod_{\substack{m = 1\\ m \neq
      r}}^L \weight_{m, j}^2\right) & \mbox{ if } r = s, \\
      4 \cdot \weight_{r, j} \weight_{s, j} \cdot \left(\prod_{\substack{\ell =
      1\\ \ell \neq r, s}}^L \big(\weight_{\ell, j}^2 + \eta^2\big) -
      \prod_{\substack{m = 1\\ m \neq r, s}}^L \weight_{m, j}^2\right) & \mbox{
      if } r \neq s.
    \end{cases}
  \end{align*} The maximal diagonal entry of $X\tran X$ cannot exceed
  $\sigmax(X\tran X)$. Using that $\sigmax(\ \cdot\ ) \leq \norm{\ \cdot\ }$ and
  $\prod_{\ell \in I} \big(\weight_{\ell, j}^2 + \eta^2\big) - \prod_{m \in I}^L
  \weight_{m, j} \leq \prod_{\ell \in I} \big(\weight_{\ell, j}^2 +
  \eta^2\big)$, we then compute \begin{align*}
    \sigmax\big(\nabla^2 R(\bftheta)\big) &\leq \sigmax\big(X\tran X\big) \cdot
    \max_{j = 1, \ldots, d}\ \norm[\big]{H_{jj}(\bfz)}_2\\
    &\leq \sigmax\big(X\tran X\big) \cdot \max_{j = 1, \ldots, d}\ \left(\sqrt{4
    \cdot \sum_{q = 1}^\ell \left(\prod_{\substack{\ell = 1\\ \ell \neq q}}^L
    \big(\weight_{\ell, j}^2 + \eta^2\big) - \prod_{\substack{m = 1\\ m \neq
    q}}^L \weight_{m, j}^2\right)^2}\right.\\
    &\qquad\qquad + \left. \sqrt{16 \cdot \sum_{\substack{r, s = 1\\ r \neq
    s}}^L\ \weight_{r, j}^2 \weight_{s, j}^2 \cdot \left(\prod_{\substack{\ell =
    1\\ \ell \neq r, s}}^L \big(\weight_{\ell, j}^2 + \eta^2\big) -
    \prod_{\substack{m = 1\\ m \neq r, s}}^L \weight_{m, j}^2\right)^2}\right)\\
    &\leq \sigmax\big(X\tran X\big) \cdot \left(2 \cdot \sqrt{\sum_{q = 1}^\ell\
    \norm*{\bigodot_{\substack{\ell = 1\\ \ell \neq q}}^L \big(\weight_{\ell}^2
    + \eta^2 \cdot \bfone\big)}_2^2} \right. \\
    &\qquad\qquad\qquad \left. +\ 4 \cdot \sqrt{\sum_{\substack{r, s = 1\\ r
    \neq s}}^L\ \norm*{\weight_{r} \odot \weight_{s} \bigodot_{\substack{\ell =
    1\\ \ell \neq r, s}}^L \big(\weight_{\ell}^2 + \eta^2 \cdot
    \bfone\big)}_2^2}\right) \\
    &\leq 2 \sigmax\big(X\tran X\big) \sqrt{L} \big(2 \sqrt{L} + 1\big) \cdot
    \max_{m = 1, \ldots L}\ \norm*{\bigodot_{\substack{\ell = 1\\ \ell \neq
    m}}^L \big(\weight_{\ell}^2 + \eta^2 \cdot \bfone\big)}_2,
  \end{align*} where the last inequality follows from the AM/GM inequality via
  $\abs[\big]{\weight_r \odot \weight_s} \leq \big(\weight_r^2 +
  \weight_s^2\big) / 2$, which is strictly bounded by $\big(\weight_r^2 + \eta^2
  \cdot \bfone + \weight_s^2 + \eta^2 \cdot \bfone\big) / 2$. We can now apply
  Lemma \ref{lem::prod_bound} to receive the final bound \begin{align}
    \label{eq::reg_diff_op}
    \sigmax\big(\nabla^2 R(\bftheta)\big) \leq \dfrac{\sqrt{L} \big(4 \sqrt{L} +
    2\big) \cdot \kappa(X\tran X)}{\eta^2} \cdot \calL_R(\bftheta).
  \end{align}
  
  To complete the proof of the lemma, it now suffices to combine
  \eqref{eq::loss_diff_op} and \eqref{eq::reg_diff_op} into an estimate for
  $\sigmax\big(\nabla^2 \calL_R(\bftheta)\big)$.
\end{CustomProof}

Employing the estimate from Lemma \ref{lem::hess_frech_bound} in
\eqref{eq::hess_exp} now shows that the strong descent condition
\eqref{eq::str_desc_main} holds whenever \begin{align*}
  1 - \dfrac{\alpha_k}{2} \cdot \sigmax\big(\nabla^2 \calL_R(\bfvartheta_k)\big)
  \geq 1 - \dfrac{\alpha_k}{2} \cdot \sqrt{L} \kappa\big(X\tran X\big) \cdot
  \dfrac{7 \sqrt{L} + 2}{\eta^2} \cdot \calL_R(\bfvartheta_k) \geq \delta.
\end{align*} Lemma \ref{lem::gf_coerc} implies coercivity of the loss $\calL_R$,
which must in turn exhibit compact sub-level sets. As shown in the proof of
Proposition 1 in \cite{letcher_2021}, the condition $\alpha_k < 2 /
\sigmax\big(\nabla^2 \calL_R(\bftheta_k)\big)$ then already ensures
$\calL_R(\bfvartheta_k) \leq \calL_R(\bftheta_k)$ for all $\bfvartheta_k$ in the
line segment between $\bftheta_k$ and $\bftheta_{k + 1}$. This implies the first
part of Theorem \ref{thm::conv_abs_disc}, with the second part following via
induction on $k \geq K$ and the same argument as in Lemma \ref{lem::gf_coerc}.

\subsection{Proof of Theorem \ref{thm::grad_desc_bal}}

The proof follows along similar lines as the proof of Theorem
\ref{thm::grad_flow_bal}, while accounting for extra terms quantifying the
discretization error. Using the definition of the gradient descent recursion
\eqref{eq::gd_def_no_noise}, \begin{equation}
  \label{eq::sq_diff_disc}
  \begin{split}
    &\weight_{\ell}^2(k + 1) - \weight_{\ell + 1}^2(k + 1)\\
    =\ &\bigg(\weight_{\ell}(k) - \alpha_k \cdot \nabla_{\weight_{\ell}(k)}
    \calL_R(\bftheta_k)\bigg)^2 - \bigg(\weight_{\ell + 1}(k) - \alpha_k
    \nabla_{\weight_{\ell + 1}(k)} \calL_R(\bftheta_k)\bigg)^2\\
    =\ &\weight_{\ell}^2(k) - \weight_{\ell + 1}^2(k) - 2 \alpha_k \cdot
    \Bigg(\nabla_{\weight_{\ell}(k)} \calL_R(\bftheta_k) \odot \weight_{\ell}(k)
    - \nabla_{\weight_{\ell + 1}(k)} \calL_R(\bftheta) \odot \weight_{\ell +
    1}(k)\Bigg)\\
    &\qquad \qquad \qquad + \alpha_k^2 \cdot
    \bigg(\nabla_{\weight_{\ell}(k)} \calL_R(\bftheta_k)^2 -
    \nabla_{\weight_{\ell + 1}(k)} \calL_R(\bftheta_k)^2\bigg)\\
    =\ &\left(\bfone - 4 \alpha_k \eta^2 \cdot \Diag\big(X\tran X\big)
    \bigodot_{\substack{r = 1\\ r \neq \ell, \ell + 1}}^L \big(\weight_r^2(k) +
    \eta^2 \cdot \bfone\big)\right) \odot \Big(\weight_{\ell}^2(k) -
    \weight_{\ell + 1}^2(k)\Big)\\
    &\qquad \qquad \qquad + \alpha_k^2 \cdot \bigg(\nabla_{\weight_{\ell}(k)}
    \calL_R(\bftheta_k)^2 - \nabla_{\weight_{\ell + 1}(k)}
    \calL_R(\bftheta_k)^2\bigg),
  \end{split}
\end{equation} where the last equality follows from the computations in
\eqref{eq::reg_cons_eq} and \eqref{eq::reg_diff_sq}. Noting that $\max_{i = 1,
\ldots, d} \abs{\bfv}_i \leq \norm{\bfv}_2$, Lemma \ref{lem::prod_bound} implies
\begin{align*}
  &\max_{i = 1, \ldots, d}\ \left(\eta^2 \cdot \Diag\big(X\tran X\big)
  \bigodot_{\substack{r = 1\\ r \neq \ell, \ell + 1}}^L \big(\weight_r^2(k) +
  \eta^2 \cdot \bfone\big)\right)_i\\
  \leq\ &\eta^2 \sigmax\big(X\tran X\big) \cdot \norm*{\bigodot_{\substack{r =
  1\\ r \neq \ell, \ell + 1}}^L \big(\weight_r^2(k) + \eta^2 \cdot
  \bfone\big)}_2 \leq \dfrac{\kappa(X\tran X)}{\eta^2} \cdot
  \calL_R(\bftheta_k),
\end{align*} By assumption $4 \alpha_k \eta^{-2} \kappa(X\tran X) \cdot
\calL_R(\bftheta_k) < 1$, which together with the previous display and
\eqref{eq::reg_diff_lower} implies \begin{align*}
  0 \leq \max_{i = 1, \ldots, d} \left(\bfone - 4 \alpha_k \eta^2 \cdot
  \Diag\big(X\tran X\big) \bigodot_{\substack{r = 1\\ r \neq \ell, \ell + 1}}^L
  \big(\weight_r^2(k) + \eta^2 \cdot \bfone\big)\right)_i \leq 1 - 4 \alpha_k
  \eta^{2 L - 2} \sigminp(X\tran X)
\end{align*} Combining this computation with \eqref{eq::sq_diff_disc} and the
triangle inequality now leads to \begin{equation}
  \label{eq::bal_rec_disc}
  \begin{split}
    \norm[\big]{\weight_{\ell}^2(k + 1) - \weight_{\ell + 1}^2(k + 1)}_2 &\leq
    \Big(1 - 4 \alpha_k \eta^{2 L - 2} \sigminp(X\tran X)\Big) \cdot
    \norm[\big]{\weight_{\ell}^2(k) - \weight_{\ell + 1}^2(k)}_2 \\
    &\qquad + \alpha_k^2 \cdot \norm[\big]{\nabla_{\weight_{\ell}(k)}
    \calL_R(\bftheta_k)^2 - \nabla_{\weight_{\ell + 1}(k)}
    \calL_R(\bftheta_k)^2}_2.
  \end{split}
\end{equation} The latter gives a discrete-time analog of
\eqref{eq::time_diff_bal}. In comparison, the difference of squared gradients
controls the discretization error, with additional attenuation by $\alpha_k^2$.
Expanding the squares, this error may be split into three terms \begin{align*}
  \nabla_{\weight_{\ell}(k)} \calL_R(\bftheta_k)^2 - \nabla_{\weight_{\ell +
  1}(k)} \calL_R(\bftheta_k)^2 = \Delta_1(k) + \Delta_2(k) + \Delta_3(k),
\end{align*} where \begin{align*}
  \Delta_1(k) &= \nabla_{\weight_{\ell}(k)} \calL(\bftheta_k)^2 -
  \nabla_{\weight_{\ell + 1}(k)} \calL(\bftheta_k)^2\\
  \Delta_2(k) &= 2 \cdot \nabla_{\weight_{\ell}(k)} \calL(\bftheta_k) \odot
  \nabla_{\weight_{\ell}(k)} R(\bftheta_k) - 2 \cdot \nabla_{\weight_{\ell +
  1}(k)} \calL(\bftheta_k) \odot \nabla_{\weight_{\ell + 1}(k)} R(\bftheta)\\
  \Delta_3(k) &= \nabla_{\weight_{\ell}(k)} \calL(\bftheta_k)^2 -
  \nabla_{\weight_{\ell + 1}(k)} \calL(\bftheta_k)^2.
\end{align*} We will now show that the norm of each $\Delta_i(k)$ cannot exceed
some constant multiple of $\norm[\big]{\weight_{\ell}^2(k) - \weight_{\ell +
1}^2(k)}$. To this end, we inspect the expressions for the gradients of $\calL$
and $R$ computed in \eqref{eq::diff_loss} and \eqref{eq::diff_reg} and find
bounds for the involved norms via Lemma \ref{lem::prod_bound}.

Recall that $\bfw(k) = \weight_L(k) \odot \cdots \odot \weight_1(k)$. The
Cauchy-Schwarz inequality implies $\norm{\bfu \odot \bfv}_2 \leq \norm{\bfu}_2
\cdot \norm{\bfv}_2$. Using the expression \eqref{eq::diff_loss}, as well as
Lemma \ref{lem::prod_bound}, the norm of $\Delta_1(k)$ then admits the estimate
\begin{align}
  \norm{\Delta_1(k)}_2 &= 4 \cdot \norm*{\big(X\tran \bfY - X\tran X
  \weight(k)\big)^2 \odot \left(\bigodot_{\substack{r = 1\\ r \neq \ell}}^L
  \weight_r^2(k) - \bigodot_{\substack{s = 1\\ s \neq \ell + 1}}^L
  \weight_s^2(k)\right)}_2 \nonumber \\
  &\leq 4 \sigmax\big(X\tran X\big) \cdot \calL(\bftheta_k) \cdot
  \norm*{\bigodot_{\substack{m = 1\\ m \neq \ell, \ell + 1}}^L \weight_m^2(k)}_2
  \cdot \norm[\big]{\weight_{\ell}^2(k) - \weight_{\ell + 1}^2(k)}_2 \nonumber\\
  &\leq \dfrac{4 \kappa(X\tran X)}{\eta^4} \cdot \calL_R(\bftheta_k)^2 \cdot
  \norm[\big]{\weight_{\ell}^2(k) - \weight_{\ell + 1}^2(k)}_2.
  \label{eq::delta_one_bound}
\end{align} 

We move on to bound $\Delta_2(k)$. Multiplying the expressions in
\eqref{eq::diff_loss} and \eqref{eq::diff_reg}, commutativity of the
element-wise product yields \begin{align*}
  &\nabla_{\weight_{\ell}(k)} \calL(\bftheta_k) \odot \nabla_{\weight_{\ell}(k)}
  R(\bftheta_k)\\
  =\ &2 \cdot \Big(\nabla_{\weight(k)} \calL(\bftheta_k)\Big) \odot \weight(k)
  \odot \Diag(X\tran X) \left(\bigodot_{\substack{r = 1\\ r \neq \ell}}^L
  \big(\weight_r^2(k) + \eta^2 \cdot \bfone\big) - \bigodot_{\substack{s = 1\\ s
  \neq \ell}}^L \weight_s^2(k)\right).
\end{align*} The latter expression depends on $\ell$ only through its exclusion
in the products over $r$ and $s$. Further, we notice that \begin{align*}
  &\left(\bigodot_{\substack{r = 1\\ r \neq \ell}}^L \big(\weight_r^2(k) +
  \eta^2 \cdot \bfone\big) - \bigodot_{\substack{s = 1\\ s \neq \ell}}^L
  \weight_s^2(k)\right) - \left(\bigodot_{\substack{r = 1\\ r \neq \ell + 1}}^L
  \big(\weight_r^2(k) + \eta^2 \cdot \bfone\big) - \prod_{\substack{s = 1\\ s
  \neq \ell + 1}}^L \weight_s^2(k)\right)\\
  =\ &\left(\bigodot_{\substack{r = 1\\ r \neq \ell, \ell + 1}}^L
  \big(\weight_r^2(k) + \eta^2 \cdot \bfone\big) - \bigodot_{\substack{s = 1\\ s
  \neq \ell, \ell + 1}}^L \weight_s^2(k)\right) \Big(\weight_\ell^2(k) -
  \weight_{\ell + 1}^2(k)\Big).
\end{align*} For any $\ell = 1, \ldots, L - 1$, Lemma \ref{lem::prod_bound} and
a point-wise application of the AM-GM inequality imply \begin{align*}
  &\norm*{\weight_{\ell + 1}(k) \odot \weight_\ell(k) \odot
  \left(\bigodot_{\substack{r = 1\\ r \neq \ell, \ell + 1}}^L
  \big(\weight_r^2(k) + \eta^2 \cdot \bfone\big) - \bigodot_{\substack{s = 1\\ s
  \neq \ell, \ell + 1}}^L \weight_s^2(k)\right)}_2\\
  \leq\ &\max_{m = 1, \ldots, L}\ \norm*{\bigodot_{\substack{r = 1\\ r \neq
  m}}^L \big(\weight_r^2(k) + \eta^2 \cdot \bfone\big)}_2 \leq \dfrac{1}{\eta^2
  \cdot \sigminp(X\tran X)} \cdot \calL_R(\bftheta_k).
\end{align*} Combining these computations, as well as applying once more Lemma
\ref{lem::prod_bound} and the AM/GM inequality, now yields the estimate
\begin{align}
  \norm{\Delta_2(k)}_2 &\leq \dfrac{4 \kappa(X\tran X)}{\eta^2} \cdot
  \norm[\big]{\nabla_{\weight(k)} \calL(\bftheta_k)}_2 \cdot
  \norm*{\bigodot_{\substack{s = 1\\ s \neq \ell, \ell + 1}}^L \weight_s(k)}_2
  \cdot \calL_R(\bftheta_k) \cdot \norm[\big]{\weight_\ell^2(k) - \weight_{\ell
  + 1}^2(k)} \nonumber \\
  &\leq \dfrac{8 \kappa(X\tran X)}{\eta^2} \sqrt{\dfrac{\kappa(X\tran
  X)}{\eta^4} \cdot \calL(\bftheta_k) \cdot R(\bftheta)} \cdot
  \calL_R(\bftheta_k) \cdot \norm[\big]{\weight_\ell^2(k) - \weight_{\ell +
  1}^2(k)} \nonumber \\
  \leq\ &\dfrac{4 \kappa(X\tran X)^{3 / 2}}{\eta^4} \cdot \calL_R(\bftheta_k)^2
  \cdot \norm[\big]{\weight_\ell^2(k) - \weight_{\ell + 1}^2(k)}.
  \label{eq::delta_two_bound}
\end{align}

Lastly, we bound the norm of $\Delta_3(k)$. To this end, observe via
\eqref{eq::diff_reg} and commutativity of the elemet-wise product that
\begin{align*}
  \nabla_{\weight_\ell} R(\bftheta_k)^2 = 4 \cdot \Diag\big(X\tran X\big)^2
  \weight_\ell^2(k) \odot \left(\bigodot_{\substack{r = 1\\ r \neq \ell}}^L
  \big(\weight_r^2(k) + \eta^2 \cdot \bfone\big) - \bigodot_{\substack{s = 1\\ s
  \neq \ell}}^L \weight_s^2(k)\right)^2.
\end{align*} Switching the roles of $\ell$ and $\ell + 1$ yields an analogous
formula for the squared gradient with respect to $\weight_{\ell + 1}(k)$.
Subtracting the squared gradients and canceling like terms results in
\begin{equation}
  \label{eq::delta_three_exp}
  \begin{split}
    &\dfrac{1}{4} \cdot \Delta_3(k)\\
    =\ &\eta^4 \cdot \Diag\big(X\tran X\big)^2 \left(\bigodot_{\substack{r = 1\\
    r \neq \ell, \ell + 1}}^L \big(\weight_r^2(k) + \eta^2 \cdot
    \bfone\big)\right)^2 \odot \Big(\weight_{\ell}^2(k) - \weight_{\ell +
    1}^2(k)\Big)\\
    &- \Diag\big(X\tran X\big)^2 \weight_\ell^2(k) \odot \weight_{\ell + 1}^2(k)
    \odot \left(\bigodot_{\substack{r = 1\\ r \neq \ell, \ell + 1}}^L
    \big(\weight_r^2(k) + \eta^2 \cdot \bfone\big) - \bigodot_{\substack{s = 1\\
    s \neq \ell, \ell + 1}}^L \weight_s^2(k)\right)^2\\ 
    &\qquad \qquad \qquad \qquad \qquad \qquad \qquad \qquad \odot
    \Big(\weight_{\ell}^2(k) - \weight_{\ell + 1}^2(k)\Big).
  \end{split}
\end{equation} To complete our analysis of $\Delta_3(k)$, we must bound the
norms of both vectors multiplying $\weight_{\ell}^2(k) - \weight_{\ell + 1}^2$
in the previous display. Using Lemma \ref{lem::prod_bound}, \begin{align*}
  \eta^4 \cdot \norm*{\Diag(X\tran X)^2 \left(\bigodot_{\substack{r = 1\\ r \neq
  \ell, \ell + 1}}^L \big(\weight_r^2(k) + \eta^2 \cdot \bfone\big)\right)^2}_2
  \leq \dfrac{\kappa(X\tran X)^2}{\eta^4} \cdot \calL_R(\bftheta_k)^2.
\end{align*} Point-wise application of the AM/GM inequality and Lemma
\ref{lem::prod_bound}, further imply \begin{align*}
  &\norm*{\Diag\big(X\tran X\big)^2 \weight_\ell^2(k) \odot \weight_{\ell +
  1}^2(k) \odot \left(\bigodot_{\substack{r = 1\\ r \neq \ell, \ell + 1}}^L
  \big(\weight_r^2(k) + \eta^2 \cdot \bfone\big) - \bigodot_{\substack{s = 1\\ s
  \neq \ell, \ell + 1}}^L \weight_s^2(k)\right)^2}_2\\
  \leq\ &\dfrac{\sigmax(X\tran X)^2}{2} \cdot \norm*{\Big(\weight_{\ell}^4(k) +
  \weight_{\ell + 1}^4(k)\Big) \odot \left(\prod_{\substack{r = 1\\ r \neq \ell,
  \ell + 1}}^L \big(\weight_r^2(k) + \eta^2 \cdot \bfone\big)\right)^2}_2\\ 
  \leq\ &\sigmax\big(X\tran X\big)^2 \cdot \max_{m = \ell, \ell + 1}\
  \norm*{\bigodot_{\substack{r = 1\\ r \neq m}}^L \big(\weight_r^2(k) + \eta^2
  \cdot \bfone\big)}^2_2 \leq \dfrac{\kappa(X\tran X)^2}{\eta^4} \cdot
  \calL_R(\bftheta_k).
\end{align*} Going back to \eqref{eq::delta_three_exp} with these computations
in hand, we now conclude that \begin{align}
  \norm{\Delta_3(k)}_2 &\leq \dfrac{8 \kappa(X\tran X)^2}{\eta^4} \cdot
  \calL_R(\bftheta_k)^2 \cdot \norm[\big]{\weight_\ell(k) - \weight_{\ell +
  1}(k)}_2. \label{eq::delta_three_bound}  
\end{align}

Together with the triangle inequality, the estimates
\eqref{eq::delta_one_bound}, \eqref{eq::delta_two_bound}, and
\eqref{eq::delta_three_bound} provide a bound for the difference of squared
gradients in \eqref{eq::bal_rec_disc}, so that \begin{align*}
  &\norm[\big]{\weight_{\ell}^2(k + 1) - \weight_{\ell + 1}^2(k + 1)}_2\\
  &\leq \left(1 - 4 \alpha_k \eta^{2 L - 2} + \alpha_k^2 \cdot \dfrac{16
  \kappa(X\tran X)^2}{\eta^4} \cdot \calL_R(\bftheta_k)^2\right) \cdot
  \norm[\big]{\weight_{\ell}^2(k) - \weight_{\ell + 1}^2(k)}_2.
\end{align*} Due to the assumption on $\alpha_k$, \begin{align*}
  \alpha_k^2 \cdot \dfrac{16 \kappa(X\tran X)^2}{\eta^4} \cdot
  \calL_R(\bftheta_k)^2 \leq 3 \eta^{2 L - 2},
\end{align*} which now implies the desired final bound \begin{align*}
  \norm[\big]{\weight_{\ell}^2(k + 1) - \weight_{\ell + 1}^2(k + 1)}_2 \leq
  \Big(1 - \alpha_k \eta^{2 L - 2}\Big) \cdot \norm[\big]{\weight_{\ell}^2(k) -
  \weight_{\ell + 1}^2(k)}_2.
\end{align*}

\subsection{Proof of Theorem \ref{thm::proj_sam_conv}}

The result follows directly from Theorem \ref{thm::proj_grad_conv} and Lemma
\ref{lem::proj_grad_conv_ball}, provided that their requirements are met. We
will first show that Theorem \ref{thm::proj_grad_conv} applies. By Lemma
\ref{lem::grad_exp_cond}, the marginalized loss $\calL_R$ is polynomial in the
entries of $X, \bfY, \bftheta$ and the noise level $\eta^2$. Consequently,
$\calL_R$ must be locally Lipschitz continuous and its graph is representable as
an intersection of polynomial inequalities. As we project the weight matrices
onto a ball in $\R^{L d}$, this proves assumptions (a) and (b) of Theorem
\ref{thm::proj_grad_conv}. Assumption (c) on the step-sizes $\alpha_k$ is
already contained in the requirements of Theorem \ref{thm::proj_sam_conv}, so it
remains to show that Assumption (d) of Theorem \ref{thm::proj_grad_conv} holds.

Write $\bfzeta(\bftheta) = \nabla \calL_R(\bftheta) - \nabla \wtL(\bftheta)$ for
the fluctuation of the stochastic gradient fluctuations. Fix a parameter value
$\bftheta$, then the first part of Assumption (d) in Theorem
\ref{thm::proj_grad_conv} states that marginalizing $\bfzeta$ over $\bfxi$ and
$(\bfX, Y)$ must yield zero. Indeed, Lemma \ref{lem::grad_exp_cond} proves that
$\E[\nabla \wtL] = \nabla \calL_R$ for any fixed parameter value, meaning
\begin{align*}
  \E\Big[\bfzeta(\bftheta) \mid \bftheta\Big] = \E\Big[\nabla \wtL(\bftheta) -
  \nabla \calL_R(\bftheta) \mid \bftheta\Big] = \bfzero.
\end{align*}

We must now show that the second and third part of Assumption (d), concerning
the expected gradient norms, hold. As a function of the  stochastic variables
$\bfxi_k$, $\bfX_k$, and $Y_k$, the randomized loss $\wtL_k(\bftheta)$ is
polynomial, with coefficients given by deterministic polynomials in the
parameter $\bftheta$. Consequently, there exist polynomials $\psi \in \R[x_1]$
and $\phi \in \R[x_1, x_2, x_3]$, both deterministic and independent of the
parameters and stochastic variables, such that \begin{align*}
  \norm[\big]{\wtL(\bftheta)}_2 \leq \psi\big(\norm{\bftheta}_2\big) \cdot
  \phi\big(\norm{\bfxi}_2, \norm{\bfX}_2, \abs{Y}\big).
\end{align*} Both the normal and uniform distribution admit finite moments of
all orders, proving that the moments of $\norm[\big]{\wtL(\bftheta)}_2$ take the
form $C(m) \cdot \psi\big(\norm{\bftheta}_2\big)^m$ for some constant $C(m)$
that depends on the corresponding moment of $\varphi$. This gives the desired
function $p(\bftheta)$ for the second part of Assumption (d) in Theorem
\ref{thm::proj_grad_conv}. In similar fashion, if $\bfvartheta_k$, $k \geq 0$
denotes any sequence of network parameters that converges as an element of
$\R^{L d}$, then \begin{align*}
  \E\Big[\sup_{k \geq 0} \norm[\big]{\nabla \wtL(\bfvartheta_k)}_2\Big] \leq
  \sup_{k \geq 0} \psi\big(\norm{\bfvartheta_k}_2\big) \cdot
  \E\Big[\phi\big(\norm{\bfxi}_2, \norm{\bfX}_2, \abs{Y}\big)\Big] < \infty,
\end{align*} which shows that the third part of Assumption (d) holds. This
concludes the first part of the proof; all requirements of Theorem
\ref{thm::proj_grad_conv} are met.

We move on to prove that Lemma \ref{lem::proj_grad_conv_ball} can be applied.
Using the expressions \eqref{eq::diff_loss} and \eqref{eq::diff_reg} for $\nabla
\calL$ and $\nabla R$, as well as treating $\bftheta$ as an element of $\R^{L
d}$, \begin{align*}
  \nabla \calL(\bftheta)\tran \bftheta &= - 2 L \cdot \sum_{i = 1}^d \Big(X\tran
  \big(\bfY - X \weight\big) \odot \weight\Big)_i \\
  &= 2 L \cdot \big(X \bfw\big)\tran \big(\bfY - X \bfw\big) = 2 L \cdot
  \Big(\norm{X \bfw}_2^2 - (X \bfw)\tran \bfY\Big) \\
  \nabla R(\bftheta)\tran \bftheta &= 2 \cdot \sum_{\ell = 1}^L \bfone\tran
  \Diag(X\tran X) \weight_\ell^2 \odot \left(\bigodot_{\substack{r = 1\\ r \neq
  \ell}}^L \big(\weight_r^2 + \eta^2 \cdot I_d\big) - \bigodot_{s = 1}^L
  \weight_s^2\right).
\end{align*} Expanding the product over $r$, note that the latter contains the
weighted norm $2 \eta^{2 L - 2} \cdot \norm{\Diag(X\tran X)^{1 / 2}
\weight_\ell}^2_2 = 2 \eta^{2 L - 2} \cdot \norm{\Diag(X\tran X)^{1 / 2}
\bftheta}_2^2$. Further, every expression in the remaining sum is non-negative,
which together with the Cauchy-Schwarz inequality and the assumed
non-singularity of $\Diag(X\tran X)$ implies \begin{equation}
  \label{eq::iprod_est}
  \dfrac{1}{2} \cdot \nabla \calL_R(\bftheta)\tran \weight \geq L \cdot \norm{X
  \bfw}_2^2 - L \norm{\bfY_2} \cdot \norm{X \bfw}_2 + \eta^{2 L - 2}
  \sigminp\big(X\tran X\big) \cdot \norm{\bftheta}_2. 
\end{equation} We must now pick a radius $r^2 = \sum_{\ell = 1}^L
\norm{\weight_\ell}^2$ for the projection map, such that Assumption (b) of Lemma
\ref{lem::proj_grad_conv_ball} holds. As a function of $z = \norm{X \weight}_2$,
the previous display takes the quadratic form $p(z) = L z^2 - L \norm{\bfY}_2 z
+ \eta^{2 L - 2} \sigminp(X\tran X) r^2$. Over the sphere of fixed radius $r >
0$ in $\R^{Ld}$, the magnitude of $\norm{X \weight}_2$ may take on any value in
some interval $[0, c_1]$, where $c_1 > 0$ depends only on $X$, $r$, $L$, and
$d$. If \begin{align*}
  r > \dfrac{\sqrt{L}}{2 \eta^{L - 1} \sqrt{\sigminp(X\tran X)}} \cdot
  \norm{\bfY}_2,
\end{align*} then the discriminant of $p$ is strictly negative. As $p$ features
a positive leading coefficient, this implies $p(z) > c_2$ over $z \in \ggR$, for
some $c_2 > 0$. In turn, $c_2$ bounds \eqref{eq::iprod_est} away from zero,
meaning the projected S-SAM recursion \eqref{eq::proj_sam} satisfies both
requirements of Lemma \ref{lem::proj_grad_conv_ball} with $c = 2
c_2$ and any $r_1 \geq r$, thereby completing the proof.

\section{Proofs for Section \ref{sec::full_lnn}}
\label{sec::full_lnn_proof}

\subsection{Proof of Lemma \ref{lem::pac_bound}}

The bound follows from Corollary 17 of \cite{chugg_wang_et_al_2023}; for
completeness we provide the steps. To start, recall the definition of the
Kullback-Leibler divergence $\KL(\rho, \nu) = \E_\rho\big[\log(\rmd \rho / \rmd
\nu)\big]$ for measures $\rho$ and $\nu$, with $\rho$ absolutely continuous with
respect to $\nu$ and $\rmd \rho / \rmd \nu$ denoting the Radon-Nikodym
derivative. Fix $\delta \in (0, 1)$, a distribution $\nu$ on the parameters
$\bftheta = (W_1, \ldots, W_L)$ that is independent of $(\bfX, \bfY)$, and
$\lambda_i > 0$ for each $i \in \{1, \ldots, n\}$. By definition, $\calL$ is
quadratic in each data point $\bfX_i$ and $\bfY_i$, so the assumption on the
moments of $(\bfX, \bfY)$ ensures that $\calL(\ \cdot\ )^2$ has finite
expectation for any fixed parameter value. Applying Corollary 17 of
\cite{chugg_wang_et_al_2023} with $Z_i = (\bfX_i, \bfY_i)$ and $f_i(Z_i,
\bftheta) = \norm{\bfY_i - W \bfX_i}^2_2$, with probability $1 - \delta$ over the
draw of the data $(\bfX_1, \bfY_1), \ldots, (\bfX_n, \bfY_n)$, \begin{align*}
  &\sum_{i = 1}^n \lambda_i \cdot \E_\rho\Big[\E_{(\bfX, \bfY)}\big[\norm{\bfY_i
  - W \bfX_i}^2_2\big] - \norm{\bfY_i - W \bfX_i}^2_2\Big]\\
  \leq\ &\sum_{i = 1}^{n} \dfrac{\lambda_i^2}{2} \cdot \E_\rho\Big[\E_{(\bfX,
  \bfY)}\big[\norm{\bfY_i - W \bfX_i}^4_2\big]\Big] + \KL(\rho, \nu) -
  \log(\delta)
\end{align*} for any (possibly data-dependent) distribution $\rho$ on $\bftheta$
that is absolutely continuous with respect to $\nu$. The tuning parameters
$\lambda_i$ may be chosen arbitrarily, as long as they are independent of the
observed data, so we may fix a constant $\lambda_i = \lambda$ to be determined
later. Noting that the data is sampled i.i.d., dividing both sides by $n
\lambda$, and re-arranging leads to the bound \begin{equation}
  \label{eq::pac_bound_post}
  \begin{split}
    \E_\rho\Big[\E_{(\bfX, \bfY)}\big[\calL(\bftheta)\big]\Big] \leq
    \E_\rho\big[\calL(\bftheta)\big] + \dfrac{\KL(\rho, \nu) - \log(\delta)}{n
    \lambda} + \dfrac{\lambda}{2} \cdot \E_\rho\Big[\E_{(\bfX,
    \bfY)}\big[\calL(\bftheta)^2\big]\Big].
  \end{split}
\end{equation} Fix an estimate $\bftheta$ for the parameters. We set the prior
$\nu$ on the parameter values such that each entry of the weight matrices is
independently $\calN(0, \eta^2)$-distributed. For the PAC Bayes posterior
$\rho$, we choose the distribution of $\wtmweight_{\ell, ij} = W_{\ell, ij} +
\xi_{\ell, ij}$ with each $\xi_{\ell, ij} \sim \calN(0, \eta^2)$ independent of
$\bfX$ and $\bfY$. The prior and posterior feature the same diagonal covariance
matrix, so the Kullback-Leibler divergence between them simplifies to
\begin{align*}
  \KL(\rho, \nu) = \dfrac{1}{2 \eta^2} \cdot \sum_{\ell = 1}^L
  \norm{W_\ell}^2,
\end{align*} see also Example 2.2 in \cite{alquier_2024}. Together with
independence of the $\xi_{\ell, ij}$, applying Jensen's Inequality to the convex
function $\norm{\ \cdot \ }^2$ now implies \begin{align*}
  \E_\rho\Big[\E_{(\bfX, \bfY)}\big[\calL(\bftheta)\big]\Big] =
  \E\Big[\norm[\big]{Y - \wtmweight X}^2\Big] \geq
  \E\Bigg[\norm[\bigg]{\E\Big[Y - \wtmweight X \bigmid X, Y,
  \bftheta\Big]}^2\Bigg] = \E\big[\calL(\bftheta)\big].
\end{align*} Inserting these computations into \eqref{eq::pac_bound_post}, we
now arrive at the bound \begin{align*}
  \E\big[\calL(\bftheta)\big] \leq \E_{\rho}\big[\calL(\bftheta)\big] +
  \dfrac{2\inv \eta^{- 2} \cdot \sum_{\ell = 1}^L \norm{W_\ell}^2 -
  \log(\delta)}{n \lambda} + \dfrac{\lambda}{2} \cdot
  \E\big[\wtL(\bftheta)^2\big].
\end{align*} To complete the proof, we choose $\lambda = n^{- 1 / 2}$ and
subtracting the empirical risk $\calL(\bftheta)$ from both sides.

\subsection{Proof of Theorem \ref{thm::gen_weight_bal}}

We start with an auxiliary lemma that is known, but stated and proved for
self-containment of the exposition.

\begin{lemma}
  \label{lem::matrix_chain_rule}
  Given positive integers $p$, $q$, $r$, and $s$, consider a $(p \times
  q)$-matrix $A$, a $(q \times r)$-matrix $B$, and an $(r \times s)$-matrix $C$.
  For differentiable $f : \R^{p\times s} \to \R$, \begin{align*}
	  \nabla_B f(ABC) = A\tran \nabla f(ABC) C\tran.
  \end{align*} Consequently, $U\tran \nabla_U f(U V) = \nabla_V f(V) V\tran$ for
  any matrices $U$ and $V$ of suitable size.
\end{lemma}

\begin{CustomProof}
  We verify the first identity entry-wise; for arbitrary indices $i$ and $j$ the
  chain rule implies \begin{align*}
    \nabla_B f(ABC)_{ij} & = \dfrac{\partial}{\partial B_{ij}} f(ABC) \\
    &= \sum_{k = 1}^{p} \sum_{\ell = 1}^s \dfrac{\partial}{\partial (A B
    C)_{kl}} f(ABC) \cdot \dfrac{\partial}{\partial B_{ij}} (ABC)_{k \ell} \\
    &= \sum_{k = 1}^{p} \sum_{\ell = 1}^s \nabla f(ABC)_{k \ell} \cdot
    \dfrac{\partial}{\partial B_{ij}} \sum_{m = 1}^{q} \sum_{n = 1}^{r} A_{k m}
    B_{m n} C_{n \ell} \\
    & = \sum_{k = 1}^{p} \sum_{\ell = 1}^s \nabla f(ABC)_{k \ell} \cdot A_{k i}
    C_{j \ell} \\
	  &= \Big(A\tran \nabla_{ABC}f(ABC) C\tran\Big)_{ij}.
  \end{align*} The second identity follows by respectively taking $p = q$ and $A
  = I_p$, or $r = s$ and $C = I_r$, which shows that both expressions evaluate
  to $U\tran \nabla f(U V) V\tran$.
\end{CustomProof}

In the context of linear neural networks, we may take $U = W_\ell$ and $V =
W_{\ell - 1}$ for some $\ell = 2, \ldots, L$ and apply Lemma
\ref{lem::matrix_chain_rule} to show that \begin{align}
    \label{eq::commutator_differential_operator}
    W_\ell\tran \nabla_{W_{\ell}} f(W_\ell W_{\ell - 1}) = \nabla_{W_{\ell - 1}}
    g(W_\ell W_{\ell - 1}) W_{\ell - 1}\tran. 
\end{align} for any differentiable $g$ that only depends on the product $W_\ell
W_{\ell - 1}$. Fix a stationary point $(W_1, \ldots, W_L)$ of
\ref{eq::fctal_general}. All partial derivatives must vanish at the point,
meaning \begin{align*}
  0 &= W_\ell\tran \nabla_{W_{\ell}} \bigg(F\big(W_L \cdots W_1\big) +
  \Tr\Big(\mathrm{Cov}\big(\wtmweight X\big)\Big)\bigg)\\
  &= \nabla_{W_{\ell-1}} \bigg(F\big(W_L \cdots W_1\big) +
  \Tr\Big(\mathrm{Cov}\big(\wtmweight X\big)\Big)\bigg) W_{\ell-1}\tran
\end{align*} for any $\ell = 2, \ldots, L$. 

Recall the definitions $Q_0 =\Sigma_{\bfX}$ and \begin{align}
  \label{eq::Q_rec_app}
  Q_\ell = W_\ell Q_{\ell - 1} W_\ell\tran + \eta^2 \Tr(Q_{\ell - 1}) \cdot
  I_{d_\ell}, \qquad \ell = 1, 2, \ldots, L 
\end{align} as well as the identity $\Tr\big(\mathrm{Cov}(\wtmweight X)\big) =
\Tr(Q_L) + \Tr(W X X\tran W)$ from \eqref{eq::reg_rewrite_gen}. By definition,
$Q_L$ computes a function of $(W_1, \ldots, W_L)$, and we may write
$\nabla_{\weight_\ell} \Tr(Q_L)$ for the gradients of its trace. Invoking
\eqref{eq::commutator_differential_operator}, we then obtain \begin{align}
  \label{eq::tr_fwd_grad}
  W_\ell\tran \nabla_{W_{\ell}} \Tr(Q_L)
  = \nabla_{W_{\ell - 1}} \Tr(Q_L) W_{\ell - 1}\tran.
\end{align} To proceed, we now derive more explicit expressions for the
gradients in the previous equation. Write $\partial_{\ell, ij}^W$ as a
short-hand notation for the differential operator $\partial / \partial W_{\ell,
ij}$ acting both on scalar and matrix-valued functions of $\weight_\ell$, then
we may exchange $\partial_{\ell, ij}^W$ with the linear trace operator. Together
with the chain rule, this implies \begin{align}
  \label{eq::tr_grad_eq}
  \partial_{\ell, ij}^W \Tr(Q_L) = \Tr\Big(\nabla_{Q_\ell} \Tr(Q_L)\tran
  \partial_{\ell, ij}^W Q_\ell\Big)
\end{align} with $\partial_{\ell, ij}^W Q_\ell$ the matrix of entry-wise partial
derivatives of $Q_\ell$.

Let $E_{ij}$ denote the standard basis of $\R^{d_\ell \times d_{\ell - 1}}$,
meaning each $E_{ij}$ has a single non-zero entry equaling $1$ in position $(i,
j)$. The recursion \eqref{eq::Q_rec_app} ensures that $Q_{\ell - 1}$ does not
depend on $W_\ell$ and in turn we compute
\begin{align*}
  \partial_{\ell, ij}^W Q_\ell = E_{ij} Q_{\ell - 1} W_\ell\tran +  W_\ell
  Q_{\ell - 1} E_{ij}\tran. 
\end{align*} Induction on $\ell$ shows that each $Q_\ell$ defines a symmetric
matrix, a property inherited by $\nabla_{Q_\ell} \Tr(Q_L)$. Combined with
\eqref{eq::tr_grad_eq}, this leads to \begin{align*}
  \partial_{\ell, ij}^W \Tr(Q_L) = \Tr\Big(\nabla_{Q_\ell}
  \Tr(Q_L)\tran \partial_{\ell, ij}^W Q_\ell\Big) = 2 \cdot \Tr
  \Big(E_{ij}\tran \nabla_{Q_\ell} \Tr(Q_L) W_\ell Q_{\ell - 1}\Big)
\end{align*} Note that $\Tr(E_{ij}\tran A) = A_{ij}$ for any matrix $A$ of
suitable dimension. In matrix form, the previous display then reads $
\nabla_{W_\ell} \Tr(Q_L) = 2 \cdot \nabla_{Q_\ell} \Tr(Q_L) W_\ell Q_{\ell - 1}
W_{\ell - 1}\tran$, with an analogous identity for $\nabla_{W_{\ell - 1}}
\Tr(Q_L)$ following by symmetry. Combined with \eqref{eq::tr_fwd_grad}, this
proves \begin{align}
  \label{eq::tr_comm_fwd}
  W_\ell\tran \big(\nabla_{Q_\ell} \Tr(Q_L)\big) W_\ell Q_{\ell - 1}
  = \nabla_{Q_{\ell - 1}} \Tr(Q_L) W_{\ell - 1} Q_{\ell - 1} W_{\ell - 1}\tran.
\end{align} Suppose for now that the second claim of Theorem
\ref{thm::gen_weight_bal} holds, namely $B_\ell = \nabla_{Q_\ell} \Tr(Q_L)$. In
this case, \eqref{eq::tr_comm_fwd} and the recursive definition
\eqref{eq::Q_rec_app} implies the first part of the theorem since \begin{align*}
  &W_\ell\tran B_\ell W_\ell \Big(W_{\ell - 1} Q_{\ell - 2} W_{\ell - 1}\tran +
  \eta^2 \Tr(Q_{\ell - 2}) \cdot I_{d_\ell - 1}\Big) \\
  =\ &\Big(W_\ell\tran B_\ell W_\ell + \eta^2 \Tr(B_\ell) \cdot I_{d_{\ell -
  1}}\Big) W_{\ell - 1} Q_{\ell - 2} W_{\ell - 1}\tran,
\end{align*} which simplifies to the desired identity \eqref{eq::wb_thm_conc_1}.

To complete the proof, it now suffices to show that $B_\ell = \nabla_{Q_\ell}
\Tr(Q_L)$, which we will accomplish via induction on $\ell$. By definition, $B_L
= I_{d_L}$ matches $\nabla_{Q_L} \Tr(Q_L)$, so we may suppose the claim holds
true for some $\ell > 1$. As $B_\ell$ results from the backward analog of the
forward recursion \eqref{eq::Q_rec_app}, it must be symmetric. Defining the
partial differential operators $\partial^Q_{\ell, ij}$ in the same way as
$\partial^W_{\ell, ij}$, the chain rule implies \begin{align*}
  \partial^Q_{\ell - 1, ij} \Tr(Q_L) &= \Tr\big(B_\ell\tran \partial^Q_{\ell -
  1, ij} Q_\ell\big) \\
  &= \Tr\Big(B_\ell\tran \big(W_\ell E_{ij} W_\ell\tran + \eta^2 \bfone_{\{i =
  j\}} \cdot I_{d_\ell}\big)\Big) \\
  &= \Tr\big(E_{ij}\tran W_\ell\tran B_\ell  W_\ell\big) + \eta^2 \bfone_{\{i =
  j\}} \cdot \Tr(B_\ell),
\end{align*} where $\bfone_{\{i = j\}}$ denotes the indicator of $i = j$.
Together with the recursive definition $B_{\ell-1} = W_\ell\tran B_\ell W_\ell +
\eta^2 \Tr(B_\ell) \cdot I_{d_\ell}$ this ensures the induction step
$\nabla_{Q_{\ell - 1}} \Tr(Q_L) = W_\ell\tran B_\ell W_\ell + \eta^2 \Tr(B_\ell)
\cdot I_{d_\ell} = B_{\ell - 1}$ and thereby completes the argument.

\subsection{Proof of Lemma \ref{lem::comm_three}}

Recall that $W_\ell$ is a $(d_\ell \times d_{\ell - 1})$-matrix for each $\ell =
1, 2, 3$. By definition, $Q_0 = X X\tran$ and $B_3 = I_{d_3}$, so Theorem
\ref{thm::gen_weight_bal} implies the two generalized weight balancing equations
\begin{align*}
  \Tr(X X\tran) \cdot W_2\tran B_2 W_2 &= \Tr(B_2) \cdot W_1 X X\tran W_1\tran\\
  \Tr(Q_1) W_3\tran W_3 &= d_3 \cdot W_2 Q_1 W_2\tran
\end{align*} where $B_2 = W_3\tran W_3 + \eta^2 d_3 \cdot I_{d_3}$ and $Q_1 =
W_1 X X\tran W_1\tran + \eta^2 \Tr(X X\tran) \cdot I_{d_1}$. Setting $\alpha =
\Tr(X X\tran)$, the conditions turn into \begin{align}
  \alpha \cdot W_2\tran W_3\tran W_3 W_2 + \alpha d_3 \eta^2 \cdot W_2\tran W_2
  &= \big(\norm{W_3}^2 + d_3 d_2 \eta^2\big) \cdot W_1 X\tran X  W_1\tran
  \label{eq::gen_bal_1} \\
  \big(\norm{W_1 X}^2 + \alpha d_1\eta^2 \big) \cdot W_3\tran W_3 &= d_3 \cdot
  W_2 W_1 X X\tran W_1\tran W_2\tran + \alpha d_3 \eta^2 \cdot  W_2W_2\tran.
  \label{eq::gen_bal_2}
\end{align} Replacing $W_3\tran W_3$ in \eqref{eq::gen_bal_1} with the
expression from \eqref{eq::gen_bal_2} yields the equation \begin{align*}
  &\dfrac{\alpha d_3}{\norm{W_1 X}^2 + \alpha d_1 \eta^2} W_2\tran W_2 W_1
  X X\tran W_1\tran W_2\tran W_2 + \dfrac{\alpha^2 d_3 \eta^2}{{W_1 X}^2 +
  \alpha d_1\eta^2} \big(W_2\tran W_2\big)^2 + \alpha d_3 \eta^2 \cdot
  W_2\tran W_2 \\
  =\ &\big(\norm{W_3}^2 +d_3 d_2 \eta^2\big) W_1 X X\tran W_1\tran.
\end{align*} Applying Lemma \ref{lem::mat_commute} below with $A = W_1 X X\tran
W_1\tran$ and $B = W_2\tran  W_2$ now proves the first commutator relationship
$W_2\tran W_2 W_1 X X\tran W_1\tran = W_1 X X\tran W_1\tran W_2\tran W_2$. The
second commutator relationship can be derived analogously.

\begin{lemma}
\label{lem::mat_commute}
  Let $A$ and $B$ denote positive semi-definite matrices of the same dimension.
  Suppose there are positive coefficients $\lambda$, $\mu$, and $\nu0$ such that
  $A = \lambda^2 B A B + \mu B + \nu B^2$, then $A$ and $B$
  commute. 
\end{lemma}

\begin{CustomProof}
  Write $B = U\Sigma U\tran$ for a singular value decomposition of $B$, with $U$
  orthogonal and $\Sigma$ diagonal with non-negative entries. Let $C = U\tran A
  U$, then the identity $A = \lambda^2 BAB +\mu B+\nu B^2$ implies $C =
  \lambda^2 \Sigma C \Sigma + \mu \Sigma + \nu \Sigma^2$. If $\Sigma_{jj} \geq
  \lambda\inv$ for some index $j$, then the identity for $C$ would result in the
  contradiction \begin{align*}
    C_{jj} < \lambda^2 \Sigma_{jj}^2 C_{jj} + \mu \Sigma_{jj} + \nu
    \Sigma_{jj}^2.
  \end{align*} We conclude that the map $A \mapsto \lambda^2 B A B$ contracts
  and so interpreting $A = \lambda^2 B A B + \mu B + \nu B^2$ as a recursion
  shows that $A$ can be expressed as a convergent sequence $\sum_{\ell =
  1}^\infty r_\ell B^\ell$ with suitable coefficients $r_\ell$, proving that $A$
  and $B$ commute. 
\end{CustomProof}

\section{Simulation Results}
\label{sec::sim}

We consider a linear regression loss $\norm{Y - X \bfw}_2^2$ with $d = 500$
features and a sparse ground truth parameter vector $\starweight$ with $s = 25$
non-zero entries featuring independent $\mathcal{N}(0,4)$ distributions.

Training and test data are generated independently from a standard normal
distribution with $n_{\textrm{train}} = 200$ and $n_{\textrm{test}} = 100$
samples. We then set $Y_{\textrm{train}} = X_{\mathrm{train}} \starweight +
\bm{\varepsilon}_{\mathrm{train}}$, where $\bm{\varepsilon}_{\mathrm{train}}
\sim \calN(0, 0.5 \cdot I_{n_{\textrm{train}}})$, with $Y_{\textrm{test}}$
defined analogously. 

We compare different types of optimization and regularization strategies for
training a diagonal linear network of depth $L = 3$. The effective model
corresponds to a linear predictor given by the product $\weight_L \odot \cdots
\odot \weight_1$ of the layer-wise weight vectors. All models are trained using
SGD with a fixed learning rate $\alpha = 0.005$, batch size $10$, and $50000$
training steps. The weights are initialized close to one with small normal
perturbations, to verify that S-SAM enforces feature learning in a setting where
(S)GD computes a kernel solution \citep{woodworth_gunasekar_et_al_2020}. As a
comparison, we run the same algorithms with weights initialized close to zero
with small normal perturbations, a setting in which unregularized (S)GD also
performs feature learning. 

We consider the following optimizers: unregularized SGD, SGD with weight decay,
and SGD with SAM noise. As baselines, we also include ridge regression and LASSO
estimators, the former computed via a closed-form solution and the latter
through coordinate descent.

Hyperparameters (ridge and LASSO regularization strength, weight decay
coefficient, and noise level for S-SAM) are selected using a validation split
comprising $20\%$ of the training data. The selected hyperparameters are then
fixed for all subsequent experiments. For S-SAM and weight decay, we tune the
noise level parameter $\eta \in \{0.3, 0.5, 0.7\}$. For ridge regression the
regularization strength $\lambda$ is chosen from $\{0.0001, 0.001, 0.01, 0.1, 1,
10, 100\}$ and for LASSO from $\{0.0001, 0.001, 0.01, 0.1, 1, 10\}$.

To account for stochasticity of the training procedure, all methods are
evaluated over 10 independent runs with different random seeds. For each
computed estimator $\widehat{\weight}$, we report the
mean of the training and test mean squared errors (MSE) \begin{align*}
  \mathrm{MSE}_{\textrm{train} / \textrm{test}}(\widehat{\bfw}) =
  \dfrac{1}{n_{\textrm{train} / \textrm{test}}} \sum_{i = 1}^{n_{\textrm{train}
  / \textrm{test}}} \Big(\mathbf{X}_{\textrm{train} / \textrm{test}, i}\tran
  \widehat{\bfw} - Y_{\textrm{train} / \textrm{test}, i} \Big)^2
\end{align*} across the runs. The performance of ridge regression and LASSO is
deterministic\footnote{While the coordinate descent steps in LASSO can feature
randomness, standard implementations such as \texttt{scikit-learn}
\citep{pedregosa_varoquax_2011} default to a deterministic update rule.} and
therefore reported without averaging.

\begin{figure}[tb]
  \begin{subfigure}[t]{0.4\textwidth}
      \centering
      \includegraphics[width=\textwidth]{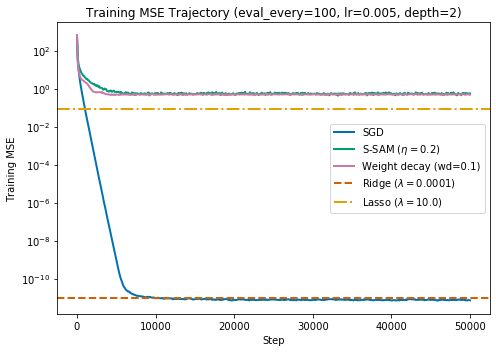}
      \caption{Training error, weights initialized close 1}
      \label{fig::d2_init1_train}
  \end{subfigure}
  \hfill
  \begin{subfigure}[t]{0.4\textwidth}
      \centering
      \includegraphics[width=\textwidth]{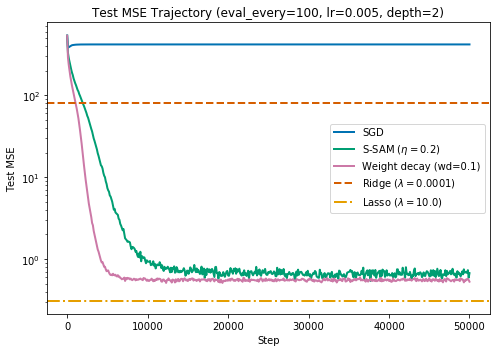}
      \caption{Test error, weights initialized close 1}
      \label{fig::d2_init1_test}
  \end{subfigure}
  \caption{Training errors (a) and test errors (b) for a diagonal linear network
  with 2 hidden layers and weights initialized closed to one. The $y$-axes are
  shown in $\log$-scale.}
  \label{fig::depth_2_setting}
\end{figure}

The training curves in Figure \ref{fig::depth_3_setting} indicate that for $L =
3$ the benefit of S-SAM over SGD, with or without weight decay, depends on the
initialization of the weights in the network. The size of the initialization
determines whether unregularized SGD operates in the kernel or feature learning
regime \citep{woodworth_gunasekar_et_al_2020}, with only the feature learning
solution able to capture the sparse ground truth, see Figure
\ref{fig::d3_init0_test}. As shown in Figure \ref{fig::d3_init1_test}, S-SAM
outperforms both SGD and SGD with weight decay for large initializations. The
induced regularizer, beyond simple parameter shrinkage, still promotes sparse
solutions when the initial predictor is dense, illustrating the importance of
the landscape regularization in Theorem \ref{thm::crit_set}. Adding weight decay
also improves over standard SGD, but its uniform shrinkage is less effective
than the additional partial-product regularization induced by S-SAM. Conversely,
small initializations cause S-SAM and SGD to behave similarly, while weight
decay can be detrimental by preventing escape from the saddle near zero. As
shown in Figure \ref{fig::depth_2_setting}, the distinction between S-SAM and
weight decay almost disappears for shallow networks as S-SAM reduces to weight
decay when matching the regularization strength, aligning with our theoretical
findings.

\section{Convergence of Gradient Descent}
\label{sec::gd_conv}

We collect and discuss some fundamental results relating to the convergence of
(stochastic) gradient descent towards stationary points of the underlying loss
function.

\subsection{{\L}ojasiewicz's Theorem and Deterministic Gradient Descent}

{\L}ojasiewicz's Theorem forms a cornerstone in the analysis of optimization
methods. It concerns the asymptotic behavior of trajectories along the gradient
vector field generated by suitable functions.

\begin{theorem}[\cite{lojasiewicz_1984}]
  \label{thm::loj_cont}
  If $f : \R^d \to \R$ is analytic and $\rmd \bftheta_t/ \rmd t = - \nabla
  f(\bftheta_t)$ for all $t \geq 0$, then $\bftheta_t$ either diverges to
  infinity as $t \to \infty$, or the limit $\bftheta_\infty$ exists and is a
  critical point of $f$.
\end{theorem}

If the orbit of $\bftheta_t$ stays bounded, the theorem ensures convergence to a
critical point of $f$. In our case, the function $f$ will be the squared
distance to observed data points, or some penalized version of it. Note that
analyticity and even differentiability may be dispensed with, provided the
function satisfies a so-called Kurdyka-{\L}ojasiewicz (KL) inequality around its
critical points. See \cite{bolte_daniilidis_et_al_2006} for more details and a
{\L}ojasiewicz-type theorem for sub-gradient dynamical systems.

The discrete-time equivalent of {\L}ojasiewicz's Theorem requires an extra
condition and provides a slightly weaker statement; see Section 3 of
\cite{absil_mahoney_et_al_2005} for more details.

\begin{definition}[Strong Descent Condition]
  \label{def::strong_desc}
  Suppose $f : \R^d \to \R$ is differentiable. A sequence $\bftheta_0,
  \bftheta_1, \ldots$ in $\R^d$ satisfies the strong descent condition with
  respect to $f$, if \begin{align*}
    f(\bftheta_k) - f(\bftheta_{k + 1}) &\geq \delta \cdot \norm*{\nabla
    f(\bftheta_k)}_2 \cdot \norm*{\bftheta_k - \bftheta_{k + 1}}_2\\
    f(\bftheta_k) = f(\bftheta_{k + 1}) &\implies \bftheta_k = \bftheta_{k + 1}
  \end{align*} for some $\delta > 0$ and all sufficiently large $k$.
\end{definition}

The sequence $\bftheta_k$ is usually generated by some form of gradient descent,
but the convergence result may be stated without explicit reference to any
recursive properties. As for the gradient flow, various generalizations to more
general function classes exists, see for example
\cite{frankel_garrigos_et_al_2015}.

\begin{theorem}[\cite{absil_mahoney_et_al_2005}, Theorem 3.2]
  \label{thm::loj_disc}
  Suppose $f : \R^d \to \R$ is analytic. If a sequence $\bftheta_0, \bftheta_1,
  \ldots$ in $\R^d$ satisfies the strong descent condition with respect to $f$,
  then $\bftheta_k$ either diverges as $k \to \infty$, or the limit
  $\bftheta_\infty$ exists and is well-defined.
\end{theorem}

Beware that the theorem does not claim for the limit $\bftheta_\infty$ to be a
critical point of $f$; this property still needs to be established in each
specific application. In the gradient descent setting, a standard argument shows
convergence to a critical point by combining boundedness of the iterates with
the first part of the strong descent condition \ref{def::strong_desc}. A proof
is provided for completeness.

\begin{lemma}
  \label{lem::gd_loj}
  Fix $\bftheta_0 \in \R^d$ and a sequence $\alpha_k > 0$ with $\sum_{j =
  0}^\infty \alpha_j = \infty$. Suppose $f : \R^d \to \R$ is analytic, with
  bounded gradient descent iterates $\bftheta_{k + 1} = \bftheta_k - \alpha_k
  \cdot \nabla f(\bftheta_k)$ that satisfy \begin{align*}
    f(\bftheta_k) - f(\bftheta_{k + 1}) \geq \delta \alpha_k \cdot
    \norm[\big]{\nabla f(\bftheta_k)}_2^2
  \end{align*} for some $\delta > 0$, then the limit $\bftheta_\infty = \lim_{k
  \to \infty} \bftheta_k$ exists and is a critical point of $f$. 
\end{lemma}

\begin{CustomProof}
  By definition, $\bftheta_k - \bftheta_{k + 1} = \alpha_k \cdot \nabla
  f(\bftheta_k)$, so the inequality in Definition \ref{def::strong_desc} may be
  rephrased as $f(\bftheta_k) - f(\bftheta_{k + 1}) \geq \delta \alpha_k \cdot
  \norm{\nabla f(\bftheta_k)}_2^2$. Moreover, $f(\bftheta_k) = f(\bftheta_{k +
  1})$ implies $\norm{\nabla f(\bftheta_k)}_2 = 0$ by the same inequality. By
  definition, $\bftheta_{k + 1} - \bftheta_k = \bfzero$ whenever the gradient
  vanishes, proving that $f$ satisfies the strong descent conditions.

  Together with boundedness of the iterates, Theorem \ref{thm::loj_disc} now
  ensures existence of the limit $\bftheta_\infty$, so it remains to show that
  $\bftheta_\infty$ is a critical point of $f$. As $f$ is analytic and the
  trajectory of $\bftheta_k$ is bounded, $\lim_{k \to \infty} f(\bftheta_k) =
  f(\bftheta_\infty)$ and $\lim_{k \to \infty} \nabla f(\bftheta_k) = \nabla
  f(\bftheta_\infty)$ are both bounded. Moreover, \begin{align*}
    f(\bftheta_0) - f(\bftheta_{k + 1}) = \sum_{j = 0}^k f(\bftheta_j) -
    f(\bftheta_{j + 1}) \geq \delta \cdot \sum_{j = 0}^k \alpha_j \cdot
    \norm[\big]{\nabla f(\bftheta_j)}_2^2
  \end{align*} so taking the limit as $k \to \infty$ leads to $f(\bftheta_0) -
  f(\bftheta_\infty) \geq \delta \cdot \sum_{j = 0}^\infty \alpha_j \cdot
  \norm{\nabla f(\bftheta_j)}_2^2$. By construction, the left-hand side is
  positive and finite. Since $\sum_{j = 0}^\infty \alpha_j = \infty$, the
  right-hand side can only satisfy the upper-bound if $\lim_{k \to \infty}
  \norm{\nabla f(\bftheta_k)}_2 = 0$.
\end{CustomProof}

\subsection{Projected Gradient Descent with Stochastic Perturbations}

General convergence theorems for stochastic gradient descent rely on adjusting
the sequence of step-sizes so that the noise cannot derail the underlying
deterministic algorithm from its trajectory towards a stationary point of the
loss function. This idea has its roots in \cite{robbins_monro_1951}, but may be
generalized to very general classes of noisy approximations of gradient descent
\citep{davis_drusvyatskiy_et_al_2020, mertikopoulos_hallak_et_al_2020,
dereich_kassing_2024}. For non-convex losses, some quantitative convergence
results exist under regularity conditions on the noise level
\citep{fehrman_gess_et_al_2020, khaled_richtarik_2023}, as well as some results
on the asymptotic distribution in a constant step-size regime
\citep{azizian_iutzeler_et_al_2024}

We will mainly need a result for projected gradient descent. Fix a closed subset
$C$ of $\R^d$ and let $\Pi_C$ denote the set-valued projection operator, meaning
$\bfy \in \Pi_C(\bfx)$ whenever $\norm{\bfy - \bfx}_2 = \inf_{\bfz \in C}
\norm{\bfz - \bfx}_2$. Closure of $C$ then ensures $\bfy \in C$. Note that the
collection of points at which $\Pi_C$ contains more than one element has
Lebesgue measure zero \citep[Corollary 2.44 of][]{fletcher_moore_2015} and
$\Pi_C$ is always single-valued for a compact convex set \citep[Corollary 2.34
of][]{fletcher_moore_2015}.

Fix $\bftheta_0 \in \R^d$, a continuously differentiable loss function $f : \R^d
\to \R$, and a sequence of step-sizes $\alpha_k > 0$. Let $\zeta : \R^d \times
\R^d \to \R$ denote any measurable function and $\bfxi_k \sim \bfxi$ a sequence
of i.i.d.\ random vectors in $\R^d$. For any set $C$, the noisy gradient descent
recursion projected onto $C$ now takes the form \begin{align}
  \label{eq::proj_grad}
  \bftheta_{k + 1} \in \Pi_C\Big(\bftheta_k - \alpha_k \cdot \big(\nabla
  f(\bftheta_k) + \zeta(\bftheta_k, \bfxi_k)\big)\Big),
\end{align} where a measurable selection of $\Pi_C$ must be made at multi-valued
points. The function $\zeta$ models the gradient noise and accounts for its
dependence on the current parameter location $\bftheta_k$. We may adapt Theorem
6.2 of \cite{davis_drusvyatskiy_et_al_2020} to get a convergence result for the
projected gradient descent \eqref{eq::proj_grad}.

\begin{theorem}
  \label{thm::proj_grad_conv}
  Suppose the following hold: \begin{enumerate}
    \item The set $C$ is compact.
    \item Both $C$ and the graph of $f$ are definable by polynomial
      inequalities.
    \item The step sizes $\alpha_k$ satisfy \begin{align*}
        \sum_{j = 0}^\infty \alpha_j &= \infty\\
        \sum_{j = 0}^\infty \alpha_j^2 &< \infty.
      \end{align*}
    \item For every $\bftheta \in \R^d$ and every convergent sequence
      $\bfvartheta_k \in \R^d$, \begin{align*}
        \E\Big[\zeta(\bftheta, \bfxi)\Big] &= 0\\
        \E\Bigg[\norm[\big]{\nabla f(\bftheta) + \zeta(\bftheta,
        \bfxi)}^2_2\Bigg] &\leq p(\bftheta)\\
        \E\Bigg[\sup_{k \in \geqZ}\ \norm[\big]{\nabla f(\bfvartheta_k) +
        \zeta(\bfvartheta_k, \bfxi)}^2_2\Bigg] &< \infty,
      \end{align*} where $p : \R^d \to \geqR$ stays bounded on bounded sets.
  \end{enumerate} Then, the function values $f(\bftheta_k)$ converge as $k \to
  \infty$ and every limit point $\bftheta_*$ of the iterates $\bftheta_k$
  satisfies $0 \in \nabla f(\bftheta_*) + N_C(\bftheta_*)$, where $N_C(\ \cdot\
  )$ denotes the Clarke normal cone with respect to $C$.
\end{theorem}

For further details on normal cones in non-smooth settings, see Chapter 2 of
\cite{clarke_1990}. When $C$ is convex, the Clarke normal cone reduces to the
usual normal cone $N_C(\bfx) = \big\{\bfy \in \R^d \mid \iprod{\bfy, \bfz -
\bfx} \leq 0 \mbox{ for all } \bfz \in C\big\}$, as frequently used in convex
analysis \citep[Proposition 2.9 of][]{clarke_2013}.

Sets that are definable in terms of polynomial inequalities are also called
semi-algebraic, see Chapter 2 of \cite{bochnak_coste_et_al_1998} for more
details. This requirement on $C$ and the graph of $f$ may be loosened to sets
definable in a so-called $o$-minimal structure. Most functions and sets
encountered in practice are definable in this way, see \cite{kurdyka_1998} for
an overview and further properties that make this wide class of objects useful
in the context of optimization.

\begin{CustomProof}[Proof of Theorem \ref{thm::proj_grad_conv}]
  We start by rephrasing the projected gradient descent via the proximal map
  \begin{align*}
    \bftheta_{k + 1} \in \argmin_{\bfvartheta \in C} \left(g(\bfvartheta) +
    \dfrac{1}{2 \alpha_k} \cdot \norm[\big]{\bftheta_k - \alpha_k \cdot
    \big(\nabla f(\bftheta_k) + \zeta(\bftheta_k, \bfxi_k)\big) -
    \bfvartheta}_2^2\right),
  \end{align*} where $g$ is simply the zero function. Consequently, $\bftheta_k$
  follows the proximal descent for the composite loss $f(\bftheta) +
  g(\bftheta)$, subject to the constraint $\bftheta \in C$. Note that
  $\bftheta_{k + 1}$ must be a measurable selection whenever the proximal map is
  multi-valued, which is always possible for any $g$ bounded from below
  \citep[Exercise 14.38 in][]{rockafellar_wets_1998}.

  Now, if all properties listed in Assumption E of
  \cite{davis_drusvyatskiy_et_al_2020} hold, then their Corollary 6.4 implies
  the desired result. The assumptions relating to properties of the function $g$
  are automatically satisfied for the zero function. Further, $\sup_{k \in
  \geqZ} \norm{\bftheta_k}_2 < \infty$ due to compactness of $C$, which is the
  final piece of the puzzle and Corollary 6.4 of
  \cite{davis_drusvyatskiy_et_al_2020} completes the proof.
\end{CustomProof}

In the previous theorem, the condition $0 \in \nabla f(\bftheta_*) +
N_C(\bftheta_*)$ at limit points $\bftheta_*$ accounts for the case where the
iterates $\bftheta_k$ asymptotically stick to the boundary of $C$ due to the
negative gradient vector field $- \nabla f$ pulling the iterates towards a point
outside the set. For sufficiently well-behaved functions this can be avoided. A
coercive function $f$, for example, has compact sub-level sets and a standard
argument shows that one may project onto a ball of sufficiently large radius
that contains all critical points. For completeness, a proof of this fact is
provided in the following lemma.

\begin{lemma}
  \label{lem::proj_grad_conv_ball}
  In the setting of Theorem \ref{thm::proj_grad_conv}, take $C = \big\{\bftheta
  \in \R^d \mid \norm{\bftheta}_2 \leq r_1\big\}$ and assume the following:
  \begin{enumerate}
    \item There exists $r_2 < r_1$, such that $\norm{\bftheta}_2 \leq r_2$ for
      every critical point $\bftheta$ of $f$.
    \item If $\norm{\bftheta}_2 = r_1$, then $\iprod[\big]{\nabla f(\bftheta),
      \bftheta} \geq c$ for some $c > 0$.
  \end{enumerate} Then, in addition to the conclusions of Theorem
  \ref{thm::proj_grad_conv}, every limit point $\bftheta_*$ of the projected
  gradient descent iterates \eqref{eq::proj_grad} lies in the interior of $C$
  and satisfies $\nabla f(\bftheta_*) = 0$.
\end{lemma}

\begin{CustomProof}
  By definition, $N_C(\bftheta) = \{0\}$ whenever $\bftheta$ lies in the
  interior of $C$. Fix $\bftheta$ on the boundary of $C$ and define the
  half-space $H = \big\{\bfvartheta \in \R^d \mid \iprod{\bftheta, \bfvartheta}
  \leq 0\big\}$. If $\bfu$ lies in the interior of $H$, then there exists a
  unique $\bfv$ on the boundary of $C$ such that $\bfv - \bftheta = \lambda
  \cdot \bfu$ for some $\lambda > 0$. Specifically, $\bfv$ is the point where
  the line segment from $\bftheta$ in the direction $\bfu$ intersects the
  boundary of $C$. The only vectors $\bfw$ satisfying $\iprod{\bfu, \bfw} =
  \lambda\inv \cdot \iprod{\bfv - \bftheta, \bfw} \leq 0$ for all such $\bfu$
  are scalar multiples of $\bftheta$, meaning $N_C(\bftheta) = \{\lambda \cdot
  \bftheta \mid \lambda \geq 0\}$. Given $\bftheta$ on the boundary of $C$, take
  $\bfvartheta \in \nabla f(\bftheta) + N_C(\bftheta)$, then together with
  assumption (b) this implies \begin{align*}
    \iprod[\big]{\bfvartheta, \bftheta} = \iprod[\big]{\nabla f(\bftheta) +
    \lambda \cdot \bftheta, \bftheta} \geq c + \lambda \cdot
    \norm{\bftheta}_2^2.
  \end{align*} Hence, $\bfvartheta = \bfzero$  would contradict the inner
  product in the previous display always being bounded away from zero, so the
  only points $\bftheta_* \in C$ satisfying $\bfzero \in \nabla f(\bftheta_*) +
  N_C(\bftheta_*)$ are critical points, which by assumption lie in the interior
  of $C$. Applying Theorem \ref{thm::proj_grad_conv} now completes the proof.
\end{CustomProof}

\section{Approximating the Shrinkage Factors in Theorem \ref{thm::crit_set_B}}
\label{sec::approx_root}

As shown in the proof of Theorem \ref{thm::crit_set_C}, any non-zero solution
$\bfz_i$ for the shrinkage factors must be a root of \begin{align*}
  p(\bfz_i) = \Big(\bfz^2_i + \eta^2 \cdot \abs{\weight_{*, i}}^{- 2 /
  L}\Big)^{L - 1} \odot \bfz^{- (L - 2)}_i - 1
\end{align*} Provided that \eqref{eq::star_thresh} holds with a strict
inequality, the latter achieves a minimum value strictly below $0$, leading to
two unique roots in $(0, 1)$. Equality in \eqref{eq::star_thresh} corresponds to
a single root at $\bfz_{*, i}$ with double multiplicity. In the former case, we
may use the secant method to approximate the two unique roots to arbitrary
precision. For generic functions, convergence of the secant method requires
initialization sufficiently close to a root, see Chapter 6 of
\cite{quarteroni_sacco_et_all_2007} for more details. We have already shown that
$p$ is strictly convex over $\ggR$, with a unique minimum at $\bfz_{*, i} \in
(0, 1)$, which makes the roots amenable to approximation by the secant method.
For the sake of completeness, a full proof of this fact is provided below.

\begin{lemma}
  \label{lem::approx_root}
  Let $a < b$ and suppose $\psi : [a, b] \to \R$ is twice continuously
  differentiable, strictly increasing, and strictly convex over $(a, b)$. If
  $\psi(a) < 0 < \psi(b)$, then $\psi$ admits a unique root $x_* \in (a , b)$
  and \begin{align*}
    x_* = \lim_{k \to \infty} \  \underbrace{\phi\ \circ \cdots \circ\ \phi}_{k
    \mbox{ times}} (a)
  \end{align*} where $\phi : [a, x_*] \to \R$ is given by \begin{align*}
    \phi(x) = - \dfrac{b - x}{\psi(b) - \psi(x)} \cdot \psi(b) + b.
  \end{align*} 
\end{lemma}

\begin{CustomProof}
  Throughout this proof, we write $\psi_x = \psi(x)$. For the sake of
  contradiction, suppose $\psi$ admits two distinct roots $x_1 < x_2$. Due to
  strict convexity, the secant line $s : [a, x_2] \to \R$ connecting $(a,
  \psi_a)$ with $(x_2, 0)$ must lie strictly above the graph of $\psi$. By
  construction, $(x_1, 0) \in \{(x, y) \mid y \geq s(x)\}$ meaning the graph of
  $\psi$ must intersect $s$ to visit $(x_1, 0)$, contradicting strict convexity.

  For any $x \in [a, x_*)$, note that $\phi(x)$ is the root of the secant line
  connecting $(x, \psi_x)$ with $(b, \psi_b)$. Due to strict convexity and
  monotonicity, $\psi_x < \psi\ \circ\ \phi(x) < 0$ and so $x < \phi(x) \leq
  x_*$. Suppose $x, y \in [a, x_*]$ are distinct. Without loss of generality, we
  may assume $y > x$, in which case $\psi(x) < \psi(y) \leq 0$. Define the
  points $X = (x, \psi_x)$, $Y = (y, \psi_y)$, $\Phi_X = (\phi(x), 0)$, $\Phi_Y
  = (\phi(y), 0)$, and $B = (b, \psi_b)$. As $\psi$ is strictly convex over $(a,
  b)$, the line segment $X B$ intersects the line $t \mapsto (t, \psi_y)$ at a
  point $Z$ with first coordinate contained in $(x, y)$. Hence, the line segment
  $Z Y$ has length strictly less that $\abs{x - y}$. The triangles $BZY$ and $B
  \Phi_X \Phi_Y$ are similar by construction. Since $Z \Phi_X$ is contained in
  $Z B$, the length of $\Phi_X \Phi_Y$ is at most the length of $Z Y$. We
  conclude that \begin{align*}
    \abs[\big]{\phi(x) - \phi(y)} < \abs[\big]{x - y}.
  \end{align*} Consequently, $\phi$ is a contraction on a compact metric space
  and the recursion $x_{k + 1} = \phi(x_k)$ converges to the unique fixed-point
  of $\phi$, starting from any $x_0 \in [a, x_*]$. This completes the proof as
  $\phi(x_*) = x_*$.
\end{CustomProof}

Suppose now that $p(\bfz_{*, i}) < 1$, write $\bfz_{i, 1}$ and $\bfz_{i, 2}$ for
the two distinct roots of $p$, and assume without loss of generality that
$\bfz_{i, 1} < \bfz_{i, 2}$. Recall from Theorem \ref{thm::crit_set_B} that
\begin{align*}
  \lambda_1 = \left(\dfrac{\eta^2}{\abs{\weight_{*, i}}^{2 / L}}\right)^{(L - 1)
  / (L - 2)} \leq \bfz_{i, 1} < \bfz_{i, 2} \leq \sqrt{1 -
  \dfrac{\eta^2}{\abs{\weight_{*, i}}^{2 / L}}} = \lambda_2,
\end{align*} which implies non-negativity of $p(\lambda_1)$ and $p(\lambda_2)$.
We may in fact assume strict positivity for either inequality, otherwise we have
already found an exact root on the boundary. Applying Lemma
\ref{lem::approx_root} over $[\lambda_1, \bfz_{*, i}]$ with $p(- \bfz_{i})$ and
over $[\bfz_{*, i}, \lambda_2]$ with $p(\bfz_i)$, the exact roots are now given
by \begin{align*}
  \bfz_{i, 1} &= \lim_{k \to \infty}\ \underbrace{\phi_1\ \circ \cdots \circ\
  \phi_1}_{k \mbox{ times}} (\bfz_{*, i}) \quad \mbox{ with } \quad \phi_1(x) =
  \dfrac{x - \lambda_1}{p(- \lambda_1) - p(- x)} \cdot p(- \lambda_1) -
  \lambda_1,\\
  \bfz_{i, 2} &= \lim_{k \to \infty}\ \underbrace{\phi_2\ \circ \cdots \circ\
  \phi_2}_{k \mbox{ times}} (\bfz_{*, i}) \quad \mbox{ with } \quad \phi_2(x) =
  \dfrac{\lambda_2 - x}{p(\lambda_2) - p(x)} \cdot p(\lambda_2) + \lambda_2.
\end{align*}


\bibliography{bibliography.bib}


\end{document}